\documentclass[Afour,sageh,times]{sagej}
\pdfoutput=1 
\usepackage[titletoc]{appendix}
\usepackage{titlesec}
\usepackage{CJKutf8}
\usepackage{moreverb,url}
\usepackage{subfigure}
\usepackage{epsfig}
\usepackage{url}
\usepackage{algpseudocode}
\usepackage{algorithm}
\usepackage{mathptmx}
\usepackage{threeparttable}
\usepackage{enumerate}

\usepackage{caption}
\usepackage[colorlinks=false,bookmarksopen,bookmarksnumbered,citecolor=black,urlcolor=black]{hyperref}
\usepackage[bookmarksopen,bookmarksnumbered]{hyperref}

\newcommand\BibTeX{{\rmfamily B\kern-.05em \textsc{i\kern-.025em b}\kern-.08em T\kern-.1667em\lower.7ex\hbox{E}\kern-.125emX}}

\graphicspath{{figures/}}

\def\volumeyear{2019}

\begin{document}

\begin{CJK*}{UTF8}{gbsn}
%中文

\runninghead{Jiang et al.}
\title{Design, Control, and Applications of a Soft Robotic Arm}
% \title{Deliberation of soft robot manipulators: a new solution for real-world tasks}

%\author{Hao Jiang, Zhanchi Wang, Yusong Jin, Xiaotong Chen, Peijin Li, Yinghao Gan, Sen Lin, Xiaoping Chen\affilnum{1}}

%\affiliation{\affilnum{1}All the authors are affiliated with University of Science and Technology of China, P.R.China}

\author{Hao Jiang, Zhanchi Wang, Yusong Jin, Xiaotong Chen, Peijin Li, Yinghao Gan, Sen Lin and Xiaoping Chen}

\affiliation{Multi-Agent Systems Lab, School of Computer Science, University of Science and Technology of China, P.R.China}

\corrauth{Hao Jiang, School of Computer Science, University of Science and Technology of China, P.R.China.}

\email{jhjh@mail.ustc.edu.cn}

\begin{abstract}

\em This paper presents the design, control, and applications of a multi-segment soft robotic arm. In order to design a soft arm with large load capacity, several design principles are proposed by analyzing two kinds of buckling issues, under which we present a novel structure named Honeycomb Pneumatic Networks (HPN). Parameter optimization method, based on finite element method (FEM), is proposed to optimize HPN Arm design parameters. Through a quick fabrication process, several prototypes with different performance are made, one of which can achieve the transverse load capacity of 3 kg under 3 bar pressure. Next, considering different internal and external conditions, we develop three controllers according to different model precision. Specifically, based on accurate model, an open-loop controller is realized by combining piece-wise constant curvature (PCC) modeling method and machine learning method. Based on inaccurate model, a feedback controller, using estimated Jacobian, is realized in 3D space. A model-free controller, using reinforcement learning to learn a control policy rather than a model, is realized in 2D plane, with minimal training data. Then, these three control methods are compared on a same experiment platform to explore the applicability of different methods under different conditions. Lastly, we figure out that soft arm can greatly simplify the perception, planning, and control of interaction tasks through its compliance, which is its main advantage over the rigid arm. Through plentiful experiments in three interaction application scenarios, human-robot interaction, free space interaction task, and confined space interaction task, we demonstrate the potential application prospect of the soft arm.

\end{abstract}

%%%% insert Keyword here %%%% 

\maketitle

\section{Introduction}
%insert section Introduction here
%The 50-year history of robotics is built based on the assumption that robot structures are kinematic chains of rigid links through joints. Robot control theories and techniques have also been built perfectly on that and reached a high level of reliability, accuracy, and efficiency \citep{siciliano2016springer}, which leads to widely applications in automation, however, it is hard for rigid robots to adapt to unstructured environments, especially those with human-robot interactions \citep{rus2015design}. The emerging of soft robots brought new opportunities for robotics, taking advantage of their easier, more natural and effective interaction with real-world unstructured environments. But soft robots do not conform to rigid body assumption, and their infinite degree of freedoms (DoFs) and compliance are unhinging the existing robotics paradigm, which poses great potential and challenges to new design and control technology. In this paper, we focus on the design, control, and applications of a soft arm and try to give our perspective and methods to several key questions. How to design powerful soft arms with big strength as well as flexibility? Can soft arms be modelled and controlled like rigid robotic arms? What are the advantages of soft arms in practical applications? 

The history of robotics is built based on the assumption that robot structures are kinematic chains of rigid links through joints. Robot control theories and techniques have also been built perfectly on that and reached a high level of reliability, accuracy, and efficiency \citep{siciliano2016springer}, which leads to widely applications in automation whose environment is structured. However, it is hard for rigid robots to adapt to unstructured environments, especially those with uncertainties and human-robot interactions \citep{rus2015design}. The emerging of soft robots brings new opportunities for robotics. Infinite passive DoFs features of soft robot enable it to be passively deformed when it interacts with environments under simple actuation, which ensures the safety of robot's interaction with the environment in unstructured environment and generates diverse behaviors to simplify task execution. However, infinite actuators are difficult to realize, so the infinite DoFs of soft robot can only be achieved passively through the contact with the outside environment. Therefore, it is reasonable that the task with strong interaction will be an important application of the soft robot, which can give full play to the potential of the soft robot. In this paper, we mainly explore the application prospect of soft arm in interaction tasks and basic requirements for the design and control of soft arm are put forward. To be specific, on the one hand, soft arm is required to be able to move flexibly while sufficient force output ability is guaranteed; on the other hand, it requires the soft arm to be able to achieve stable control in the three-dimensional space under the external forces. However, there is no mature design and control method that can meet these requirements. In addition, the specific advantages of soft arm in interaction tasks and how to use them reasonably have not been explored. To fill these gaps, we focus on the design, control, and applications of a soft arm and try to give our perspective and methods to several key questions. How to design powerful soft arms with big strength as well as flexibility? Can soft arms be modeled and controlled like rigid robotic arms if there are external influences? What are the advantages of soft arms in practical applications?

\subsection{Design: Can soft arms be of strength?}
Most designs of soft robots are inspired by natural creatures and tissues, like octopus, elephant trunks, tongues, and worms, etc \citep{rus2015design}. Soft creatures often live in soil or water, using surrounding medium to support their own body weight without a skeleton \citep{kim2013soft}. Soft arms suffer from the similar problem, lacking load capacity, and some of them even cannot support themselves from gravity, which limits their applicability. However, elephant trunks are both flexible and powerful. In this paper, our aim is to build useful soft arms like trunks.

However, the elephant trunk, as an organ composed of pure muscles, is not only flexible, but also has a large strength \citep{boas1908elephant}. It is indicated that a soft arm composed of a entirely soft material is capable of achieving a large flexibility as well as force output capability.

The use of soft materials allows the soft arms to exhibit infinite passive degree of freedom. The freedom of the rigid robot is mainly depend on the number of joints, and the motors provides driving for each joints. Under the rigid body assumption, the rigid robot design is mainly focus on the design of joint arrangement and load check. The infinite passive degree of freedom of the soft arm is mainly underactuated, then the design of the soft arm, different from the rigid one, should focus on how to design the arm structure and actuator layouts, so that the soft arm exhibits large flexibility and strength.

The upper limit of the theoretical force output capability of a soft arm is determined by the level of driving force, but the actual load capacity is also related to many factors. Since most of the soft arm using low Young's modulus as the main material, the soft arm is prone to instability when subjected to force. After the critical load of buckling is reached, some components/the whole structure is still in static equilibrium, which is however unstable\citep{sun2017force}. And the components/the whole structure will generate large bending or torsional deformation which is undesired as long as an infinitesimal disturbance is applied. In practice, there is no perfect structure, and there are always more or less disturbances, so the manipulator is prone to instable when the load reaches critical load. This instability makes the actual maximum load of the soft arm much lower than the theoretical maximum load. It also poses a challenge for the control of soft arms. By designing to solve the stability problem, the force output capability of the soft arm can be improved.

Elephant trunks are composed of three types of muscles, longitudinal, radial and oblique muscle \citep{boas1908elephant,wilson1991continuum}. Bending and elongation can be achieved by muscular hydrostats mechanisms of longitudinal and radial muscles, and torsion is achieved by oblique muscles. In addition, only senior creatures have helical array oblique muscles, while low-level arthropods have crossed-fiber helical array connective tissue, which cannot provide torsion, but their are able to resist twist when internal cavities pressured \citep{kier2012diversity}. Inspired by these natural mechanisms, we regard force output ability of longitudinal, radial and torsional as the key to design a stable and powerful soft arm. We will analyze the instability of the soft arm in the main part to prove this conjecture. Based on this hypothesis, design of existing soft manipulators is analyzed as followed:

\cite{martinez2013robotic} propose a 3D tentacle-based soft material chambers inflated by pressurized fluids. \cite{marchese2016design} build a completely soft arm in a similar way, which provides radial elongation by actuation, and contraction by elasticity of non-chamber middle parts, for bending. Whereas this kind of design lacks radial and torsional forces, and its radial force generated by only elasticity of silicone rubbers is small, which restricts the mechanical efficiency respect to the desired movement associated with the bending and elongation capabilities. In addition, this design largely rely on material expansion, is highly demanding on the fabrication process. The arm will excessively expand to the thinner parts of chambers when it has uneven wall thickness. In \citep{marchese2016dynamics}, we can see the soft arm has unpredictable multiple bulbs at one segment. This design has almost no torsion output, which not only reduces the force output of the arm but also restricts the working space to a sagittal plane (or the arm will generate unpredictable torsion). For soft continuum robots, as the actuations/constraints for elongation and contraction are mostly independent, the problem of radial deformation is more evident. Parallel continuum robots developed by \cite{orekhov2017modeling} exhibits that in the case of a rod that provides axial driving force, a large nonlinear deformation occurs without intermediate radial constraint, so that the overall bending of the arm is small and the working space is limited.

Muscles can only provide contraction forces in nature organism, so creatures need hydrostats for elongation. \cite{cianchetti2011design} focus on the morphology of octopus arm and effects of longitudinal and radial muscles in muscular hydrostats, and uses cables to provide longitudinal and radial actuations to reproduce similar behaviors. \cite{cianchetti2012design} use shape memory alloys(SMA) to contract the arm in the radial direction, with cable actuation in the longitudinal direction, to simulate muscular hydrostats and implement elongation and bending. When we design soft robots, we not only have contraction-muscle like actuations (such as cable driven, SMA), but also extension-type actuations (such as pneumatic muscles and push rods), so we can achieve elongation without muscle hydrostatic mechanism. Thus radial force output can be provided by proper constraints, which is often much easier. \cite{orekhov2017modeling} use radial constraints to reduce nonlinear deflections and increase workspace and force output. \cite{li2002design} propose a cable driven continuum arm, which uses plates as radial constraints. Extensible pneumatic muscles require mesh and plastic coupler \citep{grissom2006design} or constraint frames \citep{godage2016dynamics} when used for soft arms. Otherwise, single extension pneumatic muscle is vulnerable to buckling without constraints, which reduces the load capacity of soft arms \citep{trivedi2008soft}.

Besides soft arms, soft actuators also follow the same rules. \cite{suzumori2007bending, deimel2016novel,polygerinos2015modeling} try to use cables wrapped around to constrict radial expansion, which increase mechanical efficiency. But this method needs multistep molding process, and symmetric cable wrapping to avoid torsion during inflation or breaks due to partially excessive deformation. \cite{ilievski2011soft} propose the design of embedded pneumatic networks(PN), whose radial constraint is provided by the structure and can be easily fabricated as a entirely soft robot. \cite{mosadegh2014pneumatic} demonstrate the design of fast pneumatic networks(fPN) to create faster, linear response of Pneumatic Network actuator by adding radial constraints from structures.

It, in some extent, increases the force output of the soft actuator when radial force output is improved, but unstable phenomenon still occurs when the force is large \citep{sun2017force}, which affects the overall performance. In this case, the bottleneck of arm's performance is torsion, rather than radial force output. We will present detailed analysis in the design section.

It is common that natural creatures and organisms have mechanisms for torsions, like oblique muscles generating torsions in elephant trunks and octopus arms, and anti-twist crossed-fiber helical array connective tissues in Nematodes \citep{kier2012diversity}.  But torsions are often neglected in soft robot designs. \cite{wang2018incorporate} focus on the strengthen effect of oblique muscles to soft manipulators by providing shear forces, but it neglects torsion. \cite{doi2016proposal} emulate the octopus arm structure, fabricates a flexible arm with actuation forces in longitudinal, radial and oblique directions. It proposes that oblique muscles are able to change the arm stiffness while providing active torsion, and the stiffness can be regarded as force output to some extent. However, it does not analyze the necessity of oblique muscles in soft arms. Pneumatic muscles' meshed fibre sleeve is similar to crossed-fiber helical array connective tissue, which is able to resist twist with internal pressure. But extra mechanisms are required for providing torsion constraints when multiple pneumatic muscles are used to build an arm, like Octarm. Similar to meshed fibre sleeve in pneumatic muscles, soft fingers developed in \citep{polygerinos2015soft} demonstrates the double helix entanglement of cables provides twist resistance while providing radial restraint.

Bionic Handling Assistant \citep{grzesiak2011bionic} using relative stiffer material, nylon. The stiff material provides enough constraints to twist and make it able to move freely under gravity, but it also restricts the flexibility of the manipulator and increases power consumption. Festo BionicMotionRobot \citep{festopage_gripper} uses 3D textile knitted fabric to constrain each bellow's radial expansion, a rib to provide longitudinal constraints, and cardan joints for torsion constraints, leading to 3kg payload with 3kg self weight. A similar design of trunk robot by \cite{hannan2001analysis} use cardan joints for torsion constraints and cables to actuate. These manipulators with cardan joints belong to the range of hyper-redundant robots. 

Another mechanism improving force output is to change stiffness by jamming \citep{jiang2012design, cianchetti2013stiff, cheng2012design}. The variable-stiffness designs require other actuations, such as cable-driven, for motions, but there is no force output during the stiffness variation \citep{wang2018incorporate}. 

In summary, there are lots of works and effective methods on longitudinal and radial force output of soft actuators, but research emphasis on torsion force output is lacked. And there is no work to analyze the design of the soft arm from the perspective of instability.

In this work, we derive the principle of soft arm design by analyzing the instability of the soft arm. Then we propose Honeycomb Pneumatic Networks (HPN) structure based on design principles and fast, reliable fabrication method of HPN is proposed. We further propose a design optimization method based on finite element method (FEM) simulation. The design is valuated via several experiments. The HPN Arm exhibits a payload around 3 kg under 3 bar pressure at the length of 60cm.

\subsection{Control: Can soft arms be modeled?}

In order to utilize the soft arm, effective and reliable controller is required, and this issue has been an on-going challenge in the field due to the infinite degree of freedom and nonlinear deformation of the soft arm.

Traditional rigid robot arms use joint-link model to build accurate kinematics and dynamics model, and implement fast, accurate, reliable and energy-efficient control based on the models. In order to model and control a soft arm exactly like a rigid robot, researchers have proposed many methods. Classified by the modeling method, there are two main approaches: getting model by mechanical and geometrical analysis, or getting model by learning from data. In this paper, we refer modeling as a process of getting the mapping from task space to configuration space (or joint space, actuation space), no matter it uses analysis or learning. 

Regarding analysis model-based methods, they mathematically formulate the system’s behavior using kinematics and dynamics equations\citep{mahl2014variable}. \cite{trivedi2008geometrically} use Cosserat rod theory to build accurate forward kinematics. \cite{godage2011novel} present a 3D kinematic model for multi-segment continuum arms using a novel shape function-based approach\citep{godage2011shape}, which conforms geometrically constrained structure of the arm. \cite{hannan2003kinematics} parameterize the configuration space of a 3D continuum shape as three parameters using constant curvature (CC) assumption, which simplifies an infinite dimension structure to 3D. \cite{escande2012geometric} develop the forward kinematic model for compact bionic handling assistant also based on the constant curvature assumption. \cite{marchese2014design} proposed a closed-loop curvature controller under PCC assumption and extended the method to a 3D manipulator working in sagittal plane \citep{marchese2016design}. \cite{wang2013visual,mahl2014variable,mahl2013variable} improve the model by modeling a single segment using multiple arc pieces, to create a higher dimensional representation. These methods based on the analysis model are more interpretable and reliable, but with complex deformation and nonlinear driving response, the model of soft robot is difficult to build and its precision is low.

Learning-based methods are widely applied in control of soft manipulators
owing to their universal approximation property. \cite{giorelli2013feed,giorelli2015neural}
develop a feed-forward neural network to solve the inverse kinematics for a
cable-driven manipulator. \cite{rolf2014efficient} introduce a goal babbling approach
to learn inverse kinematics of BHA. \cite{melingui2014qualitative} propose a method learning direct mapping from task space to joint space (voltage of potentiometer), which uses a neural network to learn the forward kinematics model beforehand, and use distal supervised network to invert the obtained network. \cite{melingui2015adaptive} add adaptive controllers to handle the nonlinear mapping between joint space (voltages) and actuation space (pressure in chambers). This mapping problem is common in pneumatic actuation system. Generally, learning-based methods can figure out relatively accurate models for complex structures. However, for the manipulators with many DoFs and complicated structures, a large amount of training data and a long training process is essential. 

In this paper, different from most others, we want to build model and control from actuation space. In this process, nonlinearity in actuation (viscoelasticity, etc.) is difficult to model in analytical methods, while the amount of data required by training will be too large. In fact, a better control scheme is to combine the two previous approaches in order to take the best of them. From task space to configuration space, taking advantage of high interpretability of analytical methods, we use a model-based method to set appropriate cost function and use gradient descent algorithm to figure out optimal solutions. From configuration space to actuation space, analytical methods are extremely hard to implement due to complex deformation and nonlinear actuation response. A common technic is to add sensors in configuration space and use PID controllers, but sensors for soft manipulators are not mature and hard to install. Here we want a direct mapping from actuation space and task space, we model each segment individually, where a neural network learns the relationship after training with a proper amount of data. Moreover, control precision and stability are improved by taking viscoelasticity into consideration. Compared with existing neural-network based control approaches \citep{giorelli2013feed,giorelli2015neural}, this feature makes our approach unique. Regarding viscoelastic behavior, it is a hysteresis effect of actuators and memory phenomenon of the whole structure, impairing the drive space control precision of most manipulators made of soft materials. All the above methods are based on constant curvature (CC) assumption under ideal conditions without external forces or disturbances: (1) the manipulator is uniform in shape and symmetric in actuation design, (2) external loading effects are negligible, and (3) torsional effects are minimal. Other methods are also modeled based on deterministic behavior. Actually, modeling of soft robots, without external forces or certain forces, is a mapping from actuation to task space states, which can be represented as a function. Under this condition, accurate modeling is possible. For instance, neural networks can approximate any function fed by enough data. 

Nevertheless, uncertain external forces or disturbances are very hard to be avoided in practical scenes. Due to their passive compliance, a small force can change the shape of soft robots. For instance, the shapes of a soft manipulator holding objects with different weights, under the same actuation, are usually completely different. Sometimes even self-gravity alone can also change the shape, like the shape of the root segment is affected by the gravity torque of end segments. If the manipulator only has finite DoFs, controller can be implemented with corresponding sensors to the degrees of freedom (like encoders), even with passive compliance. But soft robots have infinite DoFs in theory, which makes it unrealistic to install enough sensors. Another point is that the soft manipulator needs to have infinite number of actuators to be able to reach any theoretically possible configuration, which is actually impossible.
So the deformation by external forces cannot always be compensated and avoided. When the disturbance is uncertain, the deformation will be unpredictable. In this case, soft robots could not be accurately modeled as rigid body connected by joints like traditional rigid robots.

In the case of imprecise modeling, precise control can be achieved through the closed-loop feedback of the task space. Different proper methods can be used under disturbances and uncertainties of different amplitudes. When uncertainty is small and model is relatively accurate, open-loop control is reliable with some extent of feedback. Based on the model from learning, \cite{rolf2014efficient} give a simple correction feedback controller in task space to handle the residual error. In this work, we use a similar method to change a two-level model-based controller to a closed-loop controller, which facilitate comparison with the methods presented below. When uncertainty is large, model error is large too.  If the former feedback control method is also adopted, the jitter of the control will affect the arrival efficiency, and even the convergence may not occur. At this time, we can realize the control by the way similar to differential inverse kinematics,which is achieved by small step iteration and has low dependence on the model. As long as the movement direction guided by the model is close to the target point, the convergence can be guaranteed and the step size can be conveniently adjusted according to the accuracy of the model.  \cite{thuruthel2016learning} used a neural network to learn the Jacobian mapping for each next step and validated the model in simulation; \cite{george2017learning} validated the method using a feedback controller of a 6 DoFs tendon-driven manipulator with the existence of external forces. However, in such a learning-based method, if external disturbance makes the arm reach a posture that has not been reached in the learning process, the performance is difficult to be guaranteed, and there may be stability problems. \cite{qi2016kinematic} constructed a fuzzy-model-based controller
using estimation of Jacobian from prior knowledge based interpolations, but the estimated Jacobian matrix will not change even disturbance make the model deviates from the actual shape, which affects its performance.

Since the disturbance makes the model inaccurate, it is not necessary to elaborate an accurate model. We propose a feedback control method based on estimated Jacobian model. Compared to \citep{qi2016kinematic} which directly estimates the Jacobian of the whole manipulator, we only estimate Jacobians of individual segments, and calculate the overall result based on end effector pose geometric estimation. The advantage is that Jacobian of a single segment has a stable positive-negative sign (moving closer or farther to the goal in one step), and disturbance will only change its magnitude but not sign. Geometrical information is easy to obtain and independent to external disturbances. And if the geometrical information is ensured to be correct, the direction indicated by the calculated Jacobian matrix of the whole arm should also be correct. This calculated Jacobian matrix should be close to the real Jacobian matrix and is reliable under disturbances.

When the disturbance is too large to make sure moving direction is controllable, it’s a good choice to abandon accurate model and use control method with real-time updating and learning ability to handle large disturbances at the cost of execution speed. All the aforementioned methods can turn to an updating adaptive controller by learning again when the model error to reality is large. But when disturbance occurs frequently, it is hard to gather enough data to adapt to different configurations and the control performance is low. While the control method using strategies is easy to update and is more adaptable to large disturbances. \cite{yip2014model} estimate Jacobian constantly to implement 2D position control, which updates the Jacobian matrix by measuring position-actuation changing ratios, without a specific model. But this method lacks verification in higher dimensions. In this work, we will discuss about building controller using Q-Learning, which only requires a little amount of data due to data reusing, to get a Q matrix, which can grow better in real-time learning and become usable in control. As it requires very little data, it is adaptive to large disturbances.

In summary, we propose and implement three controllers. Modeling control based on PCC assumption, estimated Jacobian feedback control and policy control based on Q-Learning. We will compare the three methods on the same experiment platform, show their characteristics under different disturbances, and give proper controllers for different application scenarios.

\subsection{Application: What are soft arms good at?}

Rigid robot technical paradigm is developed largely based on accurate sensing, modeling, planning, and executing which is computationally expensive and cannot guarantee the safety of robot-environment interactions especially in unstructured environments \citep{siciliano2016springer}. The emerging of soft robots brings new potential to robot applications \citep{rus2015design}. A variety of potential applications of soft robots have been developed, involving locomotion, grasping, manipulation, and medical applications. 

\cite{shepherd2011multigait} use pneumatic networks to build crawling robots, which demonstrates that soft robots do not need complex design or control to generate mobility. Besides, effective motions can be achieved by emulating natural creatures \citep{marchese2014autonomous,mazzolai2012soft}. Inspired by plant growth, \cite{hawkes2017soft} design a soft robot which navigates its environment through growth. These different motions are potential to be used for different applications.

As for soft grippers, their compliance adapts more readily to various objects, simplifying grasping tasks. Relevant works have demonstrated their potential in different areas. \citep{suzumori1991development,ilievski2011soft,deimel2016novel,brown2010universal}. Soft grippers, as the main products of some companies(for example Soft Robotics), have been put into practical use in some fields, such as food sorting..

Compared with soft grippers, applications of soft manipulators are limited. The main reason is that basic design, fabrication and control methods are not complete and reliable, high load capacity hardware and stable controllers are lacking. Another point is that there has not been a unified understanding of the advantages of soft manipulators over rigid counterpart. We believe that the soft arm differs from the rigid arm in two fundamental characteristics: passive compliance and infinite freedom. Based on these two characteristics, features such as safety, flexibility, and adaptability are exhibited.

Exploiting its safety, \cite{nguyen2018soft} use a three-segment pneumatic arm to finish mobile manipulation for daily living tasks, and demonstrate properties of soft manipulators in multitasking pick and place scenarios. Soft manipulators also have medical applications, for their flexibility and safe interaction with humans. \cite{park2014design,polygerinos2013towards,hauser2017jammjoint} use soft robots as affiliated moving devices and \cite{webster2006toward,deng2013development,cianchetti2013stiff} develop soft surgical robots.

The main characteristic of soft manipulators is passive compliance, compared with rigid manipulators, and their main advantage over multiple flexible link robots and rigid robots with active compliance are infinite DoFs. However, the number of actuators for soft manipulators is limited, and the potential of passive compliance and infinite degrees of freedom can only be exploit when interact with environment. Soft manipulators demonstrate embodied intelligence and morphological computation in robot-environment interaction scenarios  \citep{paul2006morphological,hauser2011towards,cianchetti2012design}, which tend to simplify interaction tasks. We believe applications of soft robots should lie on active interaction with environments, where passive compliance and infinite DoFs have great effects.

\cite{giri2011continuum} implement whole arm under-actuated manipulation based on Octarm. \cite{marchese2016design} demonstrate soft manipulator navigate through confined space. These tasks take advantage of soft manipulator‘s interaction features, but the whole arm manipulation and the navigation through confined space are only a small part of the task involved in the interaction, there are large number of tasks in daily life require robot-environment interaction.

In this paper, we explain and demonstrate the feature of soft manipulators when interacting with environments. Specifically,  we explain the simplifying effect of soft manipulators when they interact with environments. Taking advantage of compliance of soft manipulators, their behaviors can be modified by interaction, which guides us to use the soft arm with the concept of inaccuracy. In order to demonstrate the simplifying effect and task execution ability of soft manipulators in unstructured environments, we do experiments in free space interaction tasks, confined space interaction tasks, and human-robot interaction tasks. Moreover, in free space interaction tasks, we select the tasks with different number of end effector DoFs (open drawer, shift gear, clean glass, and turn handwheel etc.), which demonstrates the soft manipulators' ability to finish tasks without relying on accurate sensing, modeling or planning. Finally, we show the compliance completeness of soft manipulators and their ability to perform interaction tasks in confined spaces.
\section{HPN Manipulator Design}
\subsection{Design principle}
In this section, in order to design soft arm with large load capacity and decent flexibility, first, we propose the evaluation metrics. Then, the deformation and force characteristic of soft arms is analyzed. Finally, several design principles are proposed, which could help to design soft arms and could be used to evaluate the design of a soft arm.
\subsubsection{Evaluation metrics.}
There are kinds of soft manipulators that achieve appreciable performance in their specialty respectively, nevertheless, there is not a universal evaluation criterion assessing and comparing their performance. In this paper, aiming at reaching a high level of load capacity with little loss in flexibility, we set these two characteristics as the main metrics.

\begin{enumerate}

\item As for load capacity, we define it as the maximum load moment of a soft manipulator when it is able to remain stable and its loaded end is on the same height with its fixed end. lift the tip to the horizontal (similar as the load capacity defined by \cite{trivedi2008optimal}).

\item As for flexibility, we define it as a manipulator’s reachable space, which consists of all the points that the tip of the manipulator can reach with another end fixed. To realize comparison between manipulators with different shapes and scales, we calculate the ratio of that space to the dimensionality’s power of its original length, as a relative flexibility.
\end{enumerate}

\begin{figure*}[htbp]
\centering
\includegraphics[width=\textwidth]{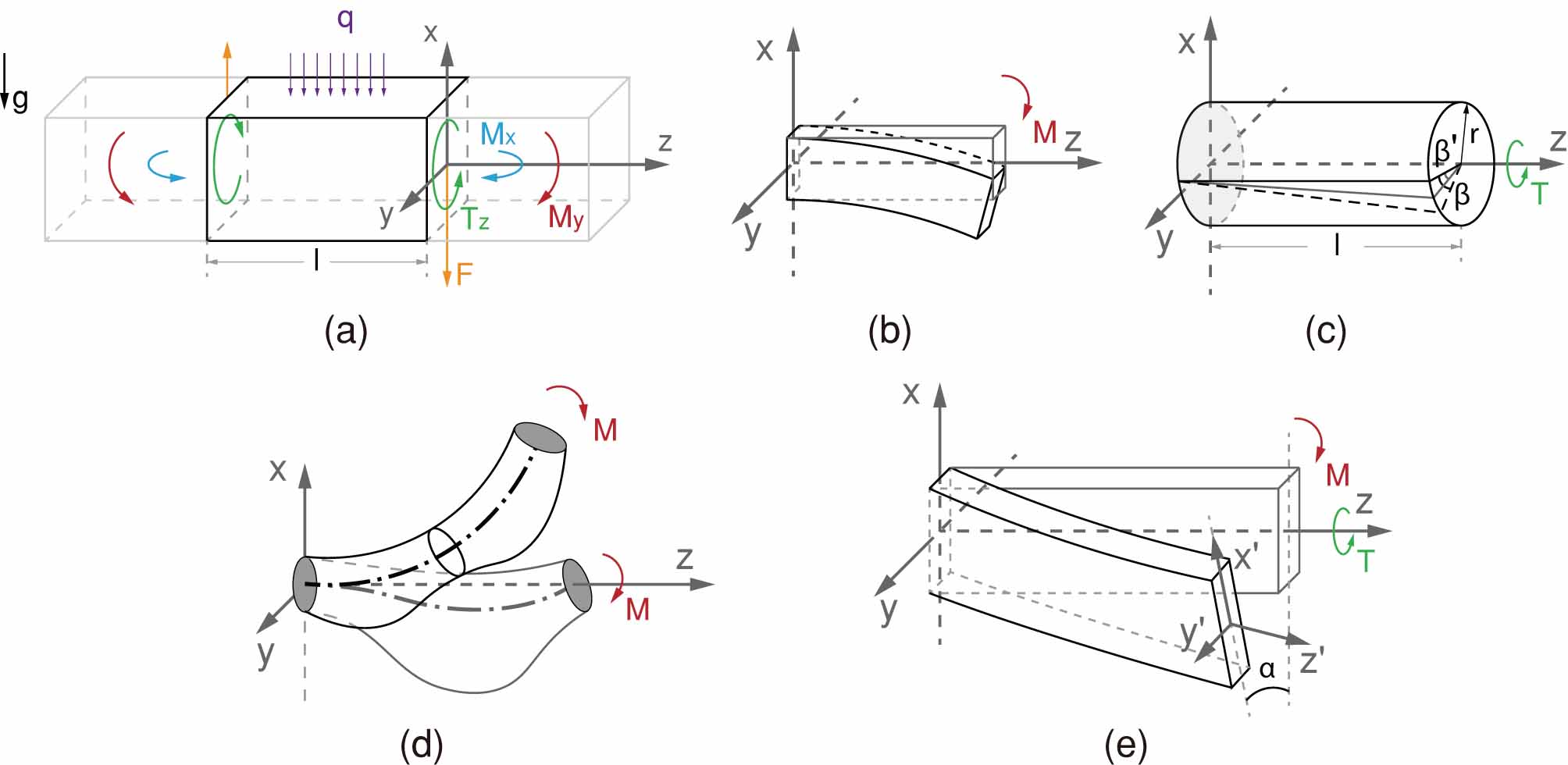}
\caption{The force and deformation analysis of the simplified beam model of soft arms under the effect of gravity. (a) demonstrates the overall force situation of the beam. (b) and (c) demonstrate the effect of actuation on bending and torsional deformation respectively.(d) illustrates the buckling of part of the arm. (e) illustrates the flexural torsional buckling of the whole arm.}
\label{beam_model}
\end{figure*}

\subsubsection{Equivalent parameters of actuation in beam model.}
As shown in Figure \ref{beam_model}(a), an infinitesimal cross section perpendicular to the longitude direction of the soft arm is considered as a homogeneous beam, and its force situation is analyzed. As for external load, here we mainly consider the bending moment on two main bending directions $M_x$ and $M_y$, the torsional torque along the longitude direction of the arm $T$, the weight of the soft arm (distributed loading) $q$, and the gravity of external load $F$.

If the soft arm is considered as a beam, the elongation length, contraction, bending and torsion of the arm is relevant to the equivalent physical parameters of the beam such as equivalent bending stiffness and equivalent torsion stiffness, which could further influence the load capacity and flexibility of the beam. Here we first figure out the relationship between actuation and the equivalent parameters. The actuation could be considered as either internal force or external force, and the choice should depend on our designing requirement. When the motion or flexibility of soft manipulator is studied, the actuation should be considered as external force, because the actuation acts on the structure and material of the manipulator and makes it deform, and the amplitude of deformation would influence its flexibility. When the interaction of the soft manipulator and its environment or the load capacity is concerned, the actuation should be considered as internal force. In this paper, we mainly care about the load capacity of the manipulator, so the actuation is considered as internal force.
\paragraph{Equivalent bending stiffness.}
%As far as bending is concerned, consider a beam subject to an external bending moment, when longitude actuation opposite to that moment is applied, the curvature of the beam will be decreased while the external bending moment remains unchanged, as shown in Figure \ref{beam_model}(b). Bending equation of beam is as follows:

Considering a beam subject to an external bending moment with longitude actuation opposite to this moment, the curvature of the beam will be decreased while the external bending moment remains unchanged, as shown in Figure \ref{beam_model}(b). Bending equation of beam is as follows:
\begin{equation}
EI = M\rho
\end{equation}
where $I$ is the moment of inertia, $E$ is Young's modulus of the material, $\rho$ is the radius of bending, $M$ is the bending moment. When the beam and actuation together is regraded as a bending beam, equivalent $E$ will increase as $\rho$ increases while $M$ and $I$ remain unchanged (the cross section is assumed to be not deformed approximately). 

The longitude actuation could be roughly divided into two kinds: one is the actuators that provide a specified force (for example, specify the pressure of a pneumatic actuator), another is the actuators that provide a specified displacement (for example, cable driven actuator and the legs of parallel continuum robot). The two kinds of actuators have different effect on the equivalent stiffness of the beam. However, to simplify the analyze, we won’t distinguish between them.

\paragraph{Equivalent torsion stiffness.}
As for oblique actuation, its force to the beam has a component which makes the beam twist. And oblique actuation could be used to control rotating of manipulators \citep{kier2012diversity, doi2016proposal, olson2018helically}. So the rotating angel caused by an external torque which makes the beam rotate in the opposite direction would decrease, as shown in Figure\ref{beam_model}(c).  The rotating formula is as follows:
\begin{equation}
GJ\theta=T
\end{equation}

where $T$ is torque, $\theta$ is the rotating angle $G$ is shear modulus, $J$ is the torsion constant of the cross section. When the beam and oblique actuation is regarded as a whole twisting beam, equivalent G will increase by oblique actuation as $\theta$ decreases while $T$ and $J$ remain unchanged.

The key problem of the mechanical design of soft manipulators is how to design corresponding structure and actuation to make soft manipulators able to work properly and not prone to stiffness failure, strength failure and stability failure under varieties of external loads. For the load capacity of soft manipulators, the main problem is stability issues under low strength and low stiffness.

\subsubsection{Buckling analysis.}
We noticed that the soft manipulator is prone to buckling \citep{sun2017force}. After the critical load of buckling is reached, the manipulator is still in static equilibrium, which is however unstable. And the manipulator will generate large bending and torsional deformation which is undesired as long as an infinitesimal disturbance is applied. In practice, there is no perfect structure, and there are always more or less disturbances, so the manipulator is sure to buckle when the load reaches critical load, which may be smaller than theoretical load capacity. Besides, after the manipulator buckles, the actual load capacity is hardly increased with the increase of the actuation on the main bending direction. So, this kind of instability would make the actual load capacity of manipulators much lower than its theoretical load capacity. Next we will consider how to increase the load capacity by designing the manipulator properly to increase its buckling critical load.

The buckling phenomenon we noticed can mainly be divided into two types, the first is the buckling of the compressed part of the manipulator while the other parts of the manipulator remains stable, and the second is the flexural torsional buckling of the whole manipulator.

Inspired by the muscle structure of elephant trunks and octopus tentacles, and cross fiber of lower animals, we suppose that the instability problem of soft manipulators could be solved by increasing the force output ability in radial, longitude and oblique directions. Here we derive the relationship between the three kinds of force output ability and the load capacity by analyzing the above-mentioned two kinds of buckling and their influencing factors. The ability of oblique actuation to provide shear force has already been studied \citep{wang2018incorporate}, so here we won’t discuss the shear force and shear deformation of the manipulator. Because few work has been done to investigate how members consists of soft materials buckle, we will use existing buckling theories for rigid members to give a rough qualitative analysis.

\paragraph{Buckling of compressed part of the manipulator.}

According to the direction of normal stress, a bending soft arm could be divided into compressed part and stretched part. %by its neutral plane. Similarly, a bending soft manipulator could also be divided into the compressed part and the stretched part. For example, for many cable driven actuators, the cables are stretched and the other structures are compressed, while for the parallel continuum robot in \citep{orekhov2017modeling},  and for soft manipulators using extending pneumatic muscle actuator (pMA) \citep{kang2013design}, \citep{grissom2006design}, and fluidic actuated chamber \citep{marchese2016design}, the actuators are compressed and other structures are stretched. 
The compressed part can be considered as a column and is prone to buckle. Figure \ref{beam_model}(d) shows the buckling of the compressed actuator of the manipulators. %Whatever it is the actuating part or the other part of the manipulator that buckles, after buckling the increase of actuation will mainly contribute to the deformation in the buckling direction and has very little effect on increasing the maximum load. And the control performance would also be affected because it is vary hard (if not totally impossible) to consider the actual effect of buckling. 
Here we use the model of compressed column to analyze this kind of buckling. When the column is fixed at one end and free at the other end, the critical load is given by Euler's formula:

\begin{equation}
    F = \frac{\pi^2EI}{(2L)^2}
\end{equation}

where $I$ is the moment of inertia, $E$ is Young's modulus of the material, $L$ is the length of the compressed beam. It can be figured out that, the critical load is proportional to the bending stiffness and is inversely proportional to the square of 2L. 

There are several methods to sovle the buckling problem. Stiffer material with larger $E$ could be used to fabricate the compressed part to increase $EI$. However, this may decrease the flexibility of the manipulator. The cross section could be properly designed to increase the moment of inertia $I$ so that $EI$ is increased. However, this could also influence the flexibility. Besides, the free length of the compressed part $L$ could be decreased. For example the arm could be divided into multi sections (see Figure~\ref{beam_model}(d)) by discrete radial constraints. Besides, increasing the radial constraint could make the stretched part, which is not easy to buckle, to provide braces for the compressed part that is easy to buckle. If the brace is continuous, involves the whole arm and is strong enough, the compressed part may be “stick” to the stretched part and is very hard to buckle by itself \citep{sun2017force}. The radial constraint could be provided by either constraint or actuation.

According to the analysis above, we could conclude that radial force or radial constraint could help to resist the buckling of compressed part of the manipulator and increase the critical load without decreasing flexibility.

\paragraph{Flexural-torsional buckling of the whole arm.}
Another kind of buckling that is common for beams, namely flexural-torsional buckling, may also influence the load capacity of the soft arm. It involves lateral displacements out of the bending plane and rotation along the shear center, as shown in figure\ref{beam_model}(e). In the analyse, we will roughly consider the whole manipulator as a simply supported bending beam without considering the shear stress, and the load is the bending moment at the two ends. Then we get the critical load of bending moment

\begin{equation}
critical~load = \pi\frac{\sqrt{GJE_yI_y}}{L}
\end{equation}

(The boundary conditions only require the torsion at the two endpoints be zero) \citep{wang2004exact},where $E_y$ is equivalent Young’s modulus, $I_y$ is moment of inertia for bending along x-axis, $E_yI_y$ is equivalent lateral bending stiffness, $G$ is shear modulus, $J$ is torsion constant, $GJ$ is equivalent torsion stiffness and $L$ is the length of the beam.

Based on the analysis above, we could derive some feasible methods that may be able to prevent this kind of buckling: increase the bending stiffness $E_yI_y$ and torsion stiffness $GJ$ while decrease the length of the soft manipulator $L$. Next we will analyze them one by one.

The bending stiffness $E_yI_y$ could be increased by using materials with large Young's modulus, or design the cross section properly to increase the moment of inertia $I_y$ or increase the longitude actuation. However, the first two methods may decrease the flexibility of the manipulator. When designing a soft finger, we only need one DoF, so we could increase the moment of inertia without much loss in flexibility. For example a layer of in-extensible material could be inserted.

There are two methods to increase torsion stiffness. An oblique actuation or constraint could be installed to render the soft manipulator twist actively. The torsion constant $J$ could be increased by properly design the structure to limit the torsional deformation of soft manipulator. For soft manipulator without torsional DoF, we could increase $J$ to limit its torsion. This won't increase the difficulty of controlling the manipulator and won't decrease the flexibility.

The length of soft manipulators $L$ is determined by application requirements and could not be decreased arbitrarily.
 
In summary, increasing oblique constraint or actuation and increasing torsion constant $J$ can increase critical load without sacrificing the flexibility.

\subsubsection{Summary for soft arm design.}
According to the analysis above, we can derive several design principles for soft arm. First of all, the ability to output longitude and torsional force together determines the load limit and stability of the manipulator. Increasing the radial constraint could reduce the buckling of compressed part of the manipulator. Secondly, increasing the ability to output torsional force could reduce flexural-torsional buckling of the manipulator. And both of these won't decrease the flexibility of the manipulator. According to this principle, a soft arm with large flexibility and load capacity could be designed.

\subsection{Honeycomb Pneumatic Network Architecture}

\begin{figure}[htbp]
\centering
\includegraphics[width=\columnwidth]{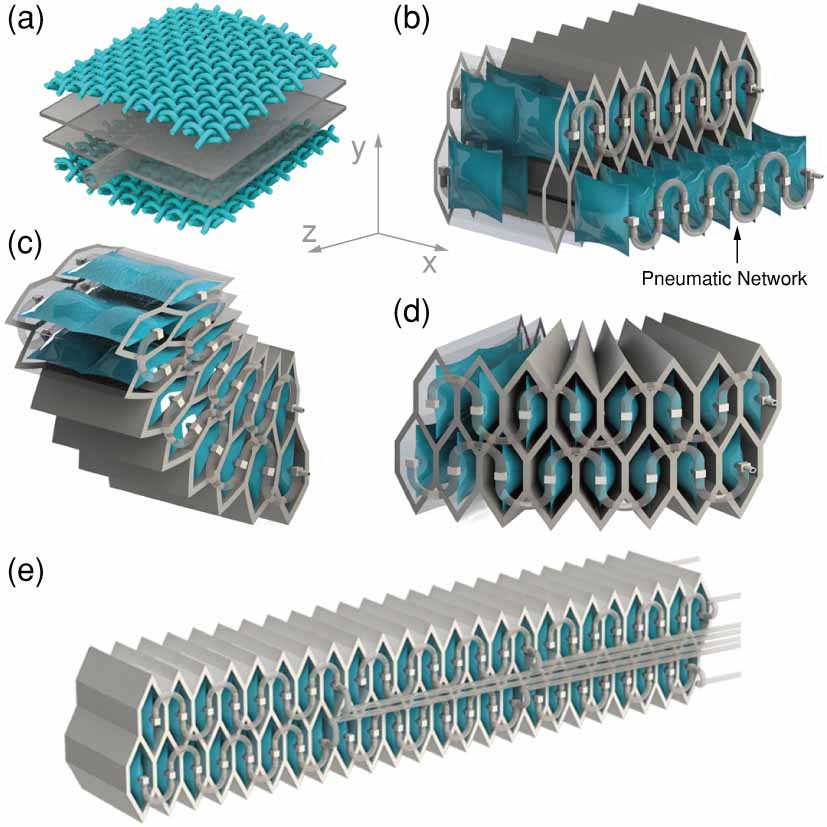}
\caption{Structure of HPN architecture. The segment (b) consists of a honeycomb structure, airbags and air tubes. An airbag is detailed in (a). The airbag consists of two layers of TPU film, and it is reinforced by processed nylon fabric. The airbag can be pressurized through one air tube with a maximum pressure of 0.5 Mpa. (c) and (d) show the bending deformations in two directions. (e)shows a HPN Arm composed of three segments}
\label{HPN_architecture}
\end{figure}

Based on the design principles above, we proposed Honeycomb pneumatic networks (HPN), which was first demonstrated in our previous work \citep{sun2016flexible}. HPN structure derives from embedded pneumatic networks (EPN) and has hexagonal honeycomb chambers. HPN is a flexible structure, and its deformation is mainly due to the change of the angel between the honeycomb walls rather than material deformation, which make mechanical efficiency much higher. In this work, soft arms are designed based on HPN. As shown in Figure \ref{HPN_architecture}, there are two columns of closely spaced honeycomb units with pneumatic networks in them. By inflating different pneumatic network separately, the HPN structure can achieve diverse bend and elongation. An HPN Arm can be composed of multiple segments to be more flexible. The Figure \ref{HPN_architecture} shows a HPN Arm composed of three segments. In order to obtain better bending and elongation performance, a compressed honeycomb structure is adopted by HPN Arm, which has greater deformation capacity and can reduce power consumption.

According to the design principle in the previous section, both the stretched and compressed part of the HPN Arm are honeycomb structure. This integrated design prevents the compressed part from buckling. The honeycomb structure has relative high Shear modulus \citep{wahl2012shear}, and can resist axial torsion so that flexural-torsional buckling of the whole arm will not occur. The pneumatic network provides axial driving force, enabling HPN to have greater elongation and bending capacity. Therefore, the design of our HPN Arm has great potential both in flexibility and load capacity, which has been proved by subsequent experiments. At present, our HPN Arm can carry a 3 kg load with 4 kg self weight.

\subsection{Fabrication}

The HPN Arm consists of two parts: the honeycomb structure and pneumatic networks. This paper presents two methods of fabrication honeycomb structure: aluminum sheet method and 3D printing technology. As for pneumatic networks which contain a mass of air chambers arranged regularly, this paper also presents two fabrication methods: handmade pneumatic network and fiber reinforced pneumatic network. After being fabricated separately, the honeycomb structure and pneumatic networks are assembled together.

\begin{figure*}[htbp]
\centering
\includegraphics[width=\textwidth]{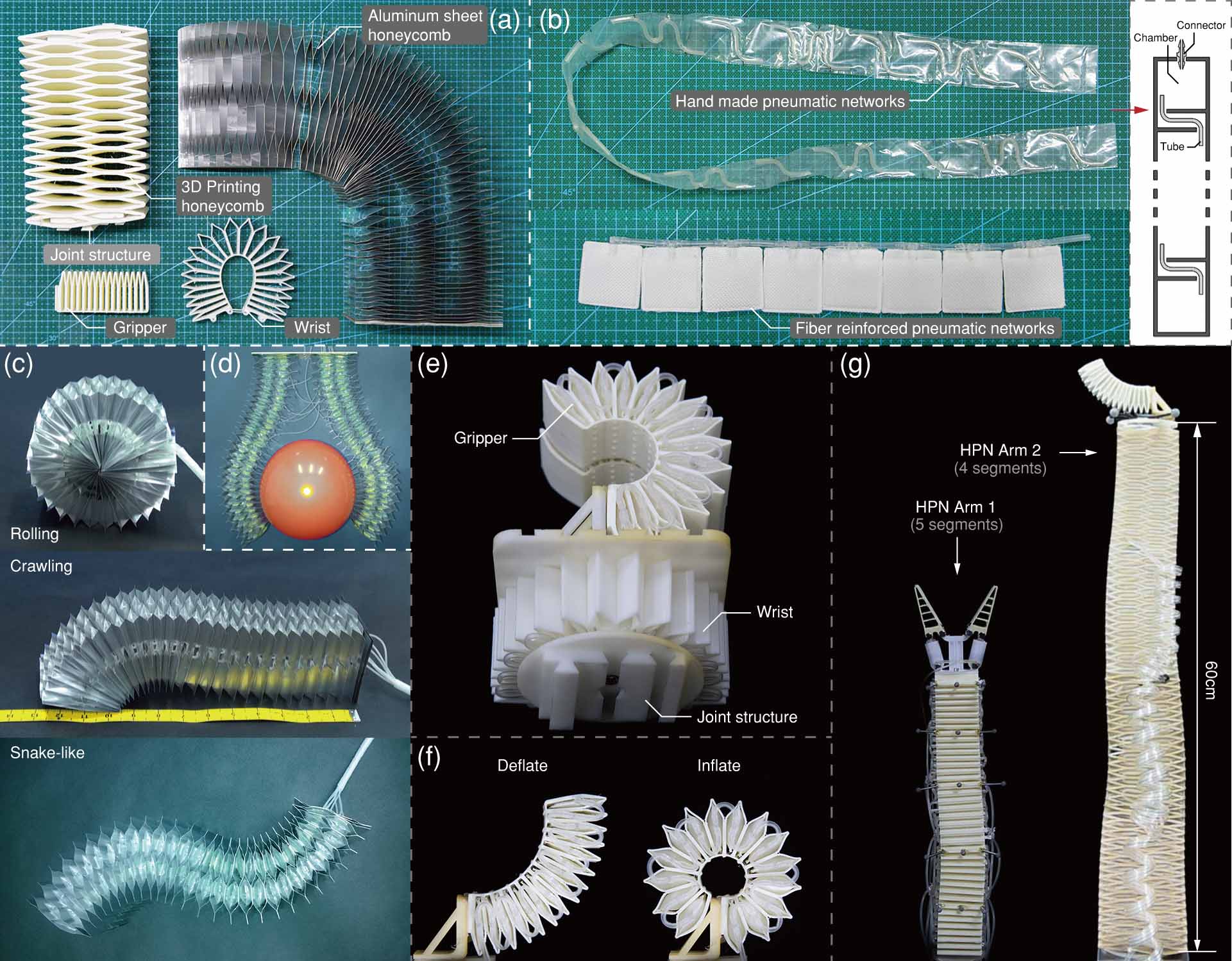}
\caption{(a) shows the honeycomb structure -- aluminum sheet honeycomb structure, 3D printed honeycomb structure, modified honeycomb structure gripper, and wrist. (b) shows the pneumatic networks -- handmade pneumatic networks and fiber reinforced pneumatic networks. Meanwhile, the structure diagram of handmade pneumatic networks, including chambers, tubes, and connector, is shown. The black area is the sealing line of the heat sealing machine. (c) Mobile robot composed of aluminum sheet honeycomb structure and handmade pneumatic networks, which can achieve curling, crawling and snake-like motion. (d)Gripper made of aluminum sheet HPN is grabbing the balloon.(e)Modified HPN gripper and wrist fabricated by 3D printing, which can be attached to end of the arm by its joint. (f) shows the working principle of HPN gripper: it will curl toward the inextensible side after pressurized. (g) HPN Arm 1 and HPN Arm 2 fabricated by 3D printing. Arm 1 is made up of five segments and Arm 2 is made up of four segments, both of which have soft gripper at their end. The length of Arm 2 at work is 60 cm and it can reach to 80 cm after elongation.
}
 \label{fabrication}
\end{figure*}

\subsubsection{Honeycomb structure.}
There are many methods to fabricate honeycomb structure, such as silicone injection molding, 3D printing, and mechanical structure connection. Here we present two fabrication methods: aluminum sheet method and 3D printing technology. Figure \ref{fabrication} shows the honeycomb structure made by aluminum sheets. We can choose some parameters of the honeycomb structure such as the thickness of aluminum sheets and size of chambers. Standard parts can be ordered after parameters are decided. The advantages of this aluminum honeycomb are lightweight, softness and high elongation (up to 400\%). What’s more, we can customize it according to our design or order standard parts directly from the factory, which means it can be mass-produced easily. However, the aluminum sheet is prone to plastic deformation and has small resilience, which limits the performance of HPN Arm under heavy load. This fabrication method is applicable to situations where large load capacity is not required, such as HPN mobile robot Figure \ref{fabrication}(c) and HPN Gripper Figure \ref{fabrication}(d). Another fabrication method we presented is 3D printing. SLA, FDM, SLS and other mature printing methods are used for printing while flexible polymer and photosensitive resin as printing materials. This method is relatively fast and can be used to design honeycomb structures with different properties by changing printing parameters. The soft arms in this paper is fabricated by desktop FDM 3D printer and 1.75mm flexible TPU material. TPU has good features of high flexibility and a large strain-to-failure, and compatible for most desktop FDM/FFF printers. Besides, TPU is not easy to suffer from material aging so it can be used for a long time. The hardness of the honeycomb structure can be changed by changing the filling rate of the printing materials. When the manipulator’s length exceeds the workspace of the printer, we can simply divide the whole structure into several segments and print them separately. To connect different parts, we design a joint structure, shown in Figure \ref{fabrication}(e). The joint structure enables the modular design: numbers of segments can be assembled with tools (such as wrist (e), gripper (f) or clamps (g)) at the end.

\subsubsection{Pneumatic networks.}
Pneumatic Networks can be made either manually or customized. The process of manual preparation of polyethylene airbags was presented in 2017 Soft Robotics Competitions \citep{toolkitpage}. As shown in Figure \ref{fabrication}(b). Even airbags made by hand can withstand a big working pressure considering the protection effects of honeycomb structure, shown in Figure \ref{protect_test}. We also developed another implementation of pneumatic networks with customized fiber reinforced TPU airbags. As shown in Figure \ref{fabrication}(b), several individual airbags are connected with each other by rubber tubes and connectors. This implementation can improve working pressure up to 0.5Mpa with low air leak rate, and is easy to fabricate and maintain. 

As shown in Figure \ref{fabrication}(c), HPN architecture can be used to create a robot with mobility, or to design soft gripper and wrist with torsion by just slightly modify or use part of the HPN structure Figure \ref{fabrication}(e) (f). All these designs inherit the advantages of HPN architecture in different aspects.

From the scattered discussion in above subsections, We can find several advantages and unique characteristics of the HPN: 

\begin{enumerate}[(a)]

\item Our HPN is characterized by a separated honeycomb structure and a pneumatic network, making the fabrication and maintenance easy. Specifically, like our HPN Arm 2 honeycomb structure, it is easy to design thanks to its extrude structure basically; when design is done, it will be sent to the company which provides 3D printing services of flexible materials; fiber-reinforcement airbag size should be designed based on the structure size and sent to the airbag processing plant for processing; we only need to simply join them together after all parts are taken back.
\item The honeycomb structure is very durable because its deformation relies on its structural deformation rather than the expansive deformation of common flexible materials. We haven’t tested it though, but seldom have there been problems with the structure during the use. If the airbag of the pneumatic network has air leakage, we only need to change the airbag. 
\item We can change the entire flexibility by setting the  filling rate of 3D printing. The range we've tried is from 5\% to 100\%. %Yet, the softer, the smaller the resilience force of the same structure. To achieve the same carrying capacity, the intermediate shaft should be thicker. 
\item The cross-section utilization rate of our design is higher. The four square airbags basically fill the cross-section. To achieve the same load capacity, the cross-section of our design is much smaller. It means the arm will be thinner, which can make the arm more adaptable to the restricted environment.

\item It should be wise that we find ways to restrict the unnecessary DoF according to demand if force output capability is needed.
\end{enumerate}

\subsection{Simulation optimization}
In this section, we first introduce two abstract model to calculate flexibility and load capacity. Wall thickness and groove depth are introduced as two key variables. Then, a series of simulation experiments are conducted to explore the relationship between the two variables and the performance (flexibility and load capacity). Based on simulation results, optimized parameters are selected for the design of HPN Arm.

\subsubsection{Calculation with abstract model.}

\paragraph{Flexibility calculation method.}
As mentioned above, we mainly explore the reachable space of HPN manipulator in x-z plane. This space has an axis-symmetric shape, so only half of the area will be calculated. The planar area has two boundaries that join together when the manipulator reaches a max bending condition. The outer one is formed by the manipulator that bends upwards after it elongates straightly to maximum length, while the inner one is formed by it directly bending upwards (see Figure  \ref{flexiblility_simulation}). To calculate the area between the two boundaries, we first calculate two areas respectively between the bending arc and outer boundary, $S_1$, and the bending arc and inner boundary, $S_0$. And then minus them.

\begin{figure}[htbp]
\centering
\includegraphics[width=\columnwidth]{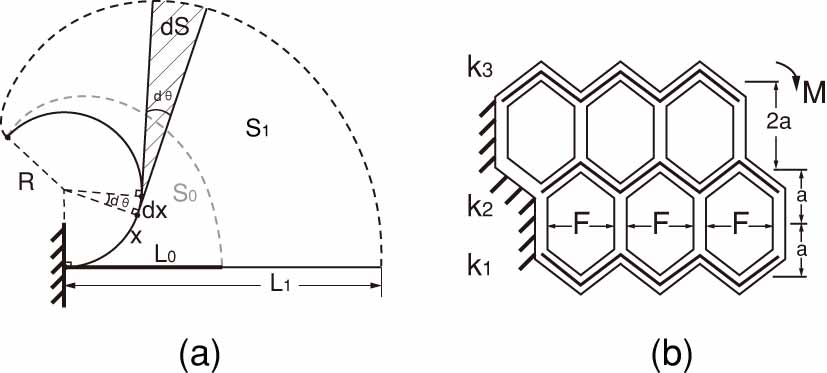}
\caption{(a) illustrates the process of cutting infinitesimal that composes the half of the reachable space. (b) shows a simplified model for calculating the load moment.}
\label{flexibility_load_cal}
\end{figure}

We represent the minimum bending radius, original length, maximum length and width of the manipulator as $R$, $L_0$, $L_1$ and $D$ respectively. To simplify the calculation, we extract two lines that remain constant lengths during the bending and use their tips' trajectories to represent the two boundaries. These two lines lie on the staggered area of the manipulator (see Figure \ref{flexibility_load_cal}(b)), and they are about $\frac{D}{3}$ away from the middle line, whose bending radius is $R$. So these two lines' bending radius $R_0$, $R_1$ are:

\begin{equation}
R_0 = R - \frac{D}{3}, R_1 = R + \frac{D}{3}
\end{equation}

To calculate the areas respectively, we separate the process into infinitesimals $dS$ with similar sector shapes (the only difference between $dS_0$ and $dS_1$ is the bending radius $R$ and length $L$): (see Figure \ref{flexibility_load_cal})

\begin{equation}
d\theta_i = \frac{dX_i}{R_i}, dS_i = \frac{1}{2}(L_i - X_i)^2d\theta_i~(i = 0, 1)
\end{equation}

Then calculate the integral:

\begin{equation}
S_i = \int^{\frac{L_i}{R_i}}_0 \frac{1}{2}(L_i - X_i)^2d\theta_i = \frac{L_i^3}{6R_i}~(i = 0, 1)
\end{equation}

Minus $S_1$ by $S_0$, and we get half of the reachable space of the manipulator. Then we divide the whole reachable space by the square of the side length to achieve an dimensionless number $f$ to represent the relative flexibility:

\begin{equation}
f = \frac{2(S_1 - S_0)}{L_0^2} = \frac{1}{3L_0^2}\left( \frac{L_1^3}{R + \frac{D}{3}} - \frac{L_0^3}{R - \frac{D}{3}}\right)
\label{space_cal}
\end{equation}

We need to mention that if the manipulator's length is relatively large, its tip can reach almost every point inside the outer boundary, so the area of inner unreachable space can be neglected.

\paragraph{Load capacity calculation method.}
There are several ways to achieve the condition that the tip and fixed end of the manipulator are on the same height under a load. We choose a relative strict condition: all parts of this manipulator must be at the same height, rather than the tip and the fixed end. Actually, the manipulator can get a better load moment with a bending shape. 

We introduce a simplified model to analyze the load moment of the HPN manipulator in this condition. In this model, we only consider the stretching deformation, because the shear deformation, which is constrained by the structure, is much smaller than stretching deformation. As shown in Figure~\ref{flexibility_load_cal}, we assume that the units are hexagons with vertical edges at a length of $2a$, and the staggered parts (with broken-lines inside) are simplified as springs. $F$ represents the internal force provided by the airbags, which is determined by the contact area $S$ and pressure $p$. $k_1$, $k_2$, $k_3$ represent the spring coefficients. $M$ represents the load moment provided by this structure. To reach a balance, we assume a elongation of $\Delta s$ for the springs. So, the force balance can be represented as:
\begin{equation}
F=(k_1+k_2+k_3)\Delta s
\end{equation}
While the load moment can be calculated as:
\begin{equation}
\label{load_result}
M=Fa-2ak_1\Delta s+2ak_3\Delta s=Spa\left[1+\frac{2(k_3-k_1)}{k_1+k_2+k_3} \right]
\end{equation}

From Equation (\ref{load_result}), we can draw the conclusions:
\begin{itemize}
\item If the manipulator has symmetrical structure ($k_1$ = $k_3$), its load moment is $Spa$. So possible ways to improve load moment are increasing pressure, edge length and contact area between airbags and walls.
%\item If the manipulator has complete different sides ($k_3 >> k_1$), its maximum load moment is $3pSa$, so additional parts in design increasing $k_3$ is useful to increase load moment.
\item If the manipulator has asymmetrical structure ($k_3 > k_1$), its load moment can be larger than $Spa$, so increasing ($k_3 - k_1$) is useful to increase load moment. In theory, the manipulator's load moment can reach a limit of $3Spa$ when $k_3 \rightarrow \infty$.
\end{itemize}

% \subsection{HPN arm design}
\subsubsection{Analysis based on nonlinear FEM.}
After using honeycomb network, its strong ability to resist torsion enables us not to worry about the flexural torsional buckling. Then we can focus on analyzing flexibility and theoretical load capability.

For the multi-segment arm, the design parameters of each arm should be different in order to achieve decent flexibility and load capacity.
% 比如,考虑到根部段要承受负载和前面所有段自重带来的力矩,根部段要设计得更 powerful.我们通过对手臂自重与力输出能力与灵活度的衡量,设计出了满足我们性能要求的手臂.(引文)

In this subsection, we first introduce two key variables, the wall thickness and groove depth. Then, a series of simulation experiments are conducted to explore the relationship between the two variables and the performance (flexibility and load capacity). Specifically, the parts of the manipulator composed of 32 units are created using SolidWorks and imported to Abaqus, where we set Young's modulus as 50, Possion's rate as 0.38 and density as \begin{math} 1.2\times10^{-6} kg/{mm}^{3} \end{math} for the material of the frame. The wall thickness of the manipulator varies from 2mm to 4.5mm (0.5mm a step) and groove depth varies from 0mm to 6mm (1mm a step), and thus we get 42 groups of data. For simplification, we keep the shape and scale of the cavity as constants during simulation experiments.

\begin{figure}[htbp]
 \centering
 \includegraphics[width=\columnwidth]{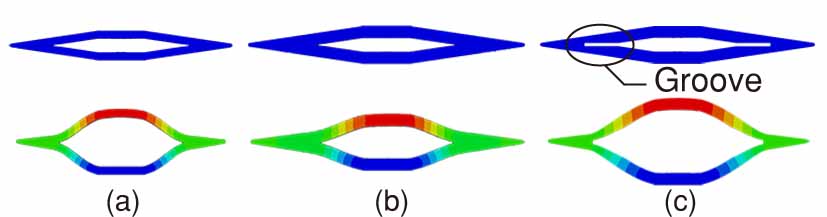}
 \caption{Deformation of HPN units with different wall thickness and groove depth under the pressure of 0.1Mpa. (a) is an HPN unit of a 2mm wall thickness and no groove; (b) is of a 3mm wall thickness and no groove; (c) is of a 3mm wall thickness and 6mm groove depth.}
 \label{Comparison}
 \end{figure}

\paragraph{Key variables.}
There are many variables affecting the manipulator's features, such as its length, width, wall thickness, etc. Using the finite element method, we analyze their impacts respectively. After several tests, we find that the wall thickness is a key variable. The deformation of three HPN units under the same pressure (100Kpa) is shown in Figure \ref{Comparison}. With 2mm wall thickness, the cavity height of unit in Figure \ref{Comparison}(a) increases from 3.5mm to 12.6mm. As the wall thickness becomes 3mm in Figure \ref{Comparison}(b), its cavity height increases from 3.5mm to 8.2mm. Since the stability of the structure can be improved by increasing the wall thickness, its flexibility is drastically lowered. So we open grooves (Figure \ref{Comparison}(c)) at the inner intersections of acute angles to mitigate the contradiction. The unit in Figure \ref{Comparison}(c) is the same with that in Figure \ref{Comparison}(b) except for grooves of 3mm depth. Its cavity height deforms from 3.5mm to 16.1mm, which is much larger than its counterpart. In further simulation experiments, we treat groove depth as another significant variable.

\paragraph{Simulation of flexibility.}
We then conduct simulation experiments on difference of wall thickness and groove depth, respectively.

\begin{figure}[htbp]
 \centering
 \includegraphics[width=\columnwidth]{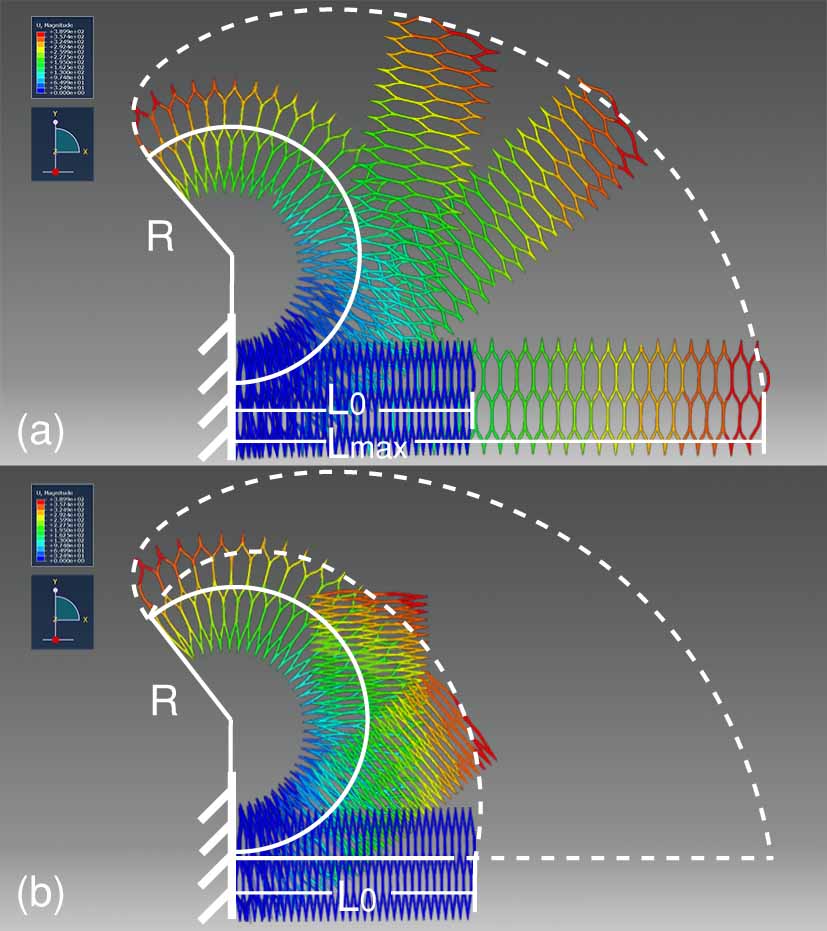}
 \caption{Bending trajectories and boundaries. In (a), the manipulator is firstly elongated from original length $L_0$ to maximum length $L_1$ by inflating all the airbags with maximum pressure (100Kpa), and then deflated from the fixed end to tip on the upper side to bend upwards during which the outer boundary (white dotted curve) is recorded. When all the airbags on the upper side are deflated, the minimum radius $R$ is recorded. Then, in (b), the manipulator is gradually deflated from tip to the fixed end and forms an inner boundary (the shorter white dotted curve).}
 \label{flexiblility_simulation}
 \end{figure}

\begin{figure}[htbp]
\centering
\includegraphics[width=\columnwidth]{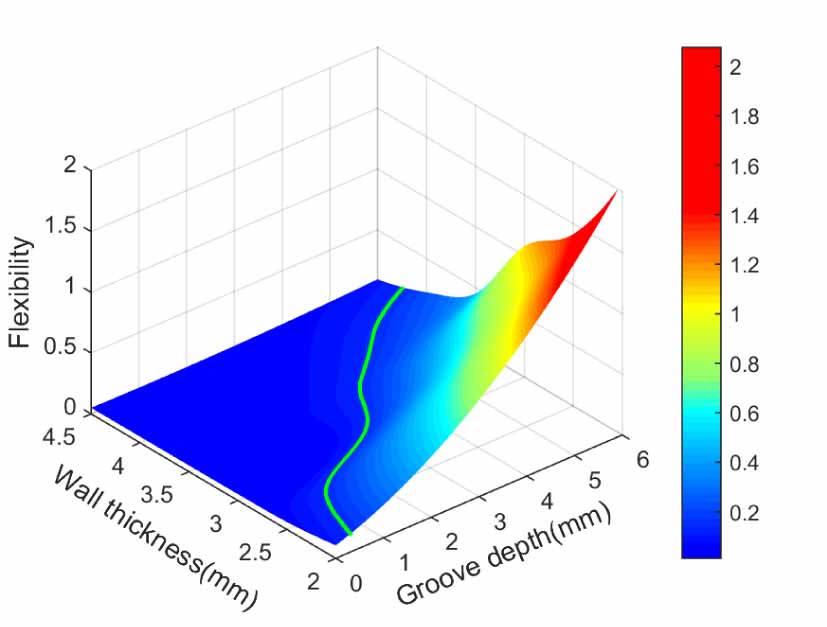}
\caption{The relationship between wall thickness, groove depth and flexibility. The solid line consists of points whose flexibility equals 0.15.}
\label{3D_simulation_flexibility}
\end{figure}

Figure \ref{flexiblility_simulation} describes a simulation process of flexibility. $L_0$, $L_1$ and $R$ represent the original length, maximum length and bending radius respectively. Half of the manipulator's reachable space is surrounded by the white dotted lines in Figure \ref{flexiblility_simulation}. We measured that area using Monte Carlo method. Furthermore, using $L_0$, $L_1$, $R$ and $D$, the reachable space can be calculated by equation \ref{space_cal}. The calculated area approximately equals to the measured counterpart, which validates the calculation method. So we only need to measure $L_1$, $L_0$ and $R$ to simplify further simulation experiments. The relationship between 42 groups of the features (wall thickness, groove depth) and the corresponding calculated flexibility is shown in Figure \ref{3D_simulation_flexibility}.

\begin{figure}[htbp]
\centering
\includegraphics[width=\columnwidth]{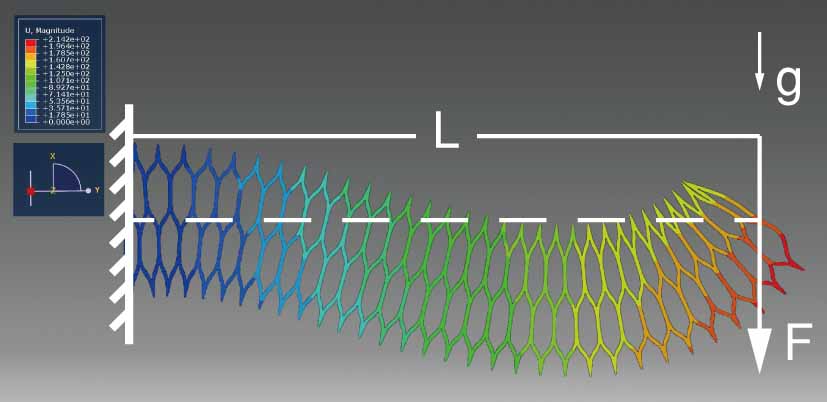}
\caption{Load capacity when bend is allowed. All the airbags are inflated on the lower side and its load increases until the two ends reach the same height. \textit{F} represents the external load, \textit{L} represents the horizontal distance between the hanging point of the load to the fixed end. }
\label{simulation_payload}
\end{figure}

\paragraph{Simulation of load capacity.}
As mentioned in the calculation method of load capacity, we measure the load moment of the HPN manipulator when it lies exactly on the horizontal line. In this condition, the load moment is only determined by its structure's parameters (such as wall thickness and groove depth). Otherwise, when the manipulator's loaded shape is allowed to be bending (Figure \ref{simulation_payload}), the load moment will be affected by its original length. So we choose the former method to measure and evaluate the load capacity of structures with different parameters. In fact, this method provides a general guidance for choosing parameters of a manipulator in practical use, where the manipulator can provide larger load capacity than that in experiment.

\begin{figure}[!htbp]
\centering
\includegraphics[width=\columnwidth]{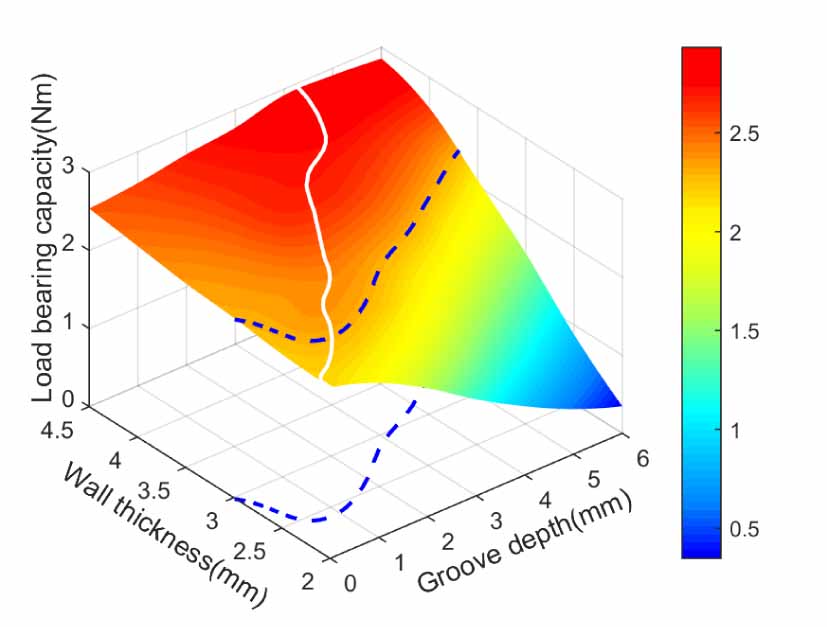}    
\caption{The relationship between wall thickness, groove depth and load capacity. The blue dotted line consists of points whose load  capacity equals 2.3 Nm, while the white solid line consists of points whose load capacity are the maximum with fixed wall thickness respectively.}
\label{3D_load_capability}
\end{figure}

The relationship between 42 groups of features (wall thickness and groove depth) and the corresponding load capacity (load moment) is shown in Figure \ref{3D_load_capability}.

The variation of flexibility is illustrated in Figure \ref{3D_simulation_flexibility}. As the groove depth grows, the flexibility increases, while as the wall thickness grows, the flexibility decreases. Both variations are monotonous. As shown in Figure \ref{3D_load_capability}, the load capacity increases monotonously as the wall thickness grows. While the variation of that with the groove depth is not monotonous: it initially increases, and then decreases. We believe the cause is that as the groove depth increases, $L$ increases while $F$ decreases (Figure \ref{simulation_payload}). Because the wall of a relatively large thickness confines the airbags' inflation, F decreases slowly when the groove depth is relatively small. So their product initially increases. Besides, the deformation of airbags is confined, so the increment of $L$ is confined while $F$ always decreases as the groove depth increases. So their product has an extreme value, and then decreases.

Because of the non-monotonous variation, the load capacity reaches an extreme value at a certain groove depth under every fixed wall thickness (white line in Figure \ref{3D_load_capability}). On its left side, the flexibility and load capacity both increase as the groove depth grows. Thus, values in this region are not worth concerning (suitable values are on the right of the white line), which simplifies the design process.

\begin{figure}[thbp]
\centering
\includegraphics[width=\columnwidth]{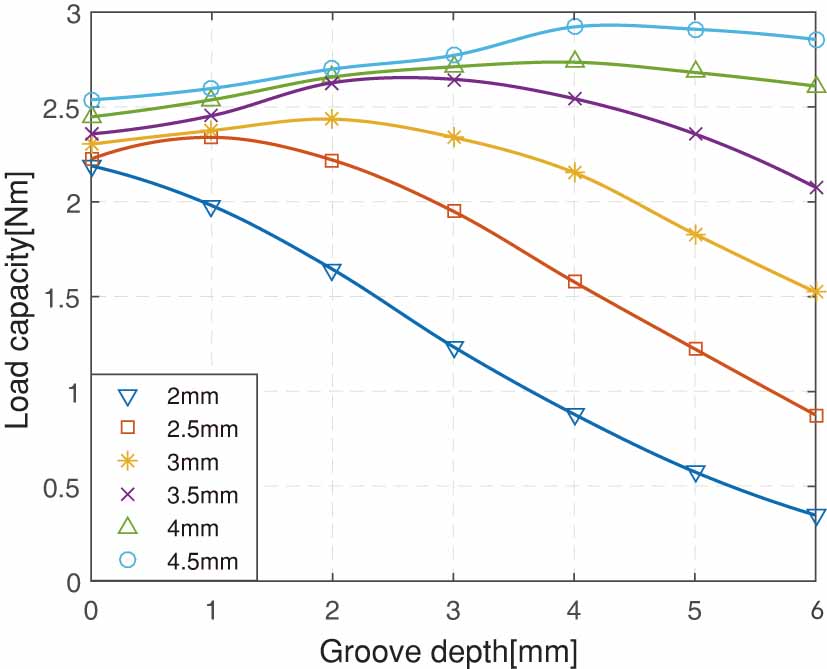}
\caption{The relationship between the load capacity and groove
depth where the wall thickness varies from 2mm to 4.5mm (0.5mm a step).}
\label{2D_load_capability}
\end{figure}

To get an explicit view of the load capacity's non-monotonous variation along with the groove depth, the performance of load capacity under different fixed wall thickness are shown in 2D plane (Figure~\ref{2D_load_capability}). From that, several conclusions can be drawn:
\begin{itemize}
\item As the wall thickness grows, the extreme point moves rightwards, so a deeper groove is required for a thicker wall to achieve maximum load capacity.
\item The curve corresponding to a 2mm wall thickness monotonously declines, with no extreme point. From the trend that the extreme point moves leftwards when wall thickness decreases, we can infer that when the wall thickness is less than a certain value, the extreme point has minus groove depth. Therefore, an additional part at the intersection, instead of a groove, is required to realize a model with a thin wall.
\item With a fixed groove depth, the load capacity always increases as the wall thickness grows, but its rising range gradually reduces. So, it's not effective to keep increasing the wall thickness when it's already large. Instead, it is useful to increase the pressure.
\end{itemize}

\begin{figure}[htbp]
\centering
\includegraphics[width=\columnwidth]{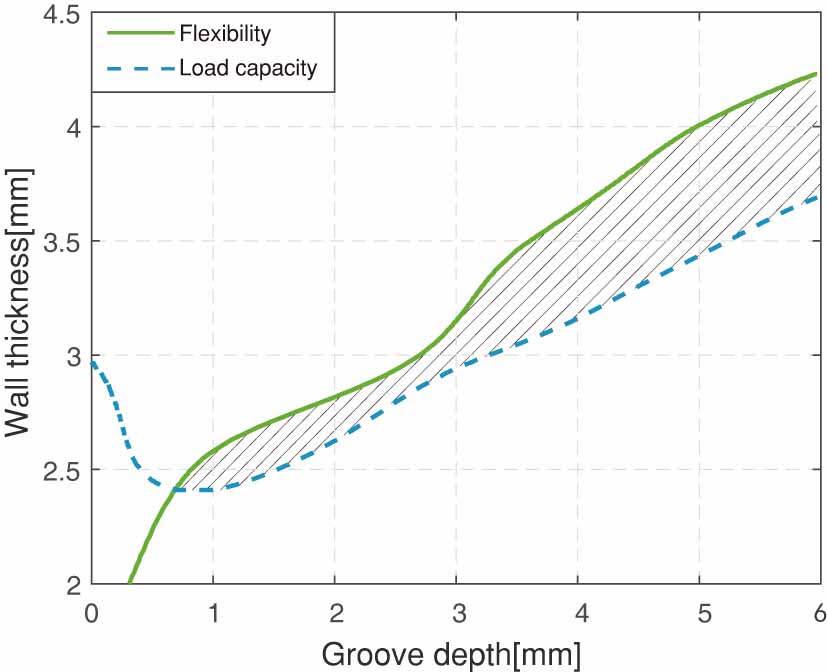}
\caption{The green solid line is the contour of flexibility in Figure~\ref{3D_simulation_flexibility} while the blue dotted line is of load capacity in Figure~\ref{3D_load_capability}. The area between these two lines meets the requirements: with flexibility more than 0.15 while load capacity more than 2.3 Nm.}
\label{simulation_load_flex}
\end{figure}

Furthermore, the simulation results also provide a guidance for determining feasible solutions to certain requirements: If either flexibility or load capacity is required, a contour can be cut from the corresponding curve surface (see the green line in Figure~\ref{3D_simulation_flexibility} or the blue dotted line in Figure~\ref{3D_load_capability}), and then projected onto the other surface to form an available region for this variable, in which the other variable's optimal value can be selected. If both flexibility and load capacity are required, there will be two contours representing the requirements separately on two curve surfaces, which can be projected onto x-y plane. If there is an available intersection formed by the two curves (see the shaded area in Figure~\ref{simulation_load_flex}), qualified solutions can be chosen from that region, otherwise, either of the two requirements needs to make compromise.

\subsection{Power system}
In this section, we present a power system that provides control of working fluid for soft manipulators. We design the control system based on solenoid valves and proportional valves respectively. Because the valve will generate much heat when working, both control systems use fans to cool the whole system to ensure the long-term stable operation.

\begin{figure}[htbp]
\centering
\includegraphics[width=\columnwidth]{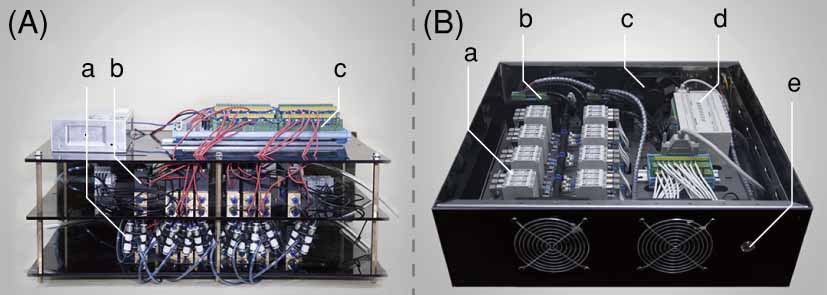}
\caption{Solenoid valve and proportional valve control system. (A)shows solenoid valve control system, which contains check valve(a), solenoid valve(b), and solenoid valve control panel(c). (B) shows proportional valve control system, which contains proportional valve(a), stepper motor driver(b), fan(c), stepper motor controller(d), and switch}
\label{control_box}
\end{figure}

\paragraph{Solenoid valve control system.}
The solenoid valve control system is shown in Figure~\ref{control_box}), including power supplies, solenoid valves, relays, data output card (Advantech PCIE-1751), cooling fans, etc. The solenoid valves are connected to the air source with positive and negative pressure, and then every two solenoid valves connected to a tee connector that can control a pneumatic network connected to it. Soft manipulators can be actuated by controlling the time, frequency and coordination of positive and negative pressure solenoid valves. In this solenoid valve system, in order to prevent gas channeling and ensure the performance of the control system, the output end of each solenoid valve is connected with a check valve and an airbag. The data output card is connected with the motherboard and adopts the bus protocol to ensure the low communication delay rate. The advantage of solenoid valve control system is its relative low cost. But there will be noise during operation, especially at high frequency.

\paragraph{Proportional valve control system.}
The proportional valve system is shown in Figure~\ref{control_box}(b), which mainly includes 24-way proportional valve (SMC ITV0030), data output card (Advantech PCI-1724U), stepper motor controller, stepper motor driver, cooling fan, etc. The proportional valve is used to control the air pressure of the connected pneumatic network. The data output card is connected with the motherboard and adopts the bus protocol to ensure the low communication delay rate. In order to expand the function of this experimental platform, two groups of step motor drivers and controllers were added into the proportional valve control box to control the movement of the base in the 2-D plane. The advantage of this proportional valve control system is that it can easily control the air pressure and achieve closed-loop in actuator space.

\subsection{Performance evaluation}

\begin{table}[!hbp]
\centering
\begin{threeparttable}

\caption{Optimal parameters}\small
\label{optimized_parameters}
\begin{tabular}{cccc}
\toprule
 Segment number\tnote{1}& 1-3& 4& 5\\
\midrule
Load moment $M_L$ (N$\cdot$m)& 0.294& 0.392& 0.490\\
Self moment $M_S$ (Nh$\cdot$m)&0.648& 1.175& 1.919\\
Required moment $M_R$ (N$\cdot$m)& 0.942 & 1.567 &2.409\\
\midrule
Wall thickness $w$ (mm)& 2 & 2.5 & 3.5\\
Groove depth $d$ (mm)& 3 & 4 & 4\\

Selected moment  $M_A$ (N$\cdot$m)& 1.236 &1.578 & 2.544\\
\bottomrule
\end{tabular}
\begin{tablenotes}
\footnotesize
    \item[1] The length of one segment is about 5cm when free and 10cm when fully inflated. Segments 1-3 are the same. 
%    \item[2] Notice that the 5th segment isn't able to afford a load of 2.899N\boldmath$\cdot$m. 20\% shortage is allowed in simulation.
    \end{tablenotes}
\end{threeparttable}

\end{table}
% \label{arm_parameter_table}

The simulation results and analysis provide guidance for designing a prototype. To simplify our design, we separate the prototype into five segments, and the wall thickness and groove depth of HPN units are the same in each segment. As the segment is closer to the root, the load capacity is expected to be stronger for holding all the segments in front. The prototype is expected to have a load capacity of 100 grams at an elongation of 50 cm. The parameters selected for each segment are shown in Table.~\ref{optimized_parameters}, and it should be mentioned that the choices listed in the table are determined according to the importance sequence as: load capacity $>$ flexibility $>$ wall thickness (the manipulator is expected to be lightweight).

A series of experiments are conducted to test the performance of our prototype.
\subsubsection{Protection test.}

\begin{figure}[htbp]
\centering
\includegraphics[width=\columnwidth]{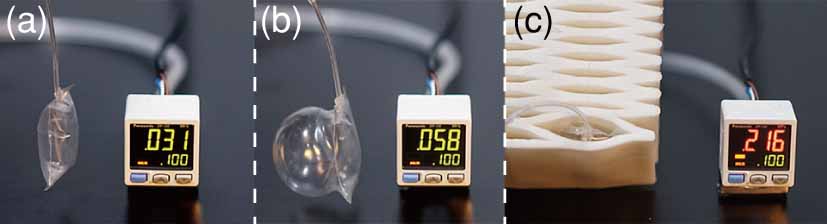}
\caption{Protection test. (a) shows a normal airbag. In (b), the airbag makes an irreversible deformation at a pressure of about 0.06 Mpa without frames while it reaches 0.216 Mpa with frames outside in (c), over 3 times of that in (b).}
\label{protect_test}
\end{figure}

As mentioned in the section of structure design, we experimentally validate the protection effect of the HPN framework. In this experiment, a manually prepared airbag made of PE/AE material is used. As shown in Figure~\ref{protect_test}, an airbag is fragile without protection: it makes irreversible deformation at a lower pressure. In contrast, it withstands a much higher pressure when surrounded by an HPN framework.The honeycomb can protect the airbag. The maximum pressure of the airbag placed in the honeycomb is more than 4 times of that in freespace.

\subsubsection{Experiments of Arm 1 flexibility.} 

\begin{figure}[htbp]
\centering
\includegraphics[width=\columnwidth]{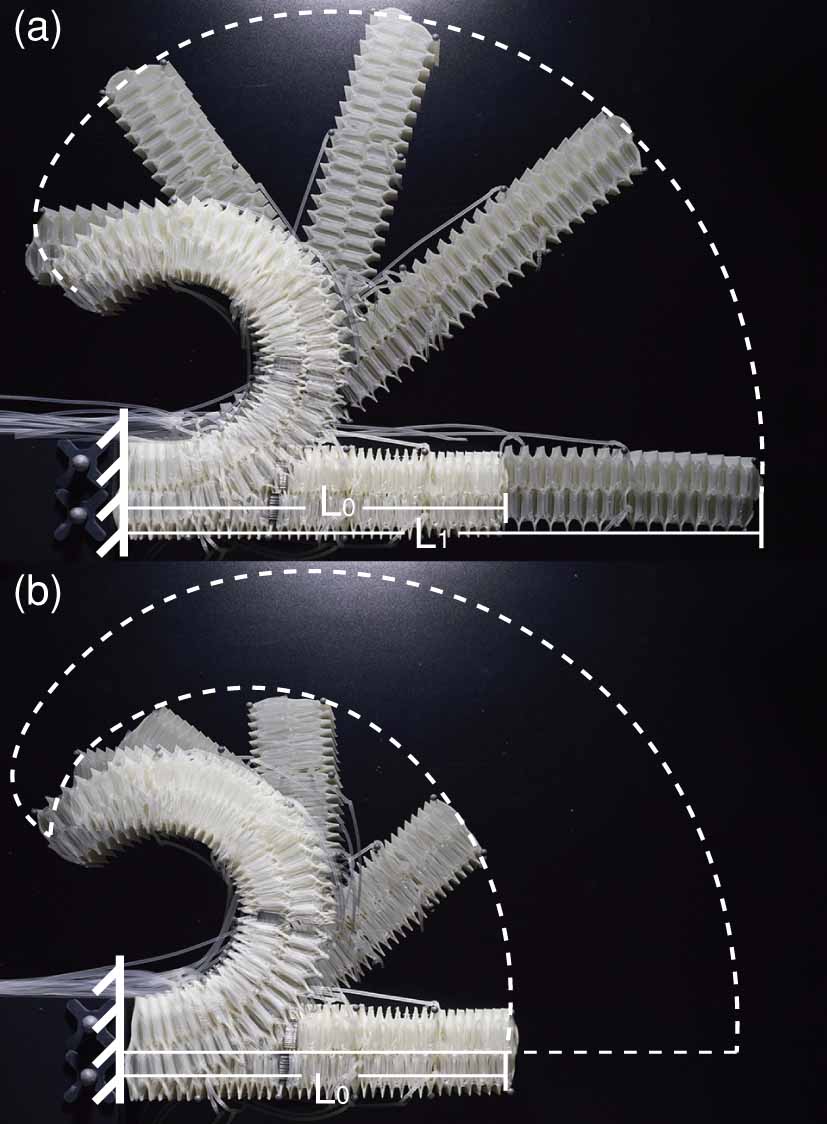}
\caption{Reachable space. In (a), firstly airbags on both sides are fully inflated to get a maximum length, then airbags on the upper side are gradually deflated from the fixed end to the tip, during which the trajectory are recorded as outer boundary. In (b), airbags are deflated from the tip to the fixed end, during which we record the trajectory as inner boundary.}
\label{flexiblility_real}
\end{figure}

Figure~\ref{flexiblility_real} shows a process of measuring HPN Arm 1 flexibility. There is a deviation between the manipulator's deformation in simulation and experiment (Figure~\ref{3D_simulation_flexibility} and Figure~\ref{flexiblility_real}). The main reason is that actual airbags have air tubes with thicknesses about 4.7mm, which enlarges the cavity height, thus the manipulator's original length is lengthened and minimum bending radius is enlarged. And the elongation rate decreases because the maximum length isn't affected.

\subsubsection{Experiments of Arm 1 Load capacity.}

To increase the load capacity, we design two additional segments as the root, whose main requirement is the load capacity. As equation (\ref{load_result}) shows, the load moment increases as ($k_3 - k_1$) increases, so the root segment is designed with asymmetrical grooves that only exist on the lower side. Therefore, the root part is capable of providing additional carrying capacity besides lifting the five segments above the horizontal line.

\begin{figure}[htbp]
\centering
\includegraphics[width=0.95\columnwidth]{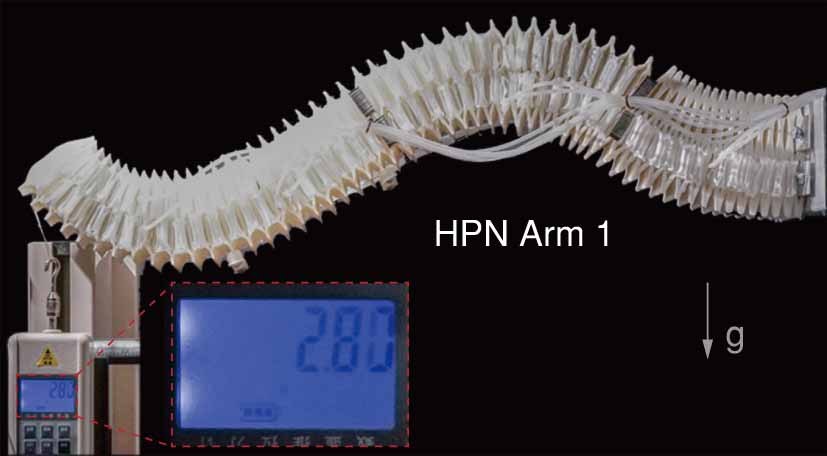}
\caption{Experimental performance of load bearing capacity. With a horizontal distance of 63cm between the hanging point and the fixed end, the HPN manipulator exhibits a load of 2.80N.}
\label{capacity_show1}
\end{figure}

Under the pressure of 90Kpa, the manipulator exhibits a payload of 2.80N at a length of 63cm (see Figure~\ref{capacity_show1}), which fully satisfies the requirements. As mentioned in simulation, the manipulator exhibits a complex bending shape instead of being straightly elongated under a load. The manipulator can withstand a heavier load in practical environments, because the manipulator usually winds to grasp or hold an object, which reduces the real load and gravity moment.

The HPN manipulator's performance shows great flexibility, load bearing capacity and cooperation ability with human (see Fig.~\ref{Low_level_app}). (first line) shows a process of grasping a ball by the HPN manipulator where it performs high compliance and flexibility; the manipulator exhibits high degrees of freedom as well as stability during movements in a fetching process in (second line and third line) demonstrates a process of fetching a hammer and passing it to a man by the HPN manipulator, which shows proper load bearing capacity and cooperation ability.

\subsubsection{Performance evaluation of Arm 2}
In the above simulation analysis and performance evaluation experiments, we only take the wall thickness and groove depth of the honeycomb structure as variables to study. In order to obtain a more powerful soft arm, we can improve the performance of the soft arm by optimizing other parameters of the arm in the design. On the basis of HPN Arm 1, we design and fabricate HPN Arm 2 for application scenarios with stronger output capacity requirements in 3D space. HPN Arm 2 is composed of four sections of honeycomb, with an overall shape of prism. The size of cells of honeycomb structure and airbags of pneumatic networks gradually increased from tip to root, thus the load capacity increases according to the analysis in the Section 1.4.1. In addition, according to Equation (\ref{load_result}), increasing K1, K2 and K3 limits the extension of the arm under specific working pressure, which can maintain a large value of effective contact area between the airbag and the honeycomb structure, thus increasing the load capacity. However, increasing K1, K2, and K3 at the same time will affect flexibility. We can also find that only increasing K2 can improve the load capacity with almost no loss of flexibility from Equation (\ref{load_result}). Considering the bending of the arm in the other direction of the 3D space, we only increase the K2 of the geometric center of Arm 2 by choosing proper wall thickness and groove depth, so as to obtain greater load capacity and better stability, which are very important performances in the 3D space.

\begin{figure*}[htbp]
\centering
\includegraphics[width=0.95\textwidth]{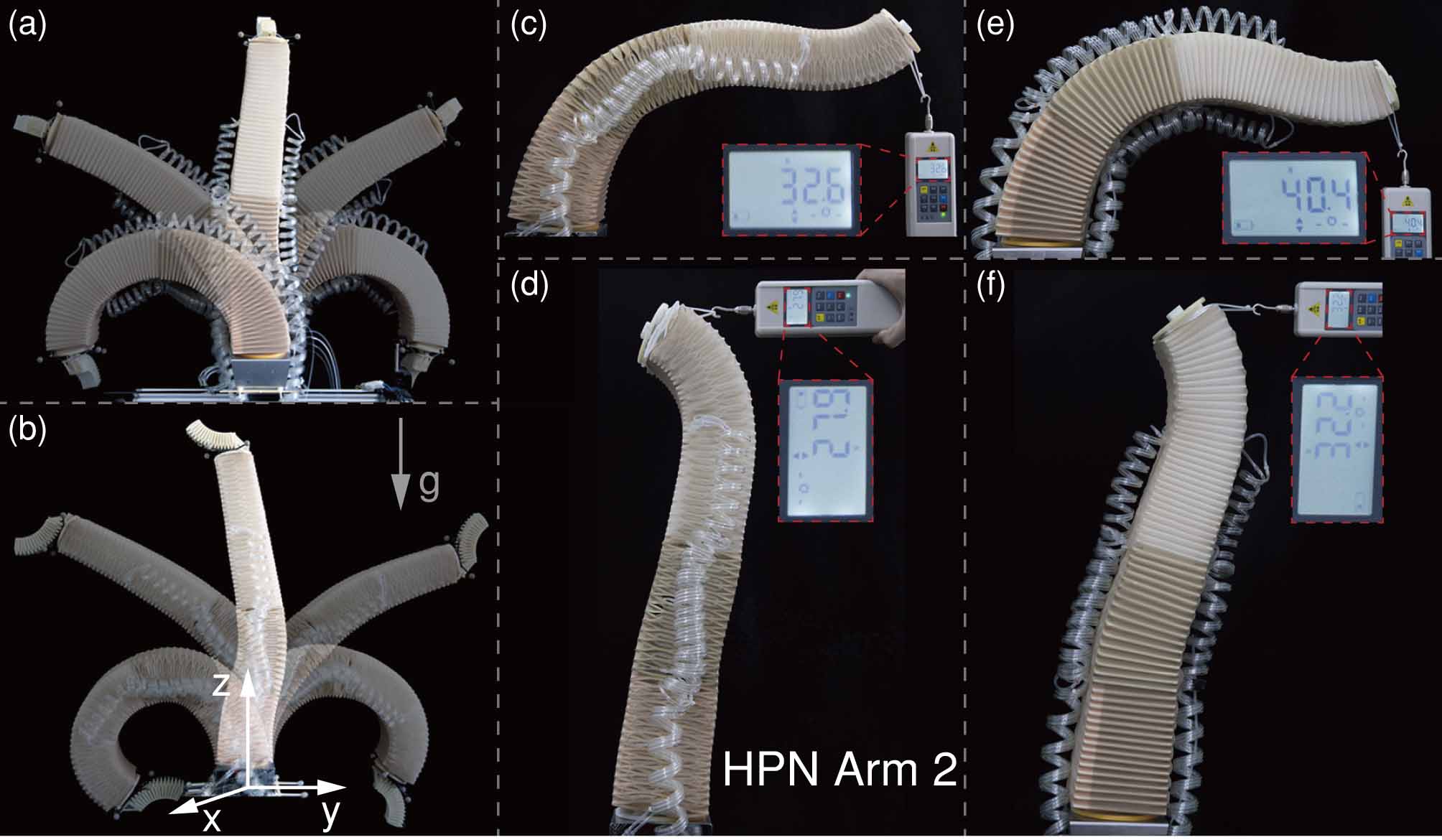}
\caption{Experimental performance of HPN Arm 2. (a) shows the workspace of Arm 2 in the x-z plane. (b) shows the workspace of Arm 2 in the y-z plane. When arm is placed upright HPN Arm 2 exhibit 32.6N and 40.4N load capacity horizontally (c)(e). Exhibit 27.9N and 32.2N load capacity vertically (d)(f).}
\label{load_and_flex}
\end{figure*}

Figure~\ref{load_and_flex} shows the flexibility and load capability of the Arm 2 placed vertically up. The honeycomb structure of the HPN arm is not isotropic. Figure~\ref{load_and_flex}(a)(b) show the workspace of Arm 2 in the x-z plane and y-z plane respectively. Figure~\ref{load_and_flex}(c)-(f) show the maximum output of the tip when the arm posture in two different directions is vertical and horizontal respectively. HPN Arm 2 can carry a 3 kg load with 4 kg self-weight and high flexibility, which provides the hardware foundation for the subsequent application in 3D space.

% \begin{tabular}{cccc}
% \hline
%   & OctArm VI& HPN Arm 1&HPN Arm\\
% \hline
% Length(m)& 1.127& 0.630& 0.6\\
% Load(N)& 8.89 & 2.8 &30\\
% Load moment(Mm)&10.019&1.764&18\\
% Weight(Kg)&6.94&1.5&4\\
% Pressure(bar)&5.51&0.9&2.5\\
% Lm/P&1.818&1.96&7.2\\
% \hline
% \end{tabular}

% 此外我们设计了另外一个四段手臂,负载3Kg, as shown in Figure \ref{capacity_show2}.

%To realize larger reachable space as well as load capacity, we add two segments with thicker walls as the manipulator's root, which is capable of lifting the five segments a front above the horizontal line and provide additional carrying capacity. Under the pressure of 90Kpa, the manipulator exhibits a payload of 2.80N at the maximum elongation, a length of 66cm (see Figure~\ref{experiment_load}). As mentioned in simulation, the manipulator exhibits a complicated shape instead of being exactly straightly elongated under a load. 
%of two reasons: the structure can protect airbags from excessive inflation, thus higher pressure and greater force are available;

% In comparison with OctArm \uppercase\expandafter{\romannumeral6}, the HPN manipulator has a less load moment. However, considering its load moment per unit pressure, HPN manipulator's performance is better than OctArm \uppercase\expandafter{\romannumeral6} (see Table.~\uppercase\expandafter{\romannumeral3}).

\subsubsection{Application demo.}
\begin{figure*}[htbp]
\centering
\includegraphics[width=0.95\textwidth]{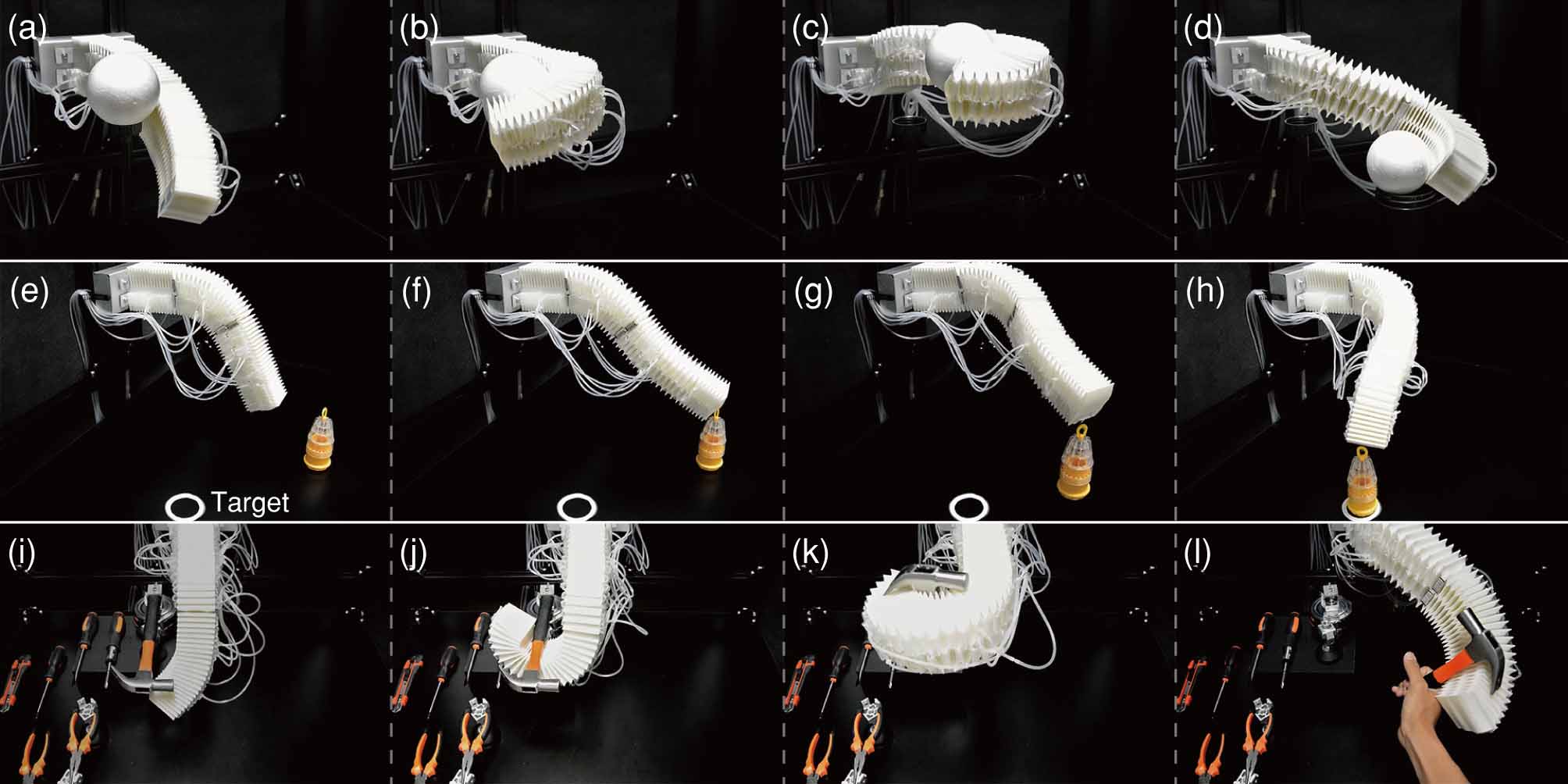}
\caption{Demonstration of HPN Arm 1 applications by controlling actuation sequence.}
\label{Low_level_app}
\end{figure*}

Here several sequences of valves' behavior are set manually for predetermined tasks to demonstrate the potential applications of HPN Arm 1, such as pick and place a ball show in Figure~\ref{Low_level_app} (see also Extension 1).

%\subsection{Design disscussion}
% \begin{enumerate}
% \item 如果需要力输出能力,根据需求把不需要的自由度想办法约束住应该是明智的.

% \item 我们设计的hpn的蜂巢结构和气动网络是分离的,使得制备简单、维护容易.具体来说,像我们的 HpnArm3 的蜂巢结构,由于基本是拉伸结构,很容易设计；设计好之后发给提供柔性材料打印服务的公司打印；fiber-reinforcement 气囊,根据结构尺寸设计气囊尺寸,发给气囊加工工厂加工；所有东西拿回来之后只需要简单的拼接就可以了.蜂巢结构非常耐用,虽然我们并没有做测试,但使用过程中结构从来没有出过任何问题；气动网络中的气囊会有漏气的问题,发现漏气,只需要更换漏气气囊就可以了.
% 可以通过设置3d打印的填充度(我们尝试过5\%-100\%)改变整体的柔软度.不过越柔软,相同结构提供的回弹力越小,要想达到相同的负载能力,中间轴就要设计的更粗.我们这种设计的截面利用率比较高,四个方形气囊基本填满了截面.在达到相同负载的情况下,这种设计的截面积比较小,也就是手臂会比较细,这样在一些受限空间工作时表现会更好.

% \end{enumerate}

\subsection{Design Conclusion}

This section has pointed out design principles of designing soft arms with larger load capacity and provided a design architecture meeting the design principles—HPN and simulating optimization methods; HPN structure and the pneumatic network are independent and separable, which enables fast and efficient fabrication and low-cost maintenance; the final experiment has verified the effectiveness of the above principles and methods. Although this paper is about the design of soft arms, the thoughts, principles, and methods are also applicable to the design of other soft robots.

Though the proposed design principles are rough, they provide a rational basis for the design of the troublesome soft robots and are expected to change the current situation that they are designed mainly based on bionics or intuition, and to produce more and better design principles, theories and specific design schemes and methods guided by the theory. 

Design thoughts of soft robots are different from that of rigid robots. Design of rigid robots can be seen as a process of increasing degrees of freedom (DoFs) to rigid links, while soft robot designs are adding constraints to arbitrary flexible structures, like inflated chambers tend to expand in every direction and balloons, and decrease DoFs. Soft robot designs need to handle unused DoFs to unleash their mechanical property.

Only when we deeply recognize the difference of design ideas for soft robots can we get rid of the shackles from the design ideas of the relatively mature rigid robots and make clear which theories, technologies, and methods are no longer applicable to soft robots and which can be still used. So that we can better develop soft robots. 

\subsubsection{Design limitation.}
% \begin{enumerate}
% \item 我们hpn手臂负载大,只在一定程度上证明了理论的有效性,还很不全面.未来会做进一步的实验验证工作,定量的研究各个约束的对于提升负载能力的作用.
% 本文设计原则部分只探讨了三种约束与负载之间的关系,与灵活度之间的关系没有研究.比如,径向约束不仅影响负载能力,还影响灵活度——对弯曲程度的影响非常大.
% \item HPN由于限制了扭转,手臂整体没有扭转自由度,在特定情况下(详见Application XX章)会有问题.通过加一个可扭转的手腕可以解决这一问题.手腕可以是被动自由度或主动自由度(如图XX).
% \item 目前3d打印结构的手臂的导管是裸露在外面的,当线散乱的时候会影响手臂运动.做控制的时候如果在中间段加标记点也会有影响.不过导管在外面,一旦出了气囊或导管漏气的问题,容易排查,在原理样机阶段,这样是比较方便的.
% \item 分离式的设计带来便利的同时也存在问题：气囊容易穿出来.目前在手臂外面套了一层弹性布料(丝袜),完美的解决了这一问题.
% \item 我们的手臂设计并不对称,也就是说两个方向的灵活度和负载能力并不相同.蜂巢抗剪切的方向(可以给个图,或者引前面图的x、y、z方向)的负载能力更强,灵活度较差,另一个方向反之.本文中只讨论了负载较弱的方向,不过这样设计出来的手臂负载能力肯定是可以满足需求的,但灵活度不能保证满足需求.
% \item 我们的手臂是可伸长的,可伸长有优势,可达区域变大,灵活度变大,后面做控制的时候也要容易；但如果不存在轴向上两个方向的驱动,只有一个方向的驱动(像我们、Octarm),相反方向靠结构的回弹力,会极大减小负载能力,而且容易不稳定.我们可以通过改变中间轴的粗细设计以满足不同的需求,比如多段手臂,根部需要负载较大,用较粗的设计参数,末端需要灵活度较大,就可以选择较细的设计参数.
% \item 仿真优化中,负载能力是按照手臂完全水平来算的.这样设计出来的手臂的实际负载能力会大于设计值.换句话说,目前的优化方法并不能根据需求设计出刚刚好的手臂.这块儿需要更好的优化设计方法,拓扑优化是一种 promising approach.【引飞飞的文章】
% \end{enumerate}
There are several notable limitations within the presented work. 

\begin{enumerate}[(a)]

\item In general, the design principles only discuss the relationship between the radial/oblique constraints and the load capacity, without considering their relationship with the flexibility. For example, the radial constraint not only influences the load capacity but also the flexibility. It has a great impact on the degree of bending. 

\item As for our specific design, the arm has no DoF of torsion because HPN restricts the torsion. In some circumstances, there will be problems. A wrist with rotational DoF can solve this problem. 

\item Furthermore, the separated design of the HPN is convenient in preparation and maintenance, but there is also a problem: the airbag easily falls out. We cover the arm with a layer of elastic cloth sheath and perfectly solved this problem. 

\item At present, the air tubes of the arm are bare and reflective markers are added in the middle segments. Sometimes, it will influence the arm’s motion. However, the bare tubes make the troubleshoot easy if there is leakage for the air gas or the air tubes. It is convenient at the principle prototype stage. After the problem of air leakage is solved, the design of putting the air tubes into the arm will be explored. 

%\item In addition, the extensible arm has advantages: wider accessible areas, greater flexibility and allowing easier control. However, if there are only one-direction actuators instead of two-way actuators axially, the arm will depend on the resilience force of the structure in the opposite direction, then the load capacity will be greatly reduced and the stability is poor. We can change the thickness of the intermediate shaft to meet different needs. Such as the multi-segment arm, the root section needs a larger load, so well adopt thicker design parameters. If it needs greater flexibility, we can choose thinner design parameters. 

\item For simulation optimization, only one direction is discussed in this paper. However, the arm designed by us are not isotropic. That is to say, the flexibility and load capacities in the two directions are different. It can be seen from the physical experiment of HPN Arm2 that the load capacity of one certain direction is better, but the flexibility is worse; while the other direction is on the contrary. This paper only discussed the direction with weak load, mainly because the other direction is difficult to simulate. Load capacity of the arm designed in this way surely can meet the requirements, but the flexibility cannot be guaranteed. 

\item Furthermore, the load capacity is calculated when the arm is horizontal in the simulation optimization. The arm designed in this way has greater actual load capacity than the designed one. In other words, the present optimization method cannot design the proper arms exactly meet the needs. A better optimum design method is needed. The topological optimization may be a promising approach in the future work \citep{chen2018topology}. 
\end{enumerate}

\section{Control of HPN manipulators}

In this section, we explore the assumptions and control methods considering the differences of internal and external circumstance to contribute to the long-term success of soft manipulators in practical applications by analyzing the performance of the controllers implemented based on HPN manipulator. Here we discuss the control of soft manipulators under quasi-static assumptions.

In the following, the hardware platform and control methods will be introduced. We will demonstrate the principle of the method, introduce the corresponding physical experiment, analyze the properties, application scenario and limitation of the methods. We will perform control experiment on HPN manipulator using model-based open-loop control, estimated model closed-loop control, and model-free closed-loop control.

\subsection{System overview}

\begin{figure}[htbp]
    \centering
    \includegraphics[width=\columnwidth]{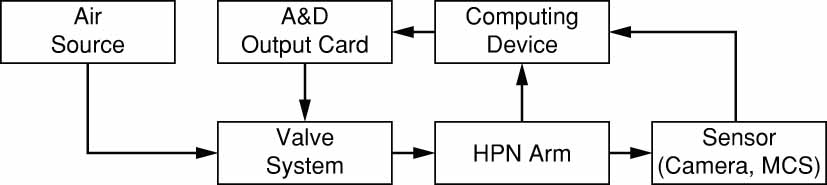}
    \caption{Organization of HPN Arm control system.}
    \label{control_flow_chart}
\end{figure}

\begin{figure}[htbp]
    \centering
    \includegraphics[width=\columnwidth]{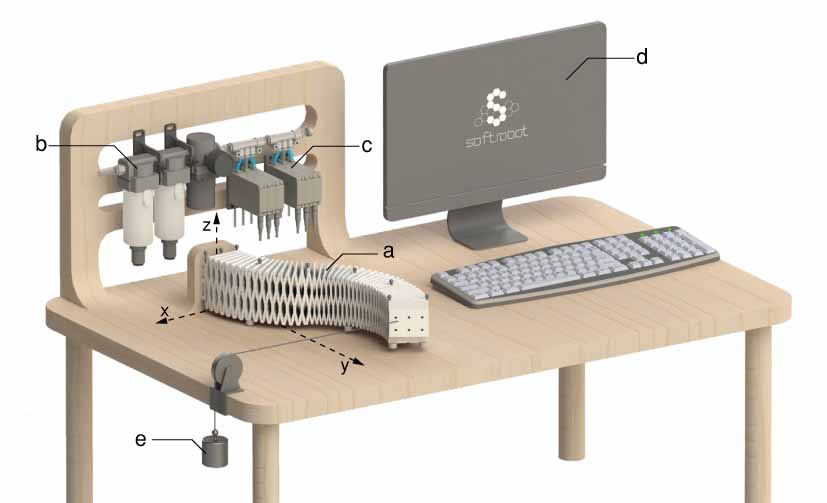}
    \caption{Hardware platform overview. The hardware of control system contains five segments HPN Arm (a), air source treatment(b), valve systems(c), the computing device (d), pulley and load system(e).}
    \label{control_platform}
\end{figure}

The organization of HPN Arm control system is illustrated in Figure \ref{control_flow_chart}. The corresponding hardware platform is illustrated in Figure \ref{control_platform}. The airflow, about 0.7 Mpa, is generated from the air source and preprocessed by the air treatment device and stabilized to 0.3 Mpa, and then directed to the valve system. The control signal from the computing device is converted to analog voltage signal by the programmable logic controller and sent to the valve to adjust the output pressure actuating the HPN Arm 1. Camera or Motion Capture System (MCS) records the real-time position and orientation information of reflective optical markers on the HPN Arm and given target, and send that to the computing device.

It is worth mentioning that, though the control system is identical, the manipulator is not the same for different experiments. The manipulators used for different experiments are as follows: a three-segments manipulator moving in 2D plane is used for model-based open-loop and closed-loop control experiments, a five-segments manipulator moving in 3D space is used for estimated model closed-loop control experiments, and a four segments manipulator moving in 2D plane is used for model-free closed-loop control experiments. Proportional valves detailed in power system section is used in all of these experiments. In 2D experiments, in order to reduce the friction between the manipulator and the table, universal wheels are added between them. In comparative experiments, pulley and load system is used to generate constant external disturbance.

\subsection{Model-based open-loop control}

Due to their material properties, complexity of their structures, non-linearity of their actuation and external disturbances, it is very hard to obtain an open-loop model for soft manipulators. To our best knowledge, from the actuator space,no work has been done to model soft manipulators with all these negative effects (especially viscoelasticity of the material). In theory, data driven method could solve the problem. However, for multi-segment soft manipulators, the dimension of the actuation increases, and the required amount of data would rise exponentially. Besides, because there are many postures to reach the same target, it may be hard to learn the correct inverse kinematic model. So we divide the inverse kinematic model into two levels using the PCC assumption, as shown in Figure~\ref{2d_pcc}: from task space which is the position and orientation of the tip of the manipulator to the configuration space which is the curvature and length of each segment, then from configuration space to actuation space which is the pressure of the airbags. And the two levels will be dealt with by pose optimization and neural network method, respectively.

In this section, we use the first three segment of the experimental platform in figure \ref{control_platform}, which have 6 motion variables.

\begin{figure}[htbp]
\centering
\includegraphics[width=0.95\columnwidth]{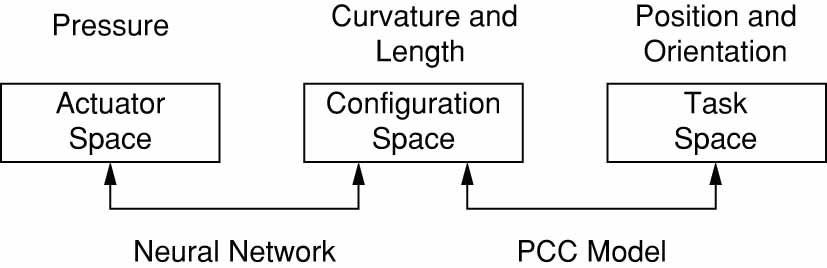}
\caption{Two level approach.}
\label{two_level_act}
\end{figure}

\subsubsection{Method.}
In this part, we realize the position estimation from configuration space to task space through kinematics modeling and find the mapping from task space to configuration space through optimization calculation. The connection between actuation space and configuration space is established by neural network.

\paragraph{Pose Estimation.}
 Before pose optimization, in order to control the pose of the soft robot arm in task space, it is first necessary to estimate an arm segment's pose (curvature and arc length) in configuration space using available localization data.

\begin{figure}[htbp]
\centering
\includegraphics[width=0.95\columnwidth]{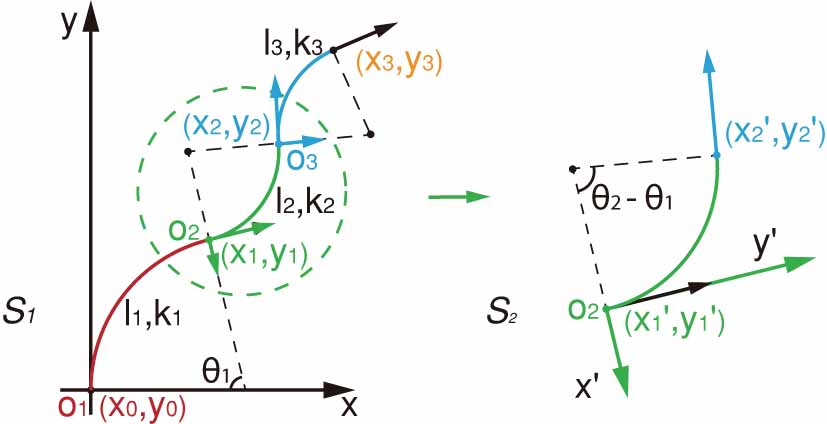}
\caption{The kinematics model of HPN Arm 1 is described in Cartesian coordinate system. ($x_i, y_i$) represents the coordinate in the global coordinate system while ($x_i^{'}, y_i^{'}$) represents the relative coordinate in the local coordinate system for segment $i$. $k_i$ and $l_i$ represent the curvature and arc length of segment $i$ respectively, and $\theta_i$ represents the sum of first $i$ segments’ bending angles.}
\label{2d_pcc}
\end{figure}

To formalize the problem of calculating $\mathcal{P}_{C-T}$ under PCC assumption, we first build right-handed Cartesian coordinate systems $S_1, S_2 , ... , S_n$ for each segment respectively, by fixing the positive direction of y-axis to the starting tangent vector of each segment's bending arc, as shown in Figure~\ref{2d_pcc}. Specifically, $S_i$ is the local coordinate system for segment $i$, while $S_1$ is also set as the global coordinate system. The calculation procedure is shown in Algorithm \ref{alg1}, for each segment $i$, we convert the global coordinate $(x_i, y_i)$ in $S_1$ to local coordinate $(x^{'}_i, y^{'}_i)$ in $S_i$, in order to figure out the curvature $k_i$ and arc length $l_i$. And $\theta_i$ represents the angle between the tangent vector at the tip of segment $i$ and y-axis in $S_1$, which can be calculated by adding $k_i l_i$ to $\theta_{i-1}$.

\begin{algorithm}
  \caption{Parameter Estimation}
  \footnotesize
  \begin{algorithmic}[1]
  \Function{Estimation}{$\vec{X}, \vec{Y}, n$}\Comment{n: segment number}
  \State $k_1 \gets \dfrac{2x_1}{x^{2}_1 + y^{2}_1},~\theta_1 \gets 2\arctan \dfrac{x_1}{y_1},~l_1 \gets \dfrac{\theta_1}{k_1}$
  \For {$i = 2, i \leq n, i++$} 
 
  \State $x^{'}_i \gets \cos{\theta}_{i-1}(x_i - x_{i-1}) - \sin{\theta}_{i-1}(y_i - y_{i-1})$
  \State $y^{'}_i \gets \sin{\theta}_{i-1}(x_i - x_{i-1}) + \cos{\theta}_{i-1}(y_i - y_{i-1})$

  \State $k_i \gets \dfrac{2x^{'}_i}{x^{'2}_i + y^{'2}_i},~l_i \gets \dfrac{2}{k_i}\arctan \dfrac{x^{'}_i}{y^{'}_i}$
  \State $\theta_i \gets k_i l_i + \theta_{i - 1}$
  \EndFor
  \EndFunction
  \end{algorithmic}
  \label{alg1}
\end{algorithm}

\paragraph{Pose optimization. }
We model the problem of getting $\mathcal{P}_{C-T}^{-1}$ from $\mathcal{P}_{C-T}$ as an optimization task. A combined cost function containing three basic functions is set for evaluating a certain pose (positions of all the segment tips). Each basic function is represented as a penalization cost term, which is proportional to the current pose's deviation from the preset requirements. Using gradient descent, an optimal pose under the measure can be figured out. 

As shown in Algorithm \ref{alg2}, the tip position of all the segments, $(\vec{X}, \vec{Y})$, are initialized by randomly selecting the position of the tips of middle segments, and transformed to the curvatures and arc lengths $(\vec{K}, \vec{L})$ using Algorithm 1, where $(x_t, y_t, \theta_t)$ represents the position and orientation of target, and $k_{max}, k_{min}, k_{avg}, l_{max}, l_{min}, l_{avg}$ represent the maximum, minimum and average of curvature and arc length respectively, which can be measured in advance. 

For the tip of each segment, the theoretical feasibility of reaching a certain position is considered in $Cost_1$, where $f$ represents whether the combination of curvature $k$ and arc length $l$ of one segment are in the available range. If $f < 0$, the corresponding $k, l$ are in the available range, and $Cost_1$ is close to 0. Otherwise, $Cost_1$ sharply grows to a great positive value, which implicates unavailable $k, l$. Besides, $\delta$ ($\delta \ll 1$) ensures smooth variance when $f$ is close to 0. 

The accuracy requirement of the arm tip's arrival orientation is considered in $Cost_2$, which grows to a relatively large value when there is a deviation, where $\theta_t, \theta_r$ represent the orientation of the target and arm root respectively, and $\theta_n$ represents the total bending angle of first $n$ segments.

$Cost_3$ ensures all segments in a certain pose with similar elongation rates and least sum of the absolute of bending angles because such poses are natural and adaptable. Specifically, the pose composed of evenly deformed segments provides space for further elongations and bends.

Finally, weighted with different empirical parameters according to the significance ($\alpha > \beta > \gamma$), three basic cost functions compose an integrated cost function $Cost$. Then, using $fminunc$ in MATLAB Toolbox, the optimal positions of each arm segment's tip $(\vec{X}^*, \vec{Y}^*)$ can be figured out, and then transformed to $(\vec{K}^*, \vec{L}^*)$ as the inputs of neural network detailed in next section.

\begin{algorithm}
  \caption{Pose Optimization}
  \label{alg2}
  \footnotesize
  \begin{algorithmic}[1]
  \Function{Optimization}{$x_t, y_t, \theta_t, n$}\Comment{n: segment number}
  \State $(\vec{X}, \vec{Y}) \gets Initialization (x_t, y_t, \theta_t)$
  \State $(\vec{K}, \vec{L}) \gets \Call{Estimation}{\vec{X}, \vec{Y}, n}$
  \State $f \gets \left| \dfrac{2(l - l_{avg})}{l_{max} - l_{min}}\right| + \left| \dfrac{2(k - k_{avg})}{k_{max} - k_{min}}\right| - 1$
  \State $Cost_1 \gets \sum_{i = 1}^n \left(\sqrt{f_i^2 + \delta^2} + f_i \right)$
  \State $Cost_2 \gets (\theta_t - \theta_r - \theta_n)^4$
  \State $Cost_3 \gets \sum_{i = 1}^n \left( \dfrac{l - l_{avg}}{l_{max} - l_{min}}\right)^2 + \left(\dfrac{k - k_{avg}}{k_{max} - k_{min}} \right)^2$
  \State $Cost \gets \alpha Cost_1 + \beta Cost_2 + \gamma Cost_3$
  \State $(\vec{X}^*, \vec{Y}^*) \gets fminunc(Cost, (\vec{X}, \vec{Y}))$ \Comment{MATLAB Toolbox}
  \State $(\vec{K}^*, \vec{L}^*) \gets \Call{Estimation}{\vec{X}^*, \vec{Y}^*, n}$
  \EndFunction
  \end{algorithmic}
  \label{algo3}
\end{algorithm}

\paragraph{Neural network.}
Another critical component to the main control algorithm is the neural network, $Net$, which learns the relationship between the configuration space and the actuation space.

 Because the adjustment of the posture of the manipulator can not complete in an instant, we fix the time interval of each position control as 6 seconds. And feed-forward neural network is chosen to be the fitting tool.

\begin{figure}[!htbp]
    \centering
    \includegraphics[width=0.95\columnwidth]{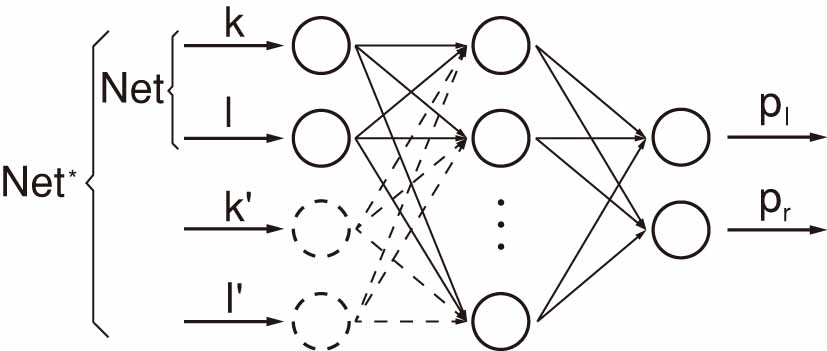}
    \caption{Neural network $Net$ is used to figure out the relationship between configuration space and actuation space with two inputs: curvature $k$ and arc length $l$, and two outputs: pressures on the left and right side, $p_l, p_r$. Advanced $Net^*$  has four inputs, added the curvature $k^{'}$ and arc length $l^{'}$ of last pose, to compensate the effect of viscoelasticity.}
    \label{network}
\end{figure}

Because different segments of HPN manipulator has different parameters, one neural network should be trained for each segment respectively. As shown in Figure~\ref{network}, We use the target curvature $k$ and the arc length $l$ as input, and the pressure of the airbags $p_l$, $p_r$ are output. In order to consider the effect of viscoelasticity, $net^{*}$ added current curvature $k^{'}$ and length $l^{'}$ as input, which is the target of the last step.

In the training process of $Net_i$ and $Net_i^*$ for the $i$th segment, we first generate enough training data by pressurizing the airbag groups using pressure samples $({P_l}, {P_r})$ randomly selected in the available pressure range $(P_{min}, P_{max})$ and record the segment tip’s positions $({X}, {Y})$ and converted to $({K}, {L})$ using Algorithm \ref{algo3}. With the data pairs, $((K,L), ({P_l}, {P_r}))$,or$((K,L,K^{'} ,L^{'} ),({P_l}, {P_r}))$ we can get $Net_i$ and $Net_i^*$ using train function in MATLAB Toolbox.

It is worth mentioning that, since $Net^*$ is well trained by a large amount of data, we use the estimated $k_0$, $l_0$ corresponding to the last target as its input instead of the real last pose performed by the arm. Which means the strategy can be used under open-loop environment.%, though it would be better to use real last pose as input. 
Thus, consecutive open-loop motion control with compensation for viscoelastic effect can be realized.

\begin{algorithm}[!htbp]
  \caption{Main Control Algorithm}
  \footnotesize
  \begin{algorithmic}[1]
    \Function{Control}{$x_t, y_t, \theta_t, n$} \Comment{n: segment number}
    %\State $Depressurization$
    \State $(\vec{K}, \vec{L}) \gets \Call{Optimization}{x_t, y_t, \theta_t, n}$ 
    \For {$i = 1, i \leq n, i++$}
    \State $(P_{li}, P_{ri}) \gets Net_i(k_i, l_i)$
    \EndFor
    \State $Pressurization~using~\vec{P_l}, \vec{P_r}$
    \EndFunction
  \end{algorithmic}
    \label{alg4}
\end{algorithm}

\subsubsection{Experiment. }%For a manipulation system, it’s essential to implement precise position and orientation control of the soft arm’s tip. 
The control algorithm is able to accurately move the tip of the manipulator to a specified position and orientation, which is referred as point-to-point task. In this section, we performed point-to-point experimental tests on single and multi-segment arms.

 \paragraph{Single segment test.}

\begin{figure*}[!htbp]
    \centering
    \includegraphics[width=\textwidth]{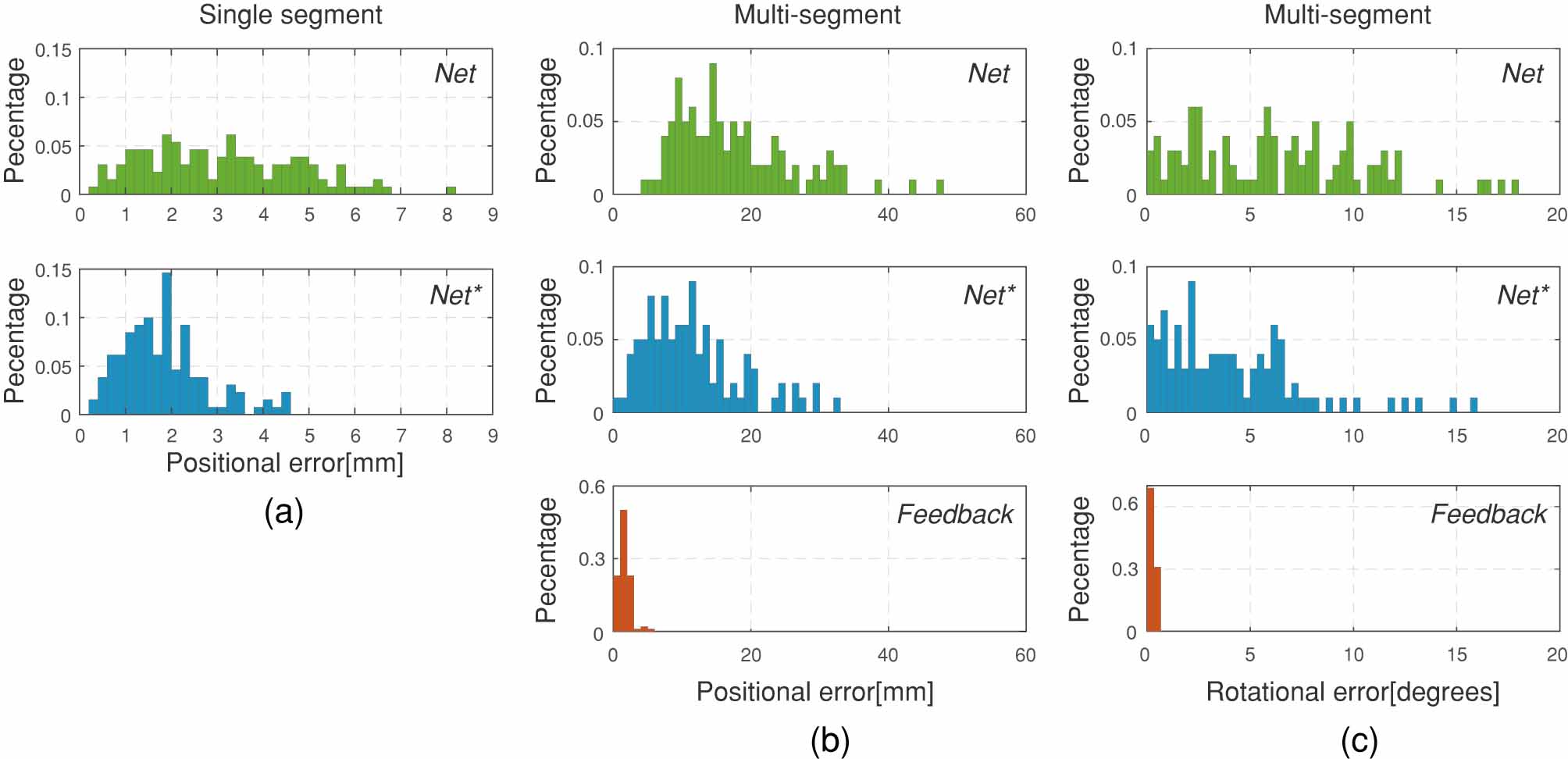}
    \caption{(a) illustrates the comparison of single segment's precision. It can be observed that the average error of normal neural network is 3.11 mm, while the average error of the network considering last target is 1.82 mm. (b) and (c) illustrate the control error of multi-segment arm. In the case of open-loop control, the normal neural network has a positional error of 17.54 mm, and a rotational error of 6.26 degrees, while the average error of $Net^*$ is 11.45 mm and 4.04 degrees. And in the case of closed-loop control, the positional and rotational errors are greatly decreased to 1.61mm and 0.26 degrees respectively.}

    \label{single_multi_modelbased_data}
\end{figure*}

A foundation for multi-segment control is that a single segment can be controlled to achieve the desired curvature and arc length. For single segment, according to the piece-wise constant curvature model, its tip's orientation can be calculated by its tip's position, so only the position is concerned in this section. We select the second segment of the arm as an example, Figure~\ref{single_multi_modelbased_data}(a) details the positional errors. We randomly generate pressure samples distributed in $0-0.3Mpa$ and collect the positions of segment's tip. Using 2000 pairs of data (pressures and positions of segment's tip), we train the Net with 25 neurons in a hidden layer. Its precision is tested using 100 random targets.

The result is shown in Figure~\ref{single_multi_modelbased_data}(a), the neural network net that does not consider previous state has an average error of 3.11 mm. The neural network net* considered previous state has an average error of 1.82 mm and provides a better solution. The accuracy of the latter is 40\% higher than that of the former.

\paragraph{Multi-segment control.}
For multi-segment control, the first three segments of the HPN Arm 1 are used for precision test. Neural networks corresponding to the three segments are trained separately, similar to the process in the single segment test. 100 targets are randomly selected in the workspace and used as input to the main control algorithm (Algorithm 3). The positional and rotational error distributions are detailed in Figure ~\ref{single_multi_modelbased_data}(b)(c).

In Figure ~\ref{single_multi_modelbased_data}(b)(c), when the last state is not considered the mean positional and rotational errors are 13.56mm and 4.39 degrees respectively. When the last state is considered, the corresponding results are 11.45mm and 4.04 degrees. It is obvious that the strategy handling viscoelasticity efficiently reduces the average positional and rotational errors. Moreover, the decrease of maximum positional error implies the strategy also improves the control stability.

\subsubsection{Limitation.}

\begin{enumerate}[(a)]
\item Open loop control is used in actuation space. Compared with the control method with configuration space feedback, this method saves the efforts of using sensors but makes our controller more sensitive to disturbances. In addition, common control methods using configuration space feedback will use PID control in the configuration space, which will be faster than our open-loop controller.

\item When training the mapping relationship between configuration space and actuation space, every data is collected in a 6 seconds time step. During this 6 seconds, the arm is not completely stationary, so the final open-loop control can only guarantee the accuracy in this 6 s, but not after.

\item Only the last pose of the segment is taken as an additional input of the neural-network ($Net^*$). However, it would be better to take more former poses into consideration if the training time is acceptable.

\item This algorithm’s application range is now restricted in 2D plane. In future work, enhanced neural networks considering the gravity effect of external load and arm itself can be designed in order to extend the algorithm to 3D space.

\item Differing from those soft arms with rigid plates between segments \citep{grissom2006design}, the HPN arm is a uniform one without obvious boundary between two segments. Therefore, it is not reasonable to assume the deformation of each segment is independent as mentioned in Section. This may be the reason why the precision of multi-segment control is not as satisfactory as that in single segment test.

\end{enumerate}

\subsection{Hybrid closed-loop control}

When the open-loop accuracy does not meet our accuracy requirement, we use closed-loop control to improve the accuracy. Here based on the hybrid open-loop control method demonstrated above, we simply use the method of translating the target to achieve closed-loop control. This requires that the control algorithm should have good monotonicity and would not cause the actual motion direction opposite to the target. In this section, we improve closed-loop control accuracy using feedback strategy, demonstrate path tracking experiment and test the stability of the system.

\subsubsection{Method.}
In this section, we use a scheme that integrates the main control algorithm with additional simple feedback strategy.

Specifically, we use the error between the desired pose $\Vec{q} = (x, y, \theta)$ and real pose $\Vec{q}^{'} = (x^{'}, y^{'}, \theta^{'})$ as a modification term of the input $\Vec{q}^* = (x^*, y^*, \theta^*)$ to the control system in the iteration process:

\begin{equation}
\Vec{q}^* \gets \Vec{q}^* + \alpha(\Vec{q} - \Vec{q}^{'})
\end{equation}
Where $\alpha$ represents the modification rate. The iteration process ends when the error falls into an acceptable range, or the number of iterations reach a limit (set as 10 in practice).

If the target is moving, dynamically compensate for the previous errors: the error in the ith subtask is inherited to the determination of the (i+1)th target, which can be represented as:

\begin{equation}
\Vec{q}^*_{i+1} \gets \Vec{q}^*_i + (\Vec{q}_{i+1} - \Vec{q}_i) + \beta (\Vec{q}_i - \Vec{q}^{'}_i), ~~i \ge 1
\end{equation}
Where $\Vec{q}^*_i$ = $(x_i, y_i, \theta_i)$ represents the $i$th input to the control system, and $q_i$, $\Vec{q^{'}_i}$ represent the $i$th desired pose and real pose respectively, while $\beta$ represents the modification rate. 

\subsubsection{Experiment }
In this section, we have done the point-to-point control experiment and the path tracking experiment of multi-segment arms.
\paragraph{point-to-point.}
In the point-to-point experiment, the process contains 5 iterations for every point on average, each taking 1s for sensing, calculation and actuation.

As shown in Figure~\ref{single_multi_modelbased_data}(b)(c), with feedback strategy, the mean positional and rotational errors of 100 trials are 1.61 mm and 0.26$^\circ$ respectively, much less than those in former open-loop control.

\begin{figure}[tbp]
    \centering
	\epsfig{figure=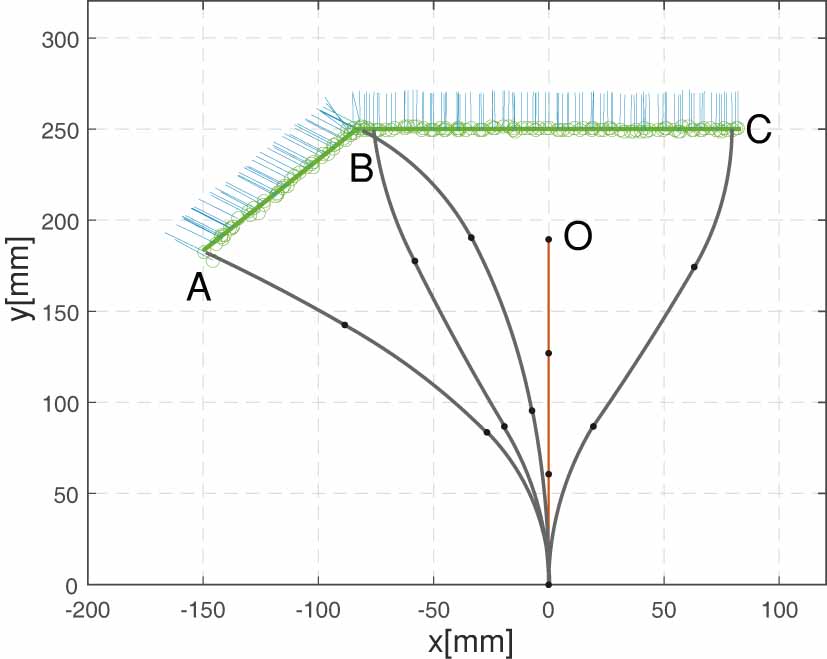,width=\columnwidth} 
    \caption{The experiment of path tracking program. The target path is shown in magenta. The inclined line, with an angle of $\frac{\pi}{4}$, has a target orientation of $\frac{\pi}{3}$ relative to y-axis, and the horizontal line has a target orientation parallel to y-axis. The arm has an original pose O, and the path's start A, end C and turning point B. The green circle represents each step's position, while the blue line represents corresponding measured orientation.}
    \label{pathtracking}
\end{figure}

\begin{figure}[htbp]
    \centering
    \epsfig{figure=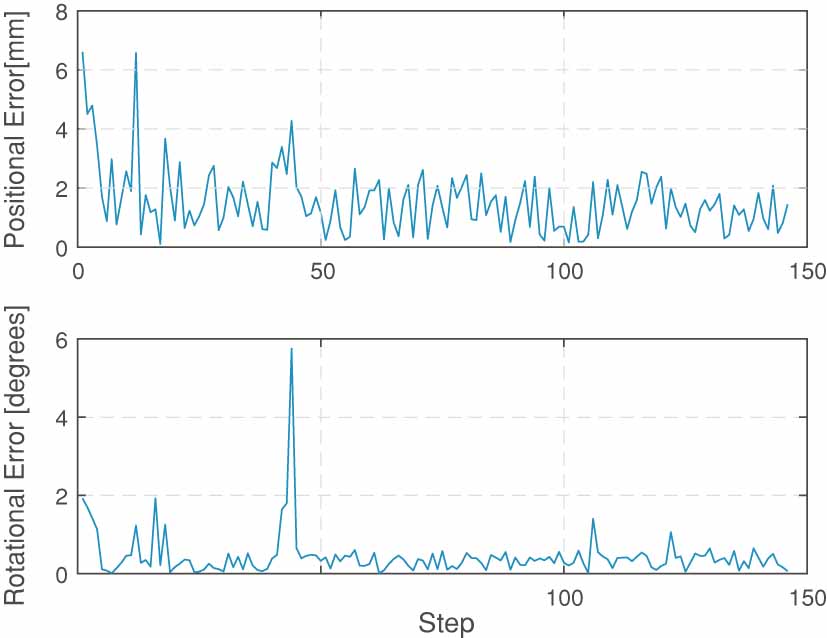,width=\columnwidth}
    \caption{Error records in path tracking. The positional error is relatively high at the beginning and the turning point. The mean positional error is 1.52 mm. The rotational error sharply grows at the turning point. The mean rotational error is 0.43$^\circ$.}
    \label{path_error}
\end{figure}

\paragraph{Path Tracking.}
The path tracking experiment is conducted to test the precision and stability of the control system. This problem is decomposed to numbers of sub-tasks: reach numerical targets distributed on the path. As the targets are distributed closely, we set $\beta$ as 1, so we get:
	
\begin{equation}
\Vec{q}^*_{i+1} \gets \Vec{q}^*_i + \Vec{q}_{i+1} - \Vec{q}^{'}_i, ~~i \ge 1
\end{equation}

figure~\ref{pathtracking} shows several arm's measured pose at the experiment's start (O), path's start (A), path's turning point (B) and path's end (C). The measured arm tip's pose is shown at each step along the path, where the position and orientation are represented as green circles and blue lines respectively. The error result is shown in figure~\ref{path_error} with a mean positional and rotational error of 1.52 mm and 0.43$^\circ$. The rotational error sharply grows at the turning point (B) because the arm has a huge state change.

From the track recorded in the experiment, we can see that an arm with a large extension ratio is able to finish a task in a natural way, which means it can move in combination of elongation and smooth bending when reaching a target. In contrast, an inextensible one may have a large reachable space, but it needs to bend large angles to reach a close target on its side. The process of these experiments can be seen in Extension 2.

\subsection{Estimated model closed-loop control}

In the interactive tasks, it is impossible to find an accurate model to represent the actual state of soft manipulators when uncertain disturbance or load is present, due to their passive compliance. In this case, it is impossible to achieve accurate control without feedback at task level or feedback at configuration level that can determine the task space accurately, such as feedback using shape sensors or length sensors.

On the other hand, if the relied accurate model is made less precise in practice, feasible solution may not be obtained. For example under external loads, the degree of bending of a single segment may exceed that estimated by the model. And in this case, for a simple feedback traslational move, the model may give a solution in the opposite direction.

Since in practice, soft manipulators can not be modeled accurately, and the control may be limited by accurate model, we turned to another way of thinking that we should find a robust model for feedback control, and should not focus on its accuracy.

We found that the method of Jacobian matrix is a good choice. For the tip of the manipulator, the Jacobian matrix always represent its response to the actuation in the current state. As long as the dot product of the direction directed by the Jacobian matrix and the real direction is positive, which means that the tip is moving near the target, the target should be reached after some times of feedback iteration. As long as the effect of actuation or combination of actuation on the state is found, and the description of the effect is robust enough, it is a robust strategy.

In this section, we introduce the feedback control method based on an estimated Jacobian model, which is implemented on a 3 dimensional HPN arm. The experimental platform is shown in figure \ref{control_platform}. A small difference is that the base of the arm is lifted up off the table. With the target as reaching desired end effector’s position and direction, the manipulator’s structure and its deformation mode are analyzed for selecting appropriate variables as representatives in actuation space and task space. Based on the analysis, the Jacobian matrix is built and simplified. The controller framework is detailed in Figure \ref{model_less_framework}.

\subsubsection{Method.}
\begin{figure*}[t]
    \centering
    \includegraphics[width=\textwidth]{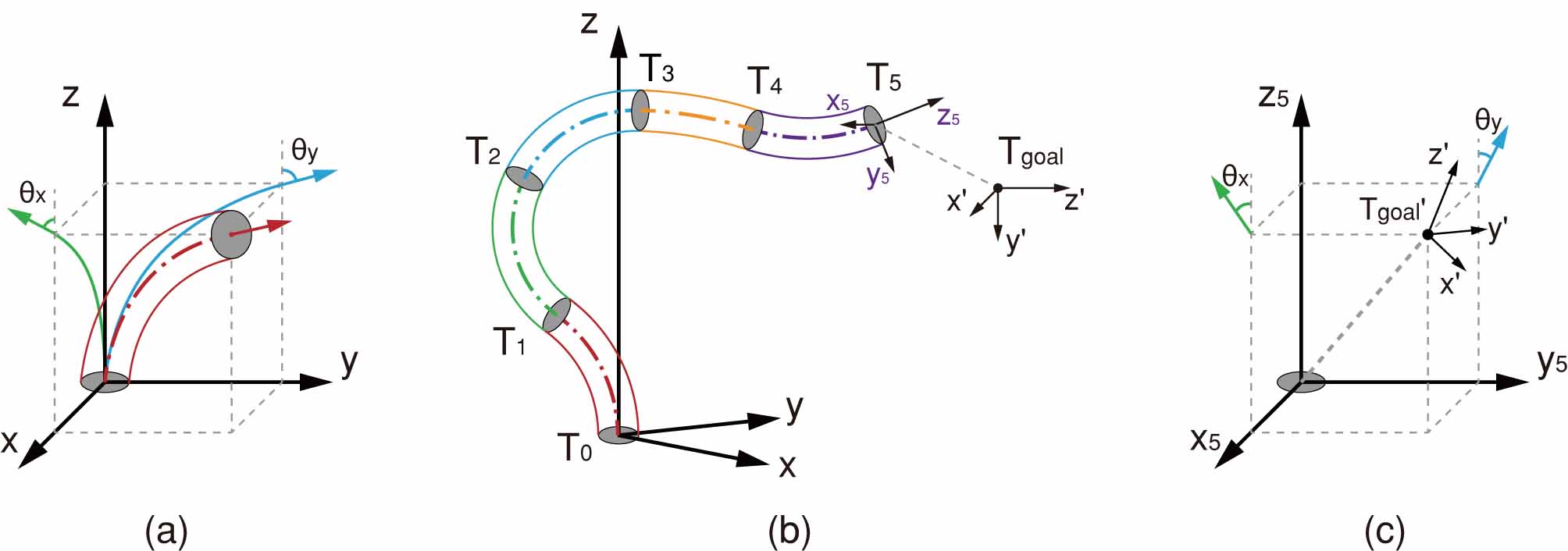}
    \caption{Estimated model schematic diagram of single segment and multi-segments. (a) represents the pose representation of a single segment. Since the single segment bend does not exceed 90 degrees and the twist is ignored, the direction can be projected onto the $x-z$ plane and the $y-z$ plane. (b) The pose information between the segments is represented by homogeneous transformation matrix. (c) shows the pose and tip error of the target point, which is shown in the tip system and projected in the same way as a single segment.}
    \label{model_less_lineart}
\end{figure*}

\begin{figure}[!htbp]
    \centering
    \includegraphics[width=0.9\columnwidth]{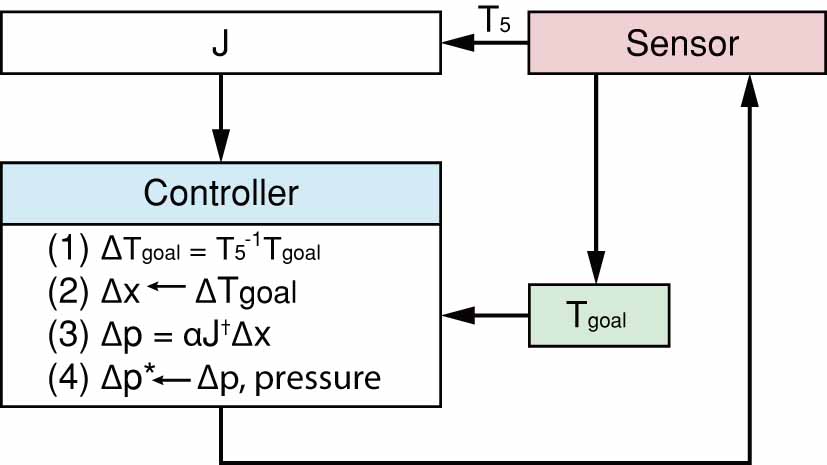}
    \caption{The figure shows the framework of a proportional feedback controller based on estimated Jacobian matrix $\textbf{\textit{J}}$. During the feedback loop, the actuation change $\Delta \textbf{\textit{p}}$ is calculated from task space distance $\Delta \textbf{\textit{x}}$ and pseudo-inverse of Jacobian $\textbf{\textit{J}}^{\dagger}$, and scaled by damping ratio a to get real change $\Delta \textbf{\textit{p}}^{*}$.}
    \label{model_less_framework}
\end{figure}

This section mainly introduces the method of estimating Jacobian, we  only  estimate Jacobians of individual segments, and calculate the overall result based on end effector pose geometric estimation as shown in figure \ref{model_less_lineart}. With the jacobian matrix, the control objectives can be achieved through an iterative approach which is shown in figure \ref{model_less_framework}.

\paragraph{Single segment Jacobian.}

For one-segment manipulators, here we define three generalized actuations: $p_{sx}$, $p_{sy}$ and $p_{sz}$, which correspond to bending in two directions and the elongating respectively. Because our one-segment manipulator would not bend more than $90^{\circ}$, the orientation of the tip of the manipulator could be projected to the $x-z$ plane and $y-z$ plane of the coordinate system at the base of the manipulator. The angles between the two projections and the $z$-axis are called $\theta_{sx}$ and $\theta_{sy}$ respectively. So the state of the tip could be represented by a vector $\Vec{X}=(x_s~ y_s ~z_s ~\theta_{sx} ~\theta_{sy})$. One limitation of this representation is that the angle between the orientation of the tip and the $z$-axis needs to be less than $90^{\circ}$ and the manipulator would not rotate around the $z$-axis. Because our one-segment manipulator could not bend over $90^{\circ}$ and has no rotational DoF, this representation could be used. Then the assumptions for simplifying are made as follows: generalized actuation $p_{sx}$ would only make the tip of the manipulator move in the $x-z$ plane along the $x$ axis and rotate around the $y$-axis, so it only changes $x_s$ and $\theta_{sx}$; generalized actuation $p_{sy}$ would only make the tip of the manipulator move in the $y-z$ plane along $y$- axis and rotate around the $x$-axis, so it only changes $y_s$ and $\theta_{sy}$; and generalized actuation $p_{sz}$ could only make the tip move along $z$-axis, so it only changes $z_s$.

\begin{figure}[!htbp]
    \centering
    \includegraphics[width=\columnwidth]{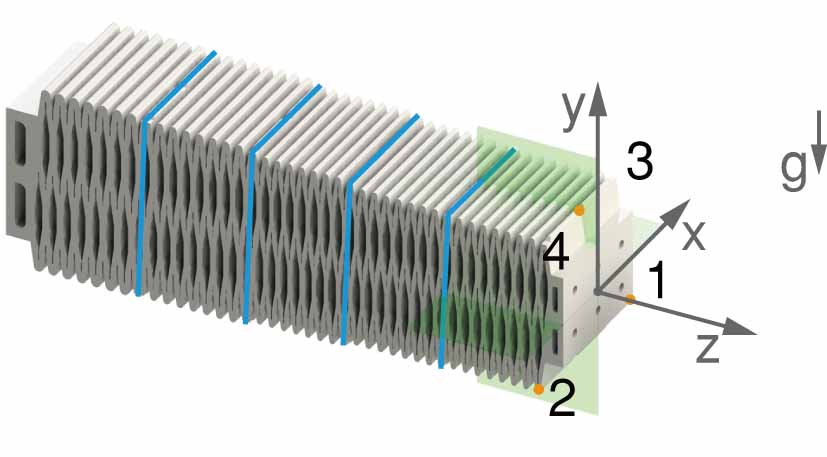}
    \caption{Five segments of the arm}
    \label{control_model_less_manipulator}
\end{figure}

According to the working principle of HPN articulated shown in figure \ref{control_model_less_manipulator}, the three generalized actuations are determined by the pressure of four groups of airbags. And they have relationships as follows:

\begin{equation}
\left\{
\begin{array}{l}
p_{sx} = -p_{s1} + p_{s2} - p_{s3} + p_{s4} \\
p_{sy} = ~~~p_{s1} + p_{s2} - p_{s3} - p_{s4} \\
p_{sz} = ~~~p_{s1} + p_{s2} + p_{s3} + p_{s4} \\
\end{array}
\right.
\end{equation}

Considering the actual working state of the manipulator, the four actual actuations are not completely independent from each other. They have to satisfy the relationship $p_{s1}+p_{s4}=p_{s2}+p_{s3}$. Then the equations above could be solved uniquely:

\begin{equation}
\left\{
\begin{aligned}
p_{s1}=(p_{sz} - p_{sx} + p_{sy})/4\\
p_{s2}=(p_{sz} + p_{sx} + p_{sy})/4\\
p_{s3}=(p_{sz} - p_{sx} - p_{sy})/4\\
p_{s4}=(p_{sz} + p_{sx} - p_{sy})/4\\
\end{aligned}
\right.
\end{equation}

So, for one-segment manipulator, three generalized actuations correspond to the change of five parameters describing the tip of the manipulator. This could be represented by a 5×3 matrix of partial differential, namely Jacobian Matrix of the one-segment bending soft manipulator. And due to our assumptions above, there are many zero elements in the matrix, which is shown below:

\begin{equation}
\textit{\textbf J}=%\left(\dfrac{\partial \textit{\textbf{x}}_i}{\partial \textit{\textbf{p}}_j}\right)_{i, j}=
\left[
\begin{matrix}
\dfrac{\partial \theta_{sx}}{\partial p_{sx}} & 0 & 0 \\
\dfrac{\partial x_s}{\partial p_{sx}} & 0 & 0 \\
0 & \dfrac{\partial \theta_{sy}}{\partial p_{sy}} & 0 \\
0 & \dfrac{\partial y_s}{\partial p_{sy}} & 0\\
0 & 0 & \dfrac{\partial z_s}{\partial p_{sz}} \\
\end{matrix}
\right]
\end{equation}

%Generalized actuations could be represented by a vector $p=[p_x, p_y, p_z]$. When the generalized actuations are change by d$p$, the position and orientation change of the tip after applying dp could be estimated by $dX=J\,\mathrm{d}p$. Then the new $X$ could be calculated from d$X$ and the old $X$.

Besides the vector $\Vec{X}=(x_s~ y_s ~z_s ~\theta_{sx} ~\theta_{sy})$ that can be used to calculate Jacobian matrix, in order to calculate information between segments, the state of each segment's tip is also represented by a coordinate system at the tip. The orientation of the tip is defined as $z^{'}$ axis. Then rotate $z$ axis to $z^{'}$ axis, and the base coordinate system is rotated with z to get the orientation of $x^{'}$ axis and $y^{'}$ axis. Then the new coordinate system is translated to the tip to get the tip coordinate system. This whole linear transformation could be represented by an augmented matrix $T$, which represents the information of $x$-axis, $y$-axis, $z$-axis and position of the tip coordinate system. $T$ is shown as follows:

\begin{equation}
\textit{\textbf T}=\left[
\begin{matrix}
ex1 & ey1 & ez1 & x\\
ex2 & ey2 & ez2 & y\\
ex3 & ey3 & ez3 & z\\
0 & 0 & 0 & 1\\

\end{matrix} 
\right]
\end{equation}

The $3\times3$ square submatrix at the upper left corner of the matrix is the rotation matrix for rotating the base coordinate system to the tip coordinate system, from which $\theta_x$ and $\theta_y$ could be calculated. When the angle between the tip and the $z$-axis of the base coordinate system is no more than 90 degrees, this representation of augmented matrix is fully equivalent to the representation of the vector $\Vec{X}(x_s~ y_s ~z_s ~\theta_{sx} ~\theta_{sy})$.

\paragraph{Multi-segment Jacobian.}

For multi-segment manipulators, let the augmented matrix of the base with respect to the base coordinate system be $T_0$, which is an identity matrix , and let the augmented matrices for the tips of each segment with respect to the base coordinate be $T_1,T_2,T_3,T_4,T_5$ respectively. Let the augmented matrix of the goal be $T_{goal}$.

Let the augmented matrices of tips of each segment with respect to its own base be ${^0}T_1$, ${^1}T_2$, ${^2}T_3$, ${^3}T_4$, ${^4}T_5$. Then we have ${^i}T_j=T_i^{-1}\,T_j$, and $T_5=T_0\,{^0}T_1\,{^1}T_2\,{^2}T_3\,{^3}T_4\,{^4}T_5$. In order to represent the effect of a single generalized actuation of a certain segment on the tip of the whole manipulator, in the multi-segment control algorithm, it can be assumed that when the configuration of a certain segment changes, the posture of other segments won’t be influenced. For example, if the posture of the second segment is changed to ${^1}T_2^{'}$, the augmented matrix of the tip will be $T_5^{'}=T_0 \,{^0}T_1 \,{^1}T_2^{'}\,{^2}T_3 \,{^3}T_4\,{^4}T_5$.

The deviation is calculated in the tip coordinate system. The augmented matrix of the goal with respect to the tip of the manipulator could be calculated as $T_{goal}^{'}=T_5^{-1}\,T_{goal}$, which represents the deviation of the position and orientation between the goal and the tip and could be transformed to the vector $\Vec{X_{goal}^{'}}=(x~ y ~z ~\theta_x ~\theta_y)$ when the angle between the $z$-axis of the tip coordinate system and the goal coordinate system is less than 90 degrees.

For the whole manipulator, there are 15 generalized actuations in total controlling 5 parameters of the tip. So the size of its Jacobian matrix is $5\times15$:

\begin{equation}
\textit{\textbf J}=\left(\dfrac{\partial \textit{\textbf{x}}_i}{\partial \textit{\textbf{p}}_j}\right)_{i, j}=\left[
\begin{matrix}
\dfrac{\partial \theta_x}{\partial p_{1x}} & \dfrac{\partial \theta_x}{\partial p_{1y}} & \dfrac{\partial \theta_x}{\partial p_{1z}} & \cdots & \dfrac{\partial \theta_x}{\partial p_{5z}}\\
\dfrac{\partial l_x}{\partial p_{1x}} & \dfrac{\partial l_x}{\partial p_{1y}} & \dfrac{\partial l_x}{\partial p_{1z}} & \cdots & \dfrac{\partial l_x}{\partial p_{5z}}\\
\dfrac{\partial \theta_y}{\partial p_{1x}} & \dfrac{\partial \theta_y}{\partial p_{1y}} & \dfrac{\partial \theta_y}{\partial p_{1z}} & \cdots & \dfrac{\partial \theta_y}{\partial p_{5z}}\\
\dfrac{\partial l_y}{\partial p_{1x}} & \dfrac{\partial l_y}{\partial p_{1y}} & \dfrac{\partial l_y}{\partial p_{1z}} & \cdots & \dfrac{\partial l_y}{\partial p_{5z}}\\
\dfrac{\partial l}{\partial p_{1x}} & \dfrac{\partial l}{\partial p_{1y}} & \dfrac{\partial l}{\partial p_{1z}} & \cdots & \dfrac{\partial l}{\partial p_{5z}}\\
\end{matrix}
\right]
\end{equation}

Each column of this Jacobian matrix could be calculated separately using numerical method. For example, to calculate the eighth column, which represents the impact of $p_{3y}$ on the tip, first a differential element $\mathrm{d}p_{3y}$ is added to $p_{3y}$, then the new position and orientation of the tip of the third segment could be approximated by ${^2}T_3^{'}$ using single-segment Jacobian matrix as shown in the last section. So the augmented matrix of the tip of the whole manipulator would be $T_5^{'}=T_0 \,{^0}T_1 \,{^1}T_2\, {^2}T_3^{'}\,{^3}T_4 \,{^4}T_5$. The deviation of $T_5^{'}$ from $T_5$ could be calculated by $T_5^{-1}\,T_5^{'}$, which could be represented by a vector (d$x$, d$y$, d$z$, d$\theta_x$, d$\theta_y$) as well. And this vector is the eighth column of the whole manipulator’s Jacobian matrix. Similarly other columns of the matrix could be calculated.

For multi-segment manipulator, the Jacobian Matrix is related to designed physical parameters. They are usually different for different design, and needs to be initialized. In the whole control process, because a single segment is short, its bending angle would be not larger than 90 degrees, it could be assumed that the Jacobian Matrix of each segment is unchanged. Due to the nonlinearity of actuation response, the Jacobian Matrix describing the characteristic of actuation should be different at different posture in principle. So in order to guarantee the stability and robustness of the control strategy, the non-zero elements of the Jacobian Matrix should be their maximum value. 

\paragraph{Feedback strategy. }

In the actual control process, the current position of the manipulator is obtained using sensors, and the Jacobian matrix is updated using the current position. $T_{goal}$ is obtained, and $T_{goal}^{'}$ is calculated and transfromed into d$x$. Then in combination with the Jacobian matrix, the generalized actuations could be calculated. Finally, the actual actuations could be calculated from the generalized actuations.

\paragraph{Approximation for simplification. }

The method presented above requires the information of the segments in the middle. If that information is not convenient to be obtained, it could be approximated using linear interpolating between $T_5$ and $T_0$. Though linear interpolation may result in a large positional error for the segments in the middle, its orientation is relatively accurate. And the angel is more important for the estimated model closed-loop control because we have to ensure the tip moves near the target for every iteration, while the effect of the position is smaller.

Because our manipulator does not have a DOF of rotation, we only control its tip's position and orientation of $z$-axis.

Because we use the Jacobian method, it is not allowed that the orientation indicated by the Jacobian matrix is wrong, so the angle between the approximated direction and the actual direction of motion should be no more than $90^{\circ}$. If the linear interpolation is used, there may be some certain postures at which the angle is above $90^{\circ}$, so they could not be reached. Thus this method may not be able to take advantage of all the performance of the manipulator.

\subsubsection{Experiment}
In this section, we have done the point-to-point control experiment and the path tracking experiment under different loads.
\paragraph{Point-to-point under different loads.}

\begin{figure}[!htbp]
\centering
    \includegraphics[width=\columnwidth]{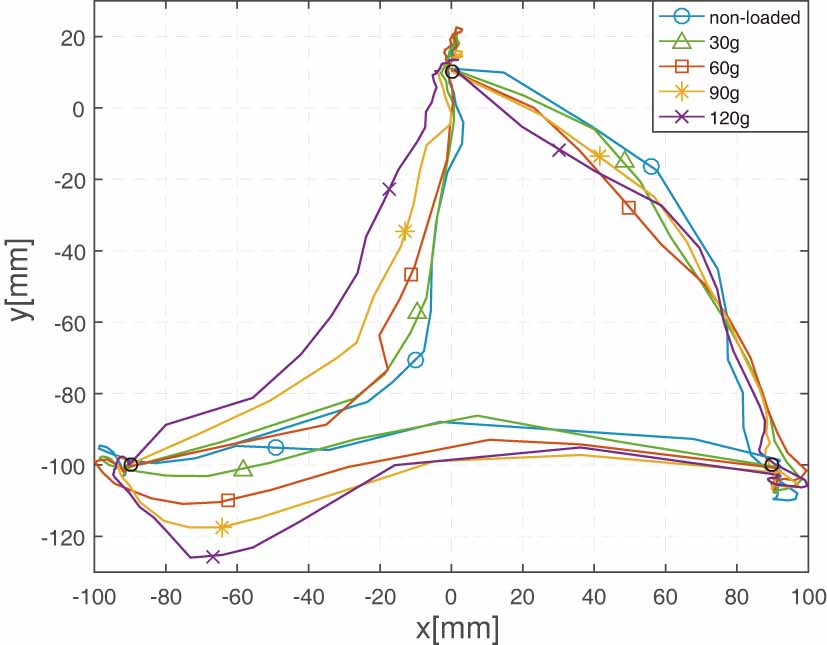}
\caption{The reaching performance in $x-y$ plane of the manipulator in point-to-point reaching test to three targets under a load of 0, 30g, 60g, 90g and 120g respectively.}
\label{reachingperf}
\end{figure}

The reaching performance in point-to-point test is shown in Figure~\ref{reachingperf}, where it can be found out that the three targets are gradually reached during the feedback control process. In this test, the end effector is only set to move in the $x-y$ plane while keeping the Z coordinate constant. From the trajectory records, we can see that the deviations are successfully corrected in the feedback process and do not influence accuracy.

\begin{figure}[!htbp]
\centering
    \includegraphics[width=\columnwidth]{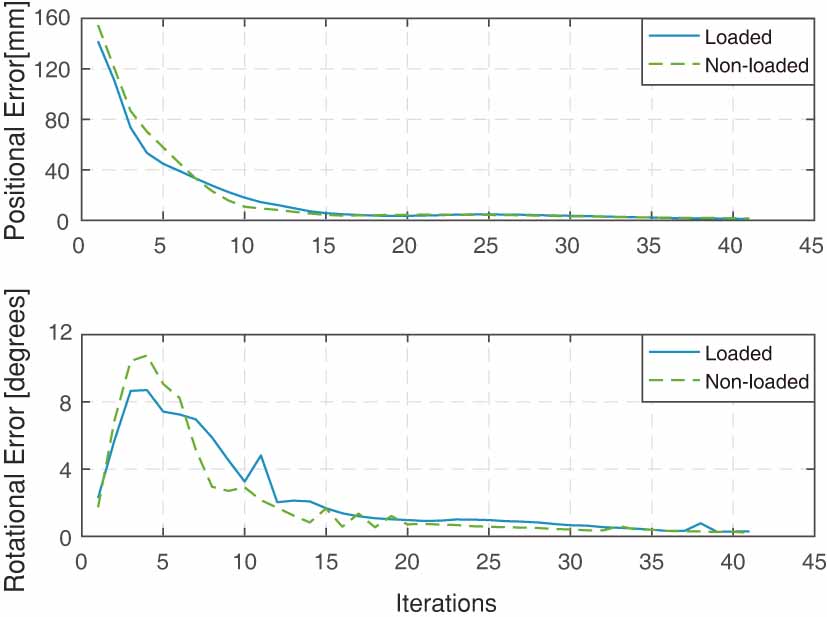}
\caption{The errors in position and direction during the point-to-point experiment are respectively illustrated with no load and 100g load. It is obvious that the errors converge quickly to less than $10mm$ and $2^{\circ}$.}
\label{p2preach}
\end{figure}

Besides, 30 tasks are set by randomly selecting targets in the manipulator workspace. Average error convergence is shown in Figure~\ref{p2preach}. From the positional error figure in Figure~\ref{p2preach}, it can be recognized that the convergence is fast and stable, where the error converges to about $5mm$ in 15 iterations. When the manipulator is under $100g$ load, the convergence rate is slightly slower yet negligible. Thus, it can be concluded that our algorithm adapts well to the load variation in the physical test. In addition, the algorithm is efficient when the error converges to about $5mm$, yet inefficient to further reduce the error.

On the other hand, reaching directions of selected targets are fixed at horizontal forth. It's shown in the rotational error figure in Figure~\ref{p2preach} that the maximum average error is less than $11^{\circ}$, which converges to less than $4^{\circ}$ in 10 iterations. This demonstrates the stability of the control method.

\paragraph{Path tracking.}
Besides, when stability is essential in a task, we can make a trade-off by setting more targets along the path and sacrificing rapidity in order to minimize the error, and this raises requirement of path tracking. We conduct path tracking tests on $x-z$ and $y-z$ vertical planes, finishing the tracking of quadrilateral paths for 5 times. Due to the similarity of results, we only show the performance of test on $x-z$ plane in following figures. Figure~\ref{pathtrack2d} illustrates the path tracking process. The movement is mainly stable, with repetitive fluctuations. The manipulator skews upwards on the upper horizon when it moves to the midpoint and downwards on the lower horizon due to marginal effect.

\begin{figure}[htbp]
\centering
\includegraphics[width=\columnwidth]{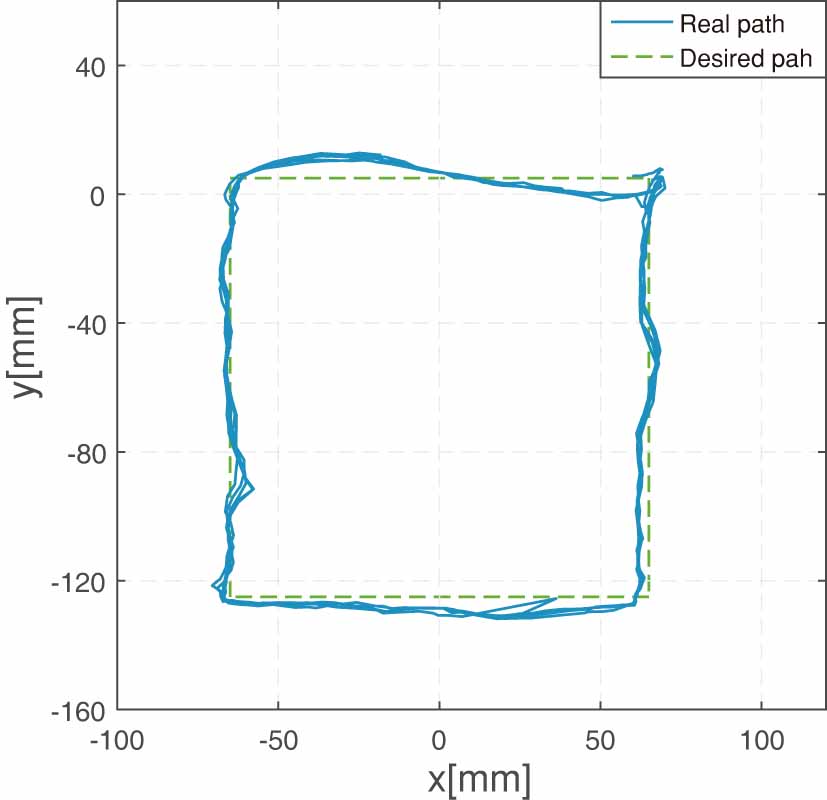}
\caption{Trajectory projection on the x-y plane of in path tracking task, where the dash line is the preset path. The clockwise cycle is repeated 5 times.}
\label{pathtrack2d}
\end{figure}

\begin{figure}[htbp]
\centering
\includegraphics[width=\columnwidth]{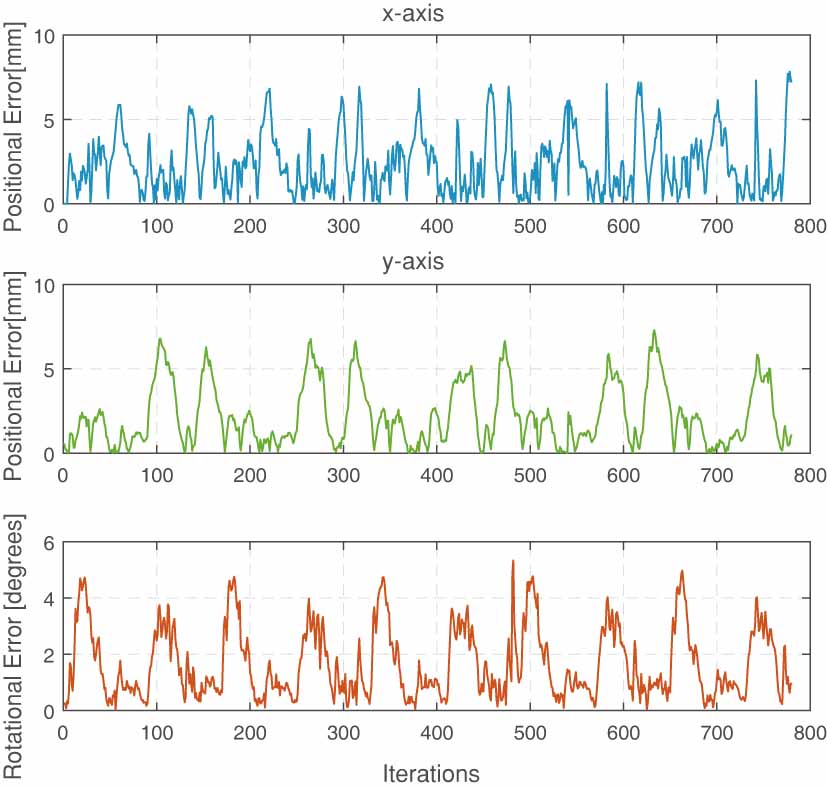}
\caption{Errors in path tracking task. Positional error in x-y plane, on z-axis, and rotational errors are shown.}
\label{pathtrackerror}
\end{figure}

\begin{figure}[htbp]
\centering
\includegraphics[width=0.8\columnwidth]{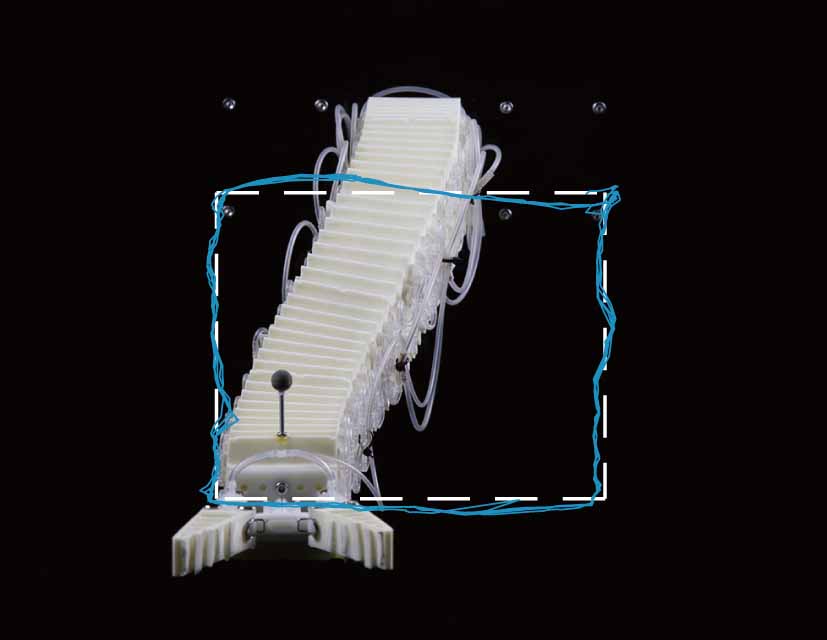}
\caption{A snapshot in path tracking task on the x-y plane. It can be figured out that even the arm curves, the direction of the end effector is stable.}
\label{tracksnapshot}
\end{figure}

\begin{figure*}[htbp]
\centering
\includegraphics[width=\textwidth]{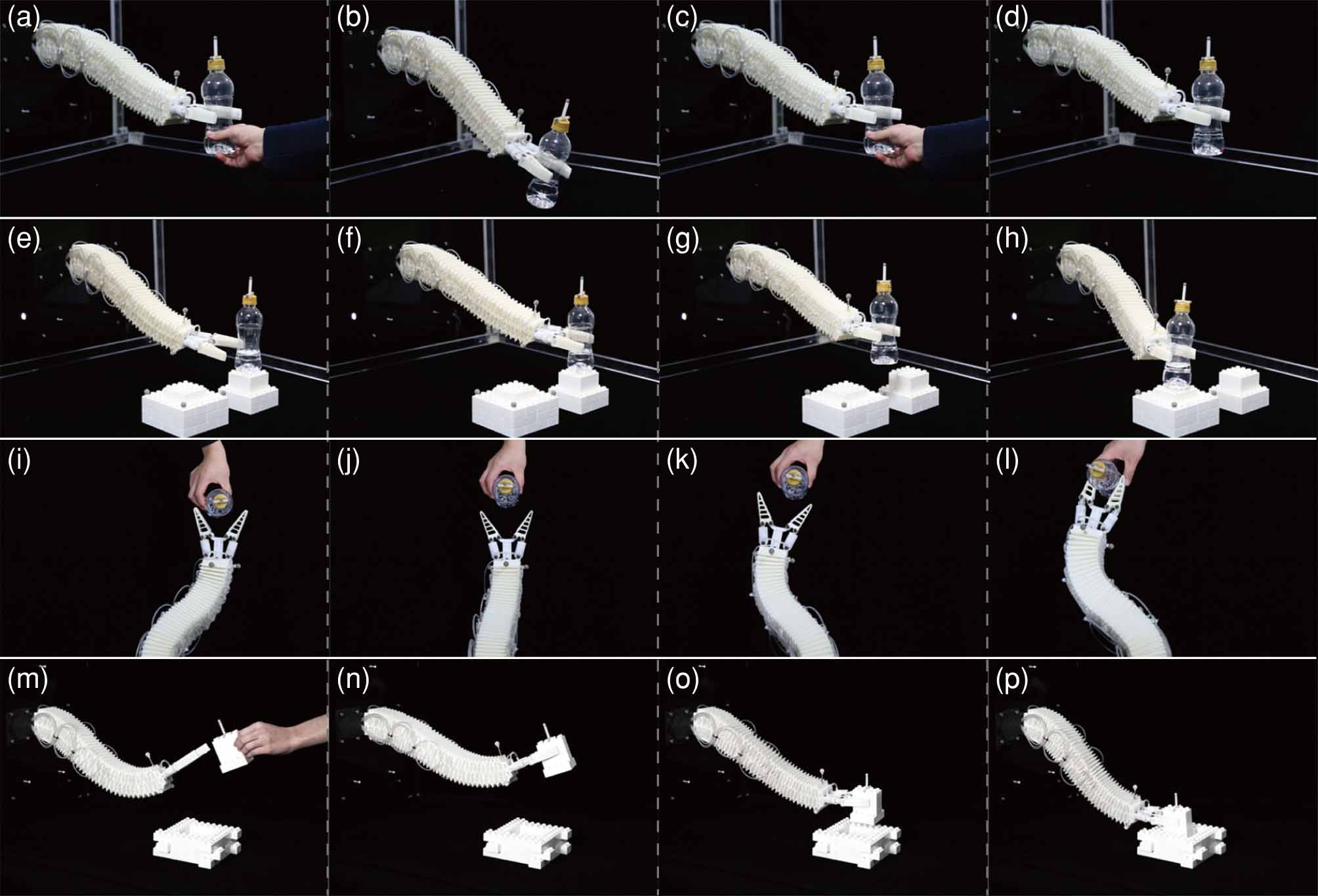}
\caption{This figure shows the process of finishing four tasks using HPN manipulator. In (a)-(d), the end effector falls when loaded with a 120g object, and then returns to the original position using feedback control. In (e)-(h), it moves the object from one position to another. In (i)-(l), it tracks the object’s position and direction, shown from overlooking view. In (m)-(p), it grasps another cube-shaped object from a hand and place the object to the position. }
\label{feedback_realtask}
\end{figure*}

Figure~\ref{pathtrackerror} shows the positional and rotational errors in path tracking task, where the manipulator shows small errors on $z$-axis and direction. It is obvious that fluctuations of error are repetitive, which means the difficulty of reaching different points on the path is not the same. The reason is that in different position the conformity of estimated Jacobian model towards the reality is different: where the estimated is less precise, the algorithm needs more iteration times to converge. In Figure~\ref{tracksnapshot} as a snapshot, the arm bends while keeping a stable direction during the movement. Besides, we test the performance of the feedback controller in several practical scenarios, as described in Figure~\ref{feedback_realtask} (see also Extension 3).

\subsubsection{Limitation.}

\begin{enumerate}[(a)]
\item Firstly, the angle between the tip orientation and the target orientation should not exceed 90 degrees. This is due to the limitation of coordinate transformation, which can be solved by interpolating intermediate targets.

\item Secondly, the angle between the approximated motion direction and the actual motion direction should be no more than 90 degrees in the middle segment, which is the requirement of relatively precise geometric information when calculating the global Jacobian matrix. If the external disturbance is too large, the estimated intermediate information will not be enough, then the control will be unstable.

\item The current single - segment Jacobian matrix is invariant, which affects the accuracy of the model and reduces the control efficiency. More accurate single-segment models can be established by machine learning methods, then higher control efficiency can be achieved.

\end{enumerate}

% The feedback control algorithm is developed on the basis of an estimated model, so the workspace in which it could be used is restricted in the area where the model ensures convergence. Firstly, the angle between the tip orientation and the target orientation should not exceed 90 degrees. This is due to the limitation of coordinate transformation, which can be solved by interpolating intermediate targets. Secondly, the angle between the approximated motion direction and the actual motion direction should be no more than 90 degrees in the middle segment, which is the requirement of relatively precise geometric information when calculating the global Jacobian matrix. If the external disturbance is too large, the estimated intermediate information will not be enough, then directional sensors will be needed in the middle.  On the other side, the algorithm presented in this paper is more portable and operable if only the position and direction of the end effector are considered, because it needs less positional sensors and simpler geometrical model. Hysteresis effect is shown as an exceeding error about 10mm during the physical experiments, which is autonomously corrected thanks to the feedback mechanism, yet it should be examined independently. Due to the limitation of this control platform (with only 1 Hz current feedback frequency), the soft arm moves slowly in the path tracking task. But in our application section, a new platform with higher feedback frequency is applied to achieve better real-time performance.

\subsection{Model-free closed-loop control}

In theory, when using closed-loop control strategy, it is not necessary to model the plant as long as the effectiveness of the actions is known.

In this section, we use the first 4 segments of the experimental platform in figure \ref{control_platform}.We adopt the reinforcement learning algorithm, Q-learning, to learn a model-free control strategy for multi-segment soft manipulators. To implement this method, state, action and reward functions are designed for the 2D horizontal plane point-to-point task and actions are designed for our HPN manipulator prototype. Then we present the main training and estimating algorithm.
\subsubsection{Method}

In this section, we firstly introduce the basic method of q-learning. Then the state and reward functions are designed for the 2D horizontal plane point-to-point task, and actions are designed for a specific manipulator prototype. Lastly, we present the main training and control algorithm.
%In this section, we first introduce the basic method of q-learning, and then state, action and reward functions are designed for the 2D horizontal plane end effector positioning task and actions are designed for a specific manipulator prototype. The last we present the main training and estimating algorithm.
\paragraph{Q-Learning.}

Q-Learning is an iterative process, which learns a strategy that ultimately gives the expected reward of taking a given action in a given state, and it is suitable for learning a control strategy for soft manipulators.

The learned strategy $\pi$ should be able to guide the agent's action at each state, which can be derived from the state-action values after the training process, i.e.
\begin{equation}
	\pi(s)=argmax_aQ(s,a)
\end{equation}
where $s, a$ are state and action, and $Q(s, a)$ is the state-action value.

Before training starts, $Q(s, a)$ returns an (arbitrary) fixed value, chosen by the designer. In each training step, the learning agent perceives its current state from state set $\mathbb{S}$, selects an action from action set $\mathbb{A}$ according to $Q(s, a)$, and then observes a reward $R$ and a new state that may depend on both the previous state and the selected action. The key procedure of Q-Learning algorithm is to evaluate and update the value of $Q(s, a)$ that represents the long-term accumulated reward:

\begin{equation}
\begin{aligned}
Q(s_t, a_t) \gets \alpha [R_t + \gamma \cdot max_aQ(s_{t+1}, a) - Q(s_t, a_t)] + Q(s_t, a_t)
\end{aligned}
\end{equation}

where $s_t$ is the system's state at step $t$, $a_t$ is the action taken in step $t$, $R_t$ is the received reward after taking action $a_t$, and $s_{t+1}$ is the state in step $t+1$ after transfer. $\alpha$ and $\gamma$ are parameters, $\alpha$ is learning rate, and $\gamma$ is called discount factor.

\paragraph{Definitions.}

In this chapter, we introduce the definitions of state, action and reward for the Q-Learning algorithm.

\begin{figure}[htbp]
    \centering
    \includegraphics[width=0.6\columnwidth]{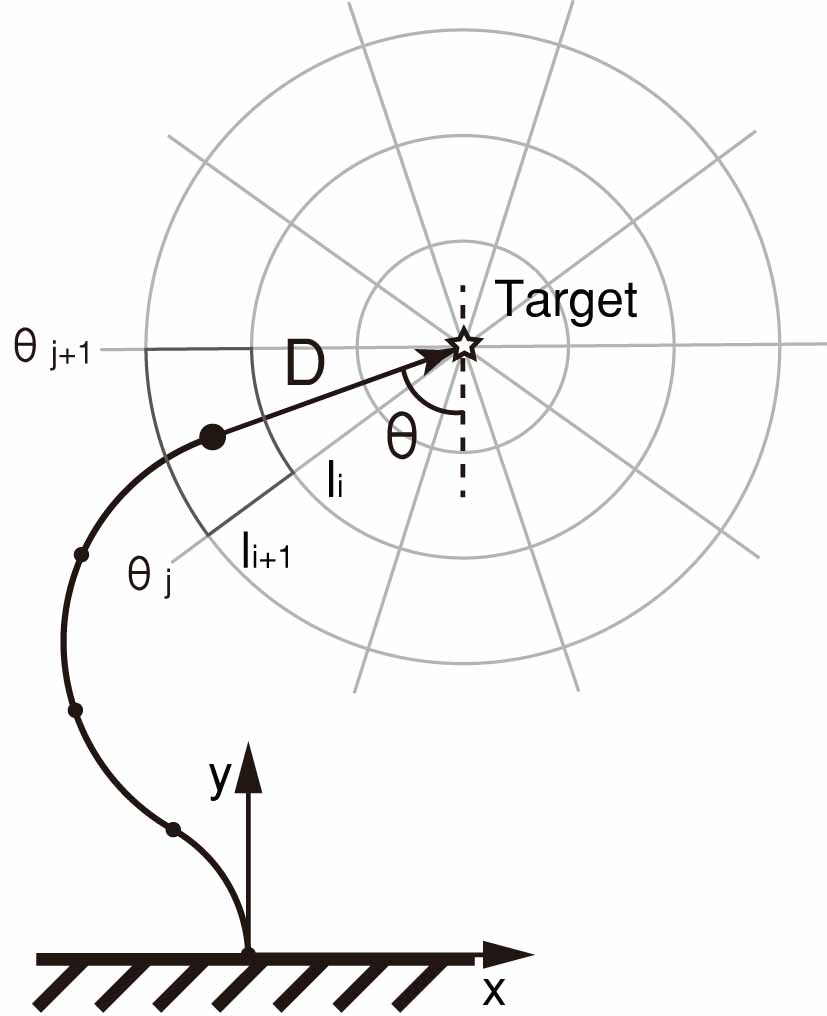}
    \caption{The definition of state through ($l$, $\theta$), where $l$ is length of vector $\vec{D}$, and $\theta$ is clockwise rotation angle from negative y-axis to $\vec{D}$. We evenly divide the range of $l$ and $\theta$ into M and N intervals separately. Each interval pair represents a certain state, i.e. when the length of vector $\vec D$ is between  ($l_i$, $l_{i+1}$) and the angle is between ($\theta_j$, $\theta_{j+1}$), the system is defined to be at state ($i$, $j$). An example of state is marked in the figure.}
    \label{statedef}
\end{figure}

For the 2D point-to-point task, state is defined by relative position of the manipulator's end effector to the target, denoted as $\vec{D}$. Specifically, it is determined by $(l,~\theta)$ where $l$ is length of $\vec{D}$, and $\theta$ is clockwise rotation angle from y-axis to $\vec{D}$, as shown in Figure~\ref{statedef}. To discretize the continuous state space, we partition $(l,~\theta)$ into intervals,  ${L_i = [l_i, l_{i+1}], i = 1,2,\cdots,m}$ and ${\Theta_j = [\theta_j, \theta_{j+1}], j = 1,2,\cdots,n}$, which are uniformly distributed in the available range of the manipulator. Thus, there are $m\cdot n$ states in total. When a certain state $(l^*,\theta^*)$ lies in $(l_i,l_{i+1})$,$({\theta}_j,{\theta}_{j+1})$, the corresponding state will be $(i, j)$, $1 \leq i \leq m, 1 \leq j \leq n$. As shown in Figure~\ref{statedef}, when the state gets close to the target, the partition granularity will be high, which guarantees control accuracy.

\begin{figure}[htbp]
    \centering
    \includegraphics[width=0.95\columnwidth]{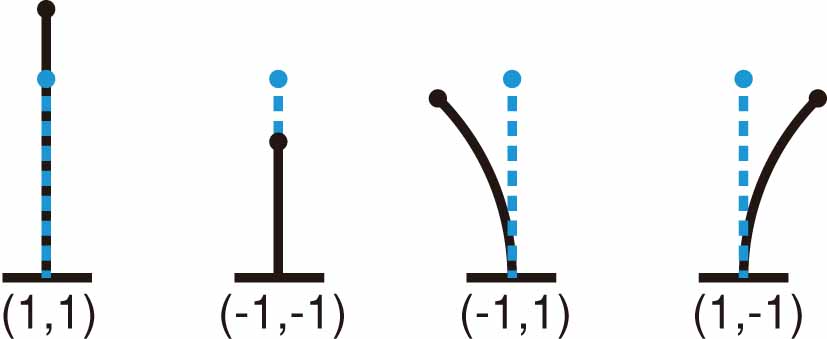}
    \caption{Four unit movements for a single segment which are defined as actions. Number 1 represents inflation at the side and -1 represents deflation. Action (1,1) is to inflate both sides, which leads to elongation, and action (-1,-1) is to deflate both sides which leads to shortening. (-1,1) is to deflate the left side and inflate the right side of the airbags, which leads to bending towards left side, and action (1,-1) is the contrary movement. In the figure, blue dashed line represents previous pose and black solid line represents the current pose after the action. All the possible pose of the manipulator can be achieved by the combination of these four basic actions.}
    \label{actiondef}
\end{figure}

Actions are represented as the manipulator end effector's position variation. For most multi-segment soft manipulators, the movement of whole manipulator can be regarded as the combination of each segment's motion. So, we can find a proper basic motion set of a segment, as the definition of actions. Generally, there are totally $k \cdot s$ actions when the manipulator has $k$ segments, each with $s$ basic motions. For instance, for our manipulator, in 2D workspace, $s$ equals 4 when segment action is defined as moving towards front, back, left and right (as shown in Figure~\ref{actiondef}). Consequently, the entire workspace of manipulator can be achieved by a sequence of these actions.

For 2D movement, two groups of airbags on one side of each segment are inflated together. Thus, the movement of one segment is actuated by different pressure combination on both sides. So actions, representing basic movement, can be defined as $a(p_l^i, p_r^i)$ where $p_l^i, p_r^i$ signs the motion (inflating or deflating) for the pressure units on left and right side of segment $i$ respectively. Specifically, as shown in Figure~\ref{actiondef}, when $p_l^i$ and $p_r^i$ are both set to 1 or -1, the segment will elongate or shorten accordingly. And when $p_l^i$ and $p_r^i$ are set to 1 and -1, the left side will be inflated and the right side will be deflated, causing the segment to bend toward right.%, and vice versa.

\begin{figure}[htbp]
    \centering
    \includegraphics[width=0.6\columnwidth]{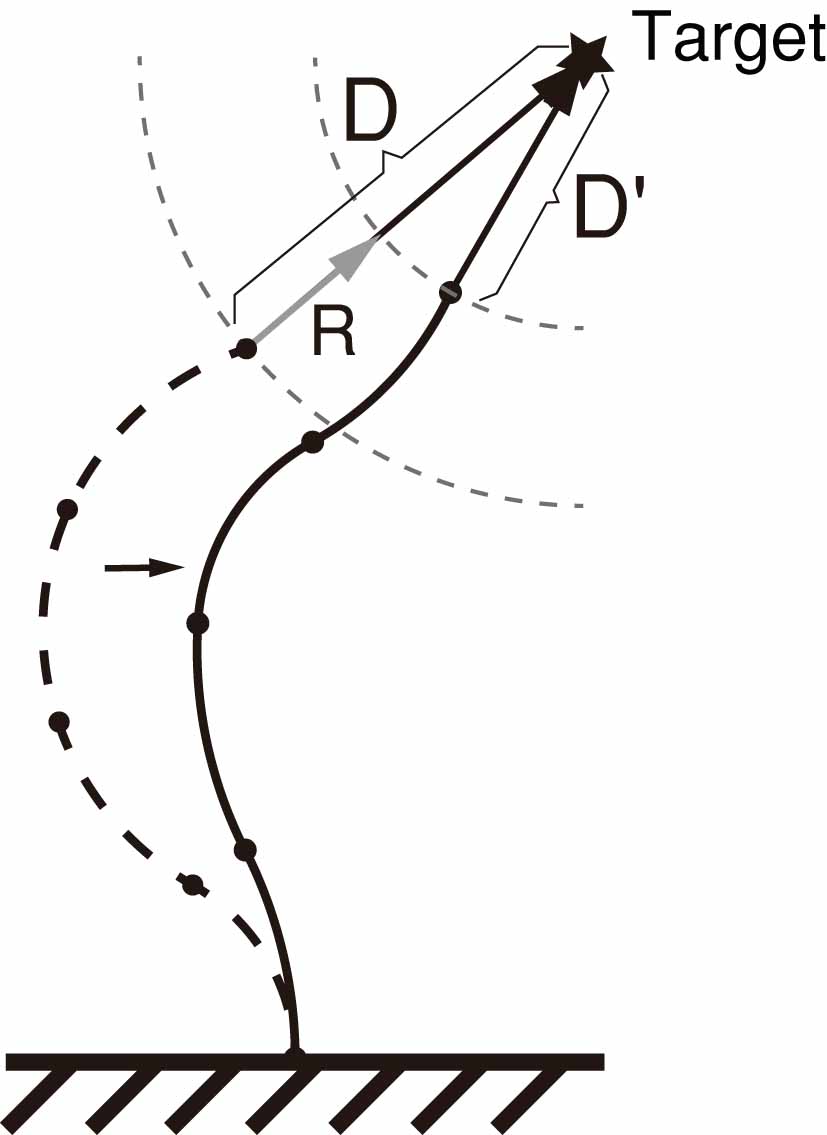}
    \caption{The design of reward function. Potted line represents the manipulator pose before executing an action and solid line represents the pose after that. In this figure, reward equals $|\vec R|$. If the end effector moves closer to the target, the reward is positive; otherwise, the reward is negative.}
 	\label{rewarddef}
\end{figure}

Reward is the evaluation of how much a certain action from a state contributes to reaching the target. Specifically, $R$ can be defined as distance variation relative to the target:

\begin{equation}
R = |\vec D| - |\vec D'|
\end{equation}

where $|\vec D|$, $|\vec D'|$ are distance between end effector and target before and after the action as shown in figure \ref{rewarddef}. When the manipulator moves towards the target, $R$ is positive, and the closer it gets, the larger the reward is; otherwise, the reward is negative. According to $R$, $Q(s, a) \in \textit{\textbf{Q}}$ gradually converges to the reasonable value during the learning process, where $\textit{\textbf{Q}} \in \mathbb{R}^{mn\times 4k}$ represents the state-action value matrix.

\paragraph{Control algorithm.}

The main training algorithm is designed as follows (shown in Algorithm \ref{algo_Q}), whose purpose is to learn a moving strategy. The training is conducted during repeated process of moving towards randomly selected targets, and it finishes after achieving an expected level of proficiency, denoted as a certain proportion of marked states.

\begin{algorithm}[b]
  \caption{Training Algorithm}
  \footnotesize
  \begin{algorithmic}[1]
  \Function{Train\_Q}{$\mathbb{S}, \mathbb{A}, \mathbb{W}$} \Comment{$\mathbb{W}$ is manipulator workspace}
  \State $\mathbb{S} \gets$ Refine$(\mathbb{W}, \mathbb{S})$
  \Repeat
  	\State $x_t \gets$ SelectTarget($\mathbb{W}$)
  	\State [$\vec{D}, s$] $\leftarrow$ GetState($x_t, \mathbb{S}$)
  	\While{$|\vec{D}| > $threshold}
      \State $a \gets \epsilon$-greedy($s, \textit{\textbf{Q}}$)
      \State Move($a$)
  	\State [$\vec{D'}, s'$] $\leftarrow$ GetState($x_t, \mathbb{S}$)
	  \State $R = |\vec D| - |\vec D'|$
  	  \State $Q(s, a) \gets Q(s, a) + \alpha [R + \gamma \cdot max_aQ(s', a) - Q(s, a)]$
  	  \If{$Q(s, a) > 0$}
  	    \State Mark($s$)
  	    \State UpdateNeighbor($s$)
  	  %\Else
  	  %  \State Unmark($s$)
  	  \EndIf
  	  \If{There is no new mark in recent T iterations}
  	    \State break
  	  \EndIf
  	\State [$\vec{D}, s$] $\leftarrow$ GetState($x_t, \mathbb{S}$)
    \EndWhile
  \Until MarkedProportion($\mathbb{S}) > P\%$
  \For{each $s \in \mathbb{S}, a \in \mathbb{A}$}
  	\State $\pi(s)$ = $argmax_aQ(s, a)$
  \EndFor
  \\
  ~~~~\Return $\pi$
  \EndFunction
  \end{algorithmic}
  \label{algo_Q}
\end{algorithm}

For the effectiveness and completeness of the learned moving policy, in essence, we need to make sure that for each available state s, there exists an action a s.t. $Q(s, a) > 0$. During the training process, when a state-action pair with a positive value is found, the state is given a mark. A marked state is definitely an available state. When training at a given target, the number of marked states Z grows fast at first, and then tends to be steady. If it does not increase in T steps, the target will be changed to another appropriate one.

Specifically, targets are selected randomly from available workspace $\mathbb{W}$, and vector $\vec{D}$ from end effector to target and corresponding state $s$ are obtained. Before each iteration, $|\vec{D}|$ is compared with a preset threshold value in order to decide if end effector has reached the position within acceptable accuracy. During iteration, an action is selected according to the $\epsilon$-greedy strategy, which means the currently known best action (with largest value of $Q(s, a)$) will be selected with a certain probability $1-\epsilon$, and a random action will be selected with probability $\epsilon$. The manipulator then performs the selected action and the new position are used to calculate $R$, which is used to update $Q(s, a)$. If the current state satisfies the mark condition ($Q(s, a) > 0$, means there exists positive action towards target from this state), then mark it. Then the adjacent unmarked states are assigned with
the average value of its neighbors. After that, we judge if it is time to choose a new target for training by checking if there are new marks in recent T iterations. When the outer iteration ends, a control strategy $\pi$ is generated by selecting the action with largest value of $Q(s, a)$ under each state. 

\begin{figure}[!htbp]
    \centering
    \includegraphics[width=0.8\columnwidth]{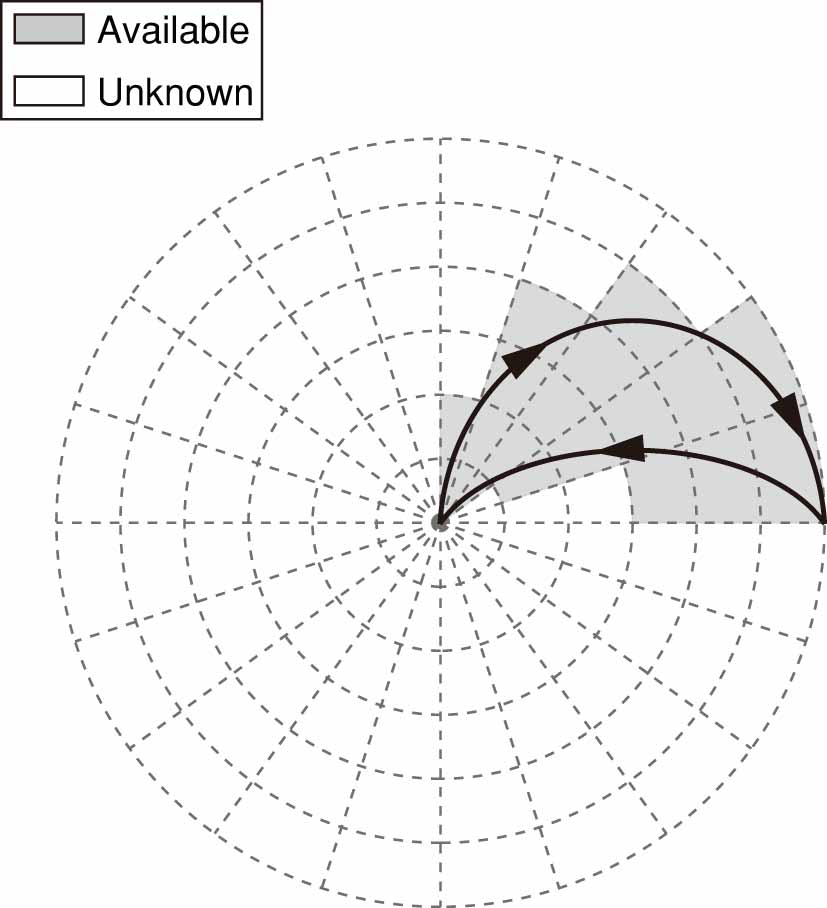}
    \caption{Searching for available states. Let the state partition graph move around the periphery of the workspace, and record all the states whose corresponding grid cell could have an intersection with it. In the schematic figure above, the black and grey states are known to be available while the white ones are not known currently.}
    \label{searchstates}
\end{figure}

For the state refining in line 2, since the definition of states only depends on the relative position of the manipulator's end effector and the target, rather than their actual position, the state partition graph is identical for any target. The center of the graph is an imaginary target, and the grid cell where the manipulator's tip is in determines the system's state. To ensure the completeness of the state set, the radius of the maximum circle is set to be the maximum distance between two points in workspace. Considering the specific shape of the workspace, only part of the states are available and others may never be reached. We propose an approach for the estimation of maximum available states, shown in Figure~\ref{searchstates}. Let the center of the state partition graph move along the periphery of the workspace, and at the same time, record the states whose corresponding grid cell intersects or is contained in the workspace. In this way, we are able to tell all available states from unavailable ones and their proportion in all states.

\begin{figure}[htbp]
    \centering
    \includegraphics[width=0.8\columnwidth]{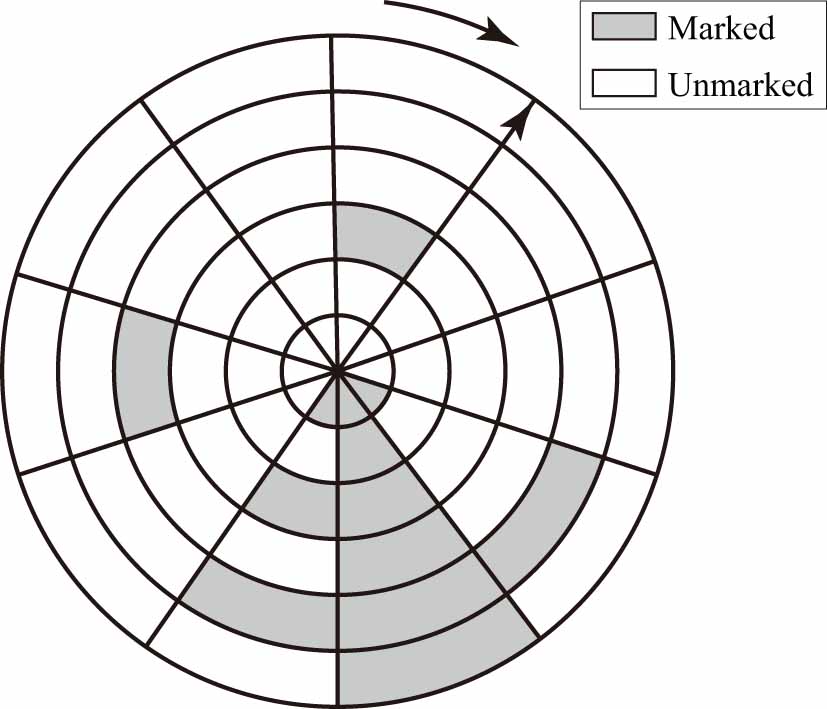}
    \caption{The state partition graph at an imaginary target. In the training process, a state $s$ will be marked when positive value of $Q(s, a)$ is found, and unmarked neighbor states (shown as neighbor areas, one state has eight neighbors) will be updated. Besides, since the states are independent of specific targets, when a new training target is selected, the partly trained graph retains and following training will build on it.}
    \label{markstates}
\end{figure}

Besides,  it is possible that some available states are not marked in the training process, and some elements of $Q$ will remain the initial random values. In line 13, we handle the problem by approximation: if a state $s$ is not marked, for every action $a$, $Q(s,a)$ is set as the average of its neighbor states. As shown in Figure~\ref{markstates}, when a state is newly marked, all the neighbor unmarked states will be updated to the average value of neighbor marked states, for example, an unmarked state will update to the marked state value to that of the only one marked state, or average of two marked states' values.

After the training process, the control strategy will be tested in workspace, where the main process is similar to the previous training, as shown in Algorithm \ref{algo_q_control}. Specifically, only the update of $Q(s, a)$ and marking of states are no longer conducted in the test.

\begin{algorithm}
  \caption{Control Algorithm}
  \footnotesize
  \begin{algorithmic}[1]
  \Function{Control\_Q}{$\mathbb{S}, \mathbb{A}, x_t, \pi$}
  	\State [$\vec{D}, s$] $\leftarrow$ GetState($x_t, \mathbb{S}$)
  	\While{$|\vec{D}| > $threshold}
      \State $a \gets$ BestAction($\pi(s)$)
      \State Move($a$)
  	  \State [$\vec{D}, s$] $\leftarrow$ GetState($x_t, \mathbb{S}$)
    \EndWhile
  \EndFunction
  \end{algorithmic}
  \label{algo_q_control}
\end{algorithm}

\subsubsection{Experiment}
In this part, we conducted point-to-point experiments to verify the effectiveness of q-learning algorithm.
 Considering the reachable range of the manipulator, we partition the distance to the target $l$ from $0mm$ to $480mm$ evenly into 16 intervals, and the angle $\theta$ from $0^{\circ}$ to $360^{\circ}$ evenly into 18 intervals. Thus there are altogether 288 states. The expected proportion of marked states $P$ is set as $50\%$. We adopted a subset of the action set, only including the elongating and bending actions. Since the manipulator consists of 4 segments, there are totally 12 actions.

\begin{figure}[!htbp]
    \centering
    \includegraphics[width=0.9\columnwidth]{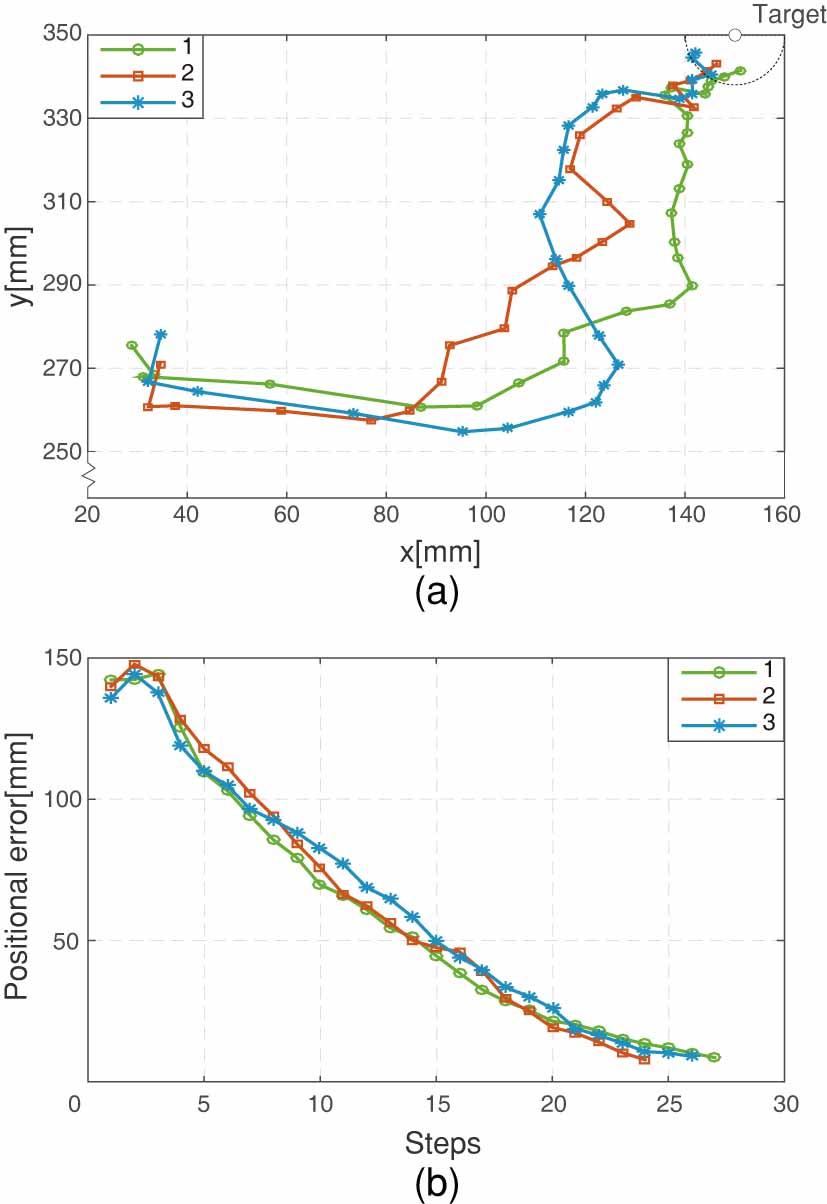}
    \caption{The fold lines in (a) represent the ordered intermediate coordinates of the manipulator's tip during the process of reaching the target. The four lines are slightly different because it has the possibility of 0.1 to conduct a random action in each step. (b) is the error graph, which demonstrates that as the manipulator's tip approaches the target [150, 350], the error gradually reduces and finally drops to less than $10mm$. There are few pointless actions, illustrating that most executed actions according to the strategy are rational and effective.}
    \label{effect}
\end{figure}

The state-action values converge after around 1000 outer iterations during learning process. After the learning process, we test the effectiveness of the learned strategy. 
We randomly choose a target within the reachable range of the manipulator and let the manipulator move under the $\epsilon$-greedy strategy to reach the target for 4 times, where $\epsilon$ is set to 0.1. We expect it to reach the targets at a precision higher than $10mm$. Figure~\ref{effect} shows that the distance between the tip of the manipulator and the target generally decreases as the manipulator takes actions, and finally the expected precision is reached. The increase of error at the first few step is probably because our manipulator would first elongate before bending when the airbags start to contact with the honeycomb wall. These experiment results attest to the effectiveness of our control strategy.

\subsubsection{Limitation.}

\begin{enumerate}[(a)]
\item The current action of q-learning is to inflate or deflate a set of airbags. Only one set of airbags can be operated in each cycle, which is inefficient. Adding more action can make it faster, but the training time will be increased as the expense. When actions can be composed into any combination, the efficiency of well-trained q-learning should theoretically be better than that of model-based methods. In fact, reinforcement learning can be used to accomplish complex tasks, which is obviously overkilled here just as an exploration of model-free controls.

\item In this section, only position control is implemented. In fact, the same control method can be used to control position and direction.

\item In addition, training data is not fully utilized here. In fact, multiple virtual targets can be defined for a single action, so that all elements of the column of Q matrix corresponding to the action can be calculated. So a relatively accurate Q matrix can be calculated after each action theoretically. In fact, the reinforcement learning controller we use in the comparative experiment section later realizes this idea of data reuse. Because very little data is required to learn the Q matrix, the controller can be updated in real time, so as to better deal with larger external disturbances.

\end{enumerate}

% In order to improve accuracy, the step length is often set small enough and the manipulator needs to finish a large number of steps before reaching a target. This limitation can make the task time-consuming, especially for control systems with low reaction frequency. One hopeful solution for this problem may be to include steps with different levels of pressure variations as actions. 
% In this section, only the basic position control is implemented. Actually the same control method could be applied to control both position and direction. In theory this needs more training data, but actually for a single action, many virtual targets could be defined so that all the elements of the Q matrix corresponding to that action could be calculated. So in theory, after every action is performed once, a relatively accurate Q matrix could be obtained. Actually, this method is also applied in later experiments.

\subsection{Comparative experiment}
We implement several different control methods: two level model based control, estimated model control and q-learning control, which base on idea of accurately modeling, idea of feedback without accurate model and idea that learning the strategy without modeling, respectively.  To study the characteristic and proper application condition for the different control methods, and to find the control method that can make best use of the advantages of soft manipulators in interactive tasks, we compare their ability to perform tasks under different external disturbances.

\subsubsection{Experiment settings.}
In order to compare the capability of the control methods, we perform the experiments on the same control platform, as shown in Figure \ref{control_platform}. The same experimental configuration as that used in the model based control method experiment is used. Estimated model control method and Q-learning control method are implemented on a 2 dimensional 3 segment HPN arm, and the information of the tip orientation is added to the reward in Q-learning to implement the orientation control.

We perform point-to-point tasks in the experiment. If the positional error is less than 15mm and the rotational error is less than 15 degrees, the task is considered to be completed. We introduce different disturbances in the experiments. For the control group, no disturbance is applied. For the experimental group with small disturbances, constant lateral forces of 0.75N and 1.5N are applied respectively. For the experimental group with large disturbances, the connections between the two groups of the airbags and the controller are exchanged. And the exchange is applied at the middle segment and the base segment respectively. The two experimental groups simulate the situations where the arm is under constant external force (for example the constant force generated in interactive task and the arm’s gravity) and the situations where the actuation mapping is changed by huge disturbances (for example when the arm bends more than 90 degrees).
	
In preliminary experiment, we discover that when the actuating pressure is kept constant and a small lateral force (0.75N) is applied at the tip, the deviation of the tip is $31mm$ on average, which is much larger than the specified accuracy requirement. To some extent, this prove that open loop control based on accurate model would be unacceptably inaccurate. So in this section we would mainly compare the three close loop control method.

The tip of the arm is controlled to follow the marking point using model-based feedback control method, and the positions and directions of 20 roughly evenly distributed reachable states are recorded. The arm is controlled to reach the states using the three control methods respectively. The reaching time is recorded to evaluate the control efficiency, and the reaching result (reached or not reached) is recorded to calculate the success rate. 100s is set as expire time, which is fairly enough for our algorithms to converge in most cases. If the target can not be reached in 100 seconds, the task is considered as failed.

\subsubsection{Results and discussion.}

\begin{table*}[!htbp]
\centering 
\begin{threeparttable}

\caption{Results of comparative experiment}
\label{lable of Comparative experiment}

\begin{tabular}{ccccccccccc}
\hline
&\multicolumn{2}{c}{free space}&\multicolumn{4}{c}{external force}&\multicolumn{4}{c}{reverse group}\\
\cmidrule(lr){2-3} \cmidrule(lr){4-7}\cmidrule(lr){8-11}
&\multicolumn{2}{c}{0g}&\multicolumn{2}{c}{75g}&\multicolumn{2}{c}{150g}&\multicolumn{2}{c}{middle}&\multicolumn{2}{c}{root}\\
\cmidrule(lr){2-3} \cmidrule(lr){4-5}\cmidrule(lr){6-7}\cmidrule(lr){8-9}\cmidrule(lr){10-11}
&time&rate&time&rate&time&rate&time&rate&time&rate\\
\hline
model-base&7.6/1\tnote{1}&1&13.1/1&0.6&13.8/1&0.55&26.1&0.35&N/A&0\\
estimated-model&12.2/1.61&1&12.3/0.94&0.7&10.8/0.78&0.55&22.5&0.3&6.2&0.1\\
Q-learning&48.0/6.31&1&53.3/4.07&0.65&40.0/2.89&0.5&52&1&49&1\\

\hline
\end{tabular}

\begin{tablenotes}
\footnotesize
    \item[1] The former number represents the real time, while the latter one represents the relative time of each method relative to the model-based method. 

\end{tablenotes}
\end{threeparttable}
\end{table*}

\begin{figure*}[htbp]
\centering
\includegraphics[width=\textwidth]{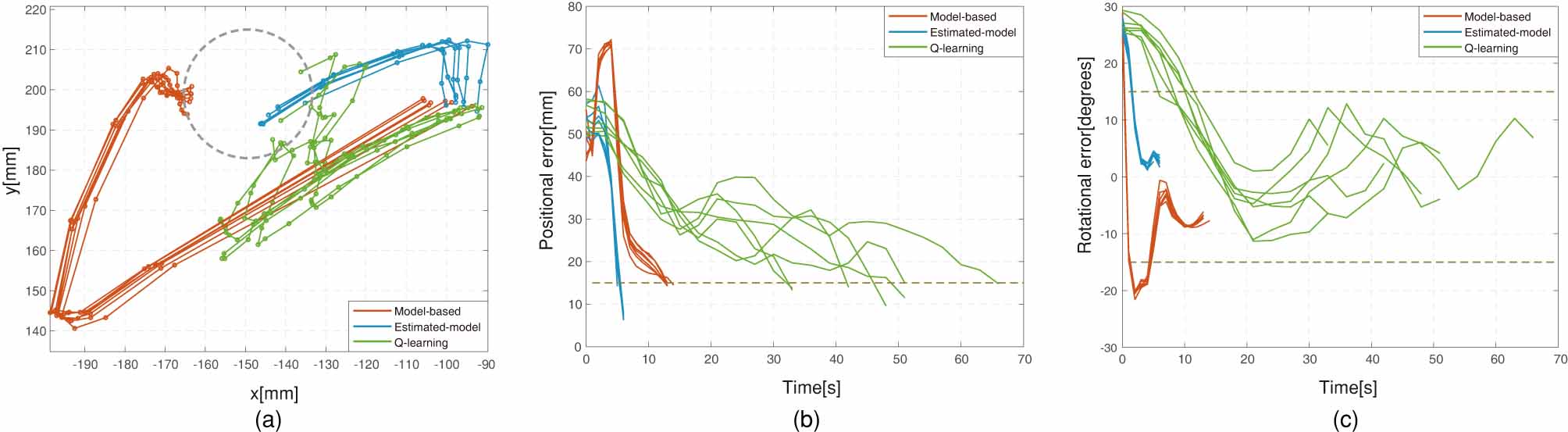}
\caption{The result of comparative experiments. (a)shows the trajectory of three algorithms arriving at the same point repeatedly under disturbances. The circle represents the scheduled arrival area. (b)shows the change in distance from the target point during the movement, and the abscissa is the running time of the algorithm.The dotted line represents the scheduled arrival area. (c)shows the error of angle during the motion.}
\label{control_compare}
\end{figure*}

In free space, all the specified states could be reached using all three control methods. In this case, the model based control method has highest efficiency.

When large disturbances that change the actuation mapping are applied, we mainly care about the success rate. All the model based method are affected greatly. While for the strategy based method, due to its online learning, it could recover in a short time (even after performing each action for one time). Even exchanging the airbags has no effect on it, it could still reach all the states and its reaching time is not apparently influenced. It’s worth mentioning that, for each experimental group, the Q-learning method is trained without any prior knowledge about the disturbances. In the result of control group and experimental group that exchanges the airbags, the time to reach the first 5 states is 61.6 seconds on average, while the time to reach the next 15 states is 45.8s on average, which demonstrates the rapid learning ability of Q-learning control method.

 The lateral disturbances would make the model deviates from the actual situation and decrease efficiency. Among the control methods, the reaching time of Q-learning control method is relatively constant and could represent the actual control difficulty. So we choose the time of Q-learning as a unit and calculate the relative reaching time with respect to it to observe the performance changes of the control methods. It could be found out that the relative reaching time of estimated model control method is roughly constant around 0.25. While the relative reaching time of model based control method rise from 0.16 to 0.34. It could be inferred from Figure \ref{control_compare} that, because the step size is large, the path of model based control method first deviate largely and then return to move towards the target under feedback. In this case, the model largely deviates from the actual situation, so if a large step size is used, the tip may move away from the target. Figure \ref{control_compare} (b) (c)  shows the positional error and rotational error while reaching the target. It can be inferred that the model based control feedback strategy is radical and causes large over-adjustment under disturbances. Estimated model control method may also cause over-adjustment, but it has a small step length so it could adjust in time without generating large deviations. In addition, it can be observed that due to the inherent randomness of Q-learning control method, the paths of reaching the same point is not the same and various from each other. While the path of model based control method has a higher repetition rate.

In conclusion, under small disturbances or zero disturbance, model based control strategy with feedback  performs the best, for it is fastest and most accurate. It could be concluded that the reaching efficiency is positively correlated with the modelling accuracy.

Generally speaking, the more accurate the model is, the better, even when external disturbances are applied. But the larger the disturbances are, the smaller the impact of the accuracy of the model is.  If the cost of modeling is not considered, then more accurate model should be chosen, or else with a large cost of modeling (as the case of soft robots), an estimated model could be used with task space feedback to achieve decent performance. Specifically, if both of the controllers using accurate model and estimated model iterates with small step size, they should have similar performance. In order to take advantages of accurate model, the step size should be large. In our experiment, with the model based control method using large step size, and when there is no disturbance, the reaching time of estimated model is 1.6 times longer than that of accurate model. Because the soft robotics is at its starting stage, most soft robots move slowly, and the 1.6 times difference is not large. But the larger the step size is, the more impact it will suffer when disturbances are applied. And its performance may even worse than the estimated model. In interactive tasks, disturbances are usually applied. So in this case using a small step size may result in a performance similar to that of using accurate model, but it is more simple to implement. For preliminary study of the control of soft arms in interactive applications, this method should be a good choice.  In the future, with the development of proprioceptive sensors and modeling technology, the control should be better and better.

When the disturbances are large, controller without model using real time learning could be used. Though model based control methods could also update its model, it’s hard to implement online update. And learning based method also has a strong cross-platform adapting capability.

\subsection{Control Conclusion}

Kinematics of the traditional robot, called rigid body kinematics, will not be affected when force is considered. Soft robot can be modeled without considering the force, but this model will change tremendously after taking force into account. It’s acceptable if the deviation is limited. But in most cases, this deviation is too large to be accepted.

Therefore, we come up with a bold hypothesis: traditional kinematics does not apply to soft robots and a new theory is needed. Instead of accurately modeling their own bodies, creatures always use visual and tactile feedback to complete complex tasks, which may provide some inspirations for soft robots to have better performances in unstructured environments.

This paper begins with the difference between the modeling of soft arms and the modeling of rigid arms to explore the control problem of the soft arm. Three new control methods are presented: accurate model, estimated model and model-free method. The characteristics and application of these different methods are obtained through comparative experiments. 

We point out that only in an ideal environment is it possible for a soft robot to be modeled accurately and achieve open-loop control in task space like a rigid robot. Task space feedback controller is indispensable in non-ideal environment and experiment at the end of this section proved this judgment to some extent. Our judgment may not be totally correct and just to consider it a prompt for open threads. What matters is the fundamental question - can a soft arm be accurately modeled? The answer to this question may inspire new control methods and avoid detours. 

However, External sensors, such as the Motion Capture System we use, are often essential for the task space feedback controller, and it may limit the application of the system. Traditional rigid robots do not need sensors in the task space for their control, so there has been little demand of them in robotics. With the development of soft robots, the demand for portable, reliable and cheap task space sensors will continue to grow.

\subsubsection{Limitation}

\begin{enumerate}[(a)]
\item At present, what we have done is the simplest motion control without force control or dynamic control.

\item All the control methods in this paper need to use sensors that can feedback the task space information, but there are some limitations in using such external sensors. For example, we use MCS to feedback task space information, which is expensive and poor in portability (or mobility) -- MCS need to be set up in the peripheral space in advance. 

\item Only the position control of the end-effector is considered in our work. However, for the planning tasks that require collision avoidance, attitude control of the arm is required. To some extent, attitude control can be achieved by adding reflective markers in the middle segment. What’s more, the state of the arm is estimated based on the position of the tip point in our model-less control. It means that if the pose of the middle segment is known, the estimated Jacobian model will be more accurate and the overall efficiency of the algorithm will be improved.
\item The current control method based on accurate model and estimation model is feedback control with fixed step size. Performance of the feedback control method can be better by adopting the method of adjusting the step size dynamically, such as PID method.

\end{enumerate}

% \subsubsection{Control limitation.}
% At present, what we have done is the simplest motion control without force control or dynamic control. And only the position control of the end-effector is considered in our work. However, for the planning tasks that require collision avoidance, attitude control of the arm is required. To some extent, attitude control can be achieved by adding reflective markers in the middle segment. What’s more, the state of the arm is estimated based on the position of the tip point in our estimated model control. It means that if the pose of the middle segment is known, the estimated Jacobian model will be more accurate and the overall efficiency of the algorithm will be improved. 

% There are some limitations of using MCS as sensors. But the main point of our work is to prove the feasibility of this technical system. MCS can then be replaced by monocular or TOF cameras. 

\section{Application}

When soft robots interact with the environment, their behavior resulting from passive degrees of freedom is highly diverse. The behavior is much more complex than the actuation which tend to contribute to interaction tasks in turn. But why soft robots exhibit such property? And what it contributes to the development of robots?  How should we utilize this property?

The objective of this section is to explain and demonstrate the features of soft manipulators when interacting with environments. 

\subsection{Explanation of simplification}
\begin{figure}[htbp]
    \centering
    \includegraphics[width=0.95\columnwidth]{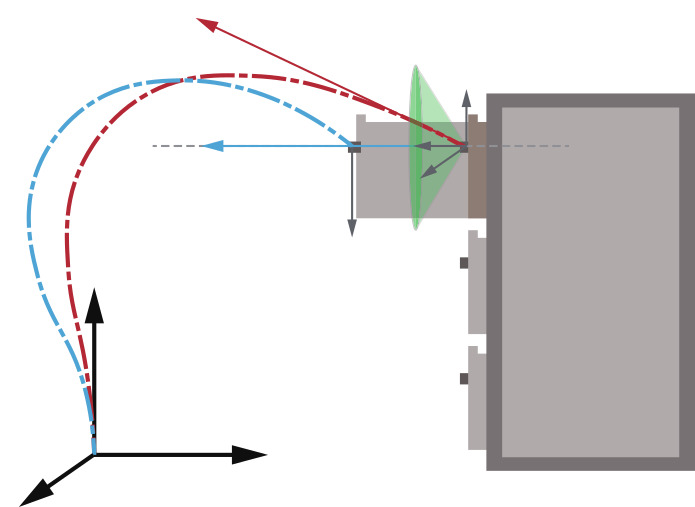}
    \caption{Schematic of the drawer opening task. The red and blue curves are the initial state and the execution state of the arm, respectively. The red arrow is planned motion, indicating that the arm will move in this direction when there is no external force. The blue arrow is the real motion, which represents the actual direction of motion within the drawer constraints. The green cone represents the set of directions in which the soft arm can open the drawer. F is the reaction force to the soft arm due to the inconsistencies between planned motion and the drawer's natural constraints.}
    \label{open_drawer_demo}
\end{figure}

In order to clearly explain the mechanism and characteristics of the interaction between the soft manipulator and the environment, we choose an interaction task, opening the drawer, as shown in Figure \ref{open_drawer_demo}, for analysis.

Many robotic tasks have a set of orthogonal reference frames in which the task specification is very easy and intuitive. Such frames are called task frames or compliance frames. An interaction tasks can be specified by assigning a desired force / torque or a desired linear / angular velocity along / about each of the frame axes. The desired quantities are known as artificial constraints because they are imposed by the controller; these constraints, in the case of rigid contact, are complementary to those imposed by the environment, known as natural constraints.

For the rigid manipulator, this interaction task of opening the drawer in task frame $S_t$ can be described as:

Desired forces and torques:

\begin{enumerate}

\item Zero forces along the $x_t$ and $ z_t$ axes 
\item Zero torques about the $x_t$, $y_t$, and $z_t$ axes
\end{enumerate}

Desired velocities:

\begin{enumerate}[3.]%[resume]
\item A nonzero linear velocity along the $y_t$-axis
\end{enumerate}

1 and 2 are natural constraints, which are determined by the task itself. 3 is artificial constraint determined by the strategy, such as opening or closing a drawer. Under this task specification, rigid manipulator needs to follow the solution that meets all constraints and avoid high contact forces caused by implementation errors through real-time perception.

Soft manipulator exhibits the whole body's inherent safety and passive compliance because of its passive compliance. The reaction force of undesired force, which is generated by the soft manipulator against environmental natural constraints, will change the state of the soft manipulator without damaging the environment or itself. This makes the soft manipulator don't need to planning a precise motion to conform to the natural constraints, and allows planned motion to fight against the natural constraints. So constraint space that describes the interaction task then will change, take the task of opening drawers above as an example, constraint space can be expressed as:

Desired forces and torques:

\begin{enumerate}

\item No constraints
\end{enumerate}

Desired velocities:

\begin{enumerate}[2.]%[resume]
\item A nonzero linear velocity along the $y_t$-axis
\end{enumerate}

Under this task specification, the soft manipulator is able to complete the drawer opening task as long as it has the motion component of $y_t$-axis, so the feasible solution of the task is expanded to a region. The task planning of the soft manipulator is simplified because there are fewer constraints to solve.

Suppose we plan to generate a planned motion as shown in Figure \ref{open_drawer_demo}. Since it does not meet the natural constraints of the drawer, the drawer will generate force F on the arm when planned motion interacts against natural constraints. F will change the motion to comply with natural constraints for the arm’s passive.compliance Therefore, as long as the y-axis motion component conforms to the natural constraints, the soft manipulator can open the drawer with reaction force F, as in Figure \ref{open_drawer_demo}, a planned motion in the green cone area can open the drawer.

%To sum up, because of passive compliance, soft manipulator has the following properties in interaction tasks: (1) soft manipulator does not rely on accurate perception and modeling of tasks, (2) constraints describing the interaction task are reduced, and (3) precise real-time perception of error is not needed in the control process. In this way, the soft manipulator can simplify perception, planning, and control in interaction task.
To sum up, because of passive compliance, soft manipulator don't need to take natural constraints into special consideration when perform the interaction task, which makes soft manipulator does not rely on accurate perception and modeling of tasks. And the feasible solution of the task is expanded to a region, as long as the solution have the kinematic component that corresponds to natural constraints, it is possible to complete the task, which makes soft manipulator does not rely on accurate planning and control. Therefore, soft manipulator can simplify perception, planning and control while performing the interaction task. 

 Based on the analysis of the property simplification, we can define a finite number of  the trend of basic motions and accomplish interaction tasks through the combination of these basic motions. We define the trend of basic motions as atom behaviors, and define the combination of atom behaviors in chronological order as the behavior pattern, and use behavior pattern to perform the interaction task. Then next section are based on this idea to perform the physical experiment.

\subsection{Experimental Method}
In this work, based on HPN arm2, we define six translational atom behaviors in space: moving up and down, moving left and right, moving forward and backward, and four rotational atom behaviors: rotating along x-axis and rotating along y-axis. Because we abandon the DoF of rotating along the longitude direction of the arm, no corresponding atom behavior is defined. The atom behaviors are implemented by the abovementioned estimated model approach. Specifically, a target point which remains a certain deviation with the tip of the arm is defined, and the target point remains relatively static with the tip, so the arm is able to move towards the direction where the target point in. The deviation could be distance and angle of orientation.

In order to show the advantage of the soft manipulator to perform interaction tasks as comprehensively as possible, three kinds of experiments are performed: free space interaction tasks, human-robot interaction tasks and confined space interaction tasks. There are several free space interaction tasks, including cleaning glass, turning handwheel, shifting gear, unscrewing bottle cap, opening drawer and opening door. For each task, corresponding behavior patterns are given as prior knowledge. For example, the behavior pattern of opening and closing a drawer could be “moving forward and backward”, and the behavior pattern of opening a door could be “moving down and backward”. For an interaction task, We define the rigid body that directly interacts with the end effector of the manipulator in the task as the task handle. During the experiments, a task handle is specified, and the manipulator is controlled using estimated model approach to reach the target then grasped the task handle. Then the corresponding behavior pattern is performed. In the experiment, same behavior patterns are used to perform task under different physical configurations, for example, the type of the task may be the same while the position and orientation of the task handle is changed. Furthermore, we use the same behavior patterns to perform interaction tasks in confined space, and the results are compared with that of the experiments in the free space to research the ability of the manipulator to perform tasks in confined space.

Next, the specific experimental setting is proposed:
\begin{enumerate}
\item The demonstration and quantitative analysis of atom behaviors

The pre-defined atom behaviors are demonstrated, and the same behaviors are used to manipulate the slider on the slide rail of different orientation. The position of the slider and the readings of the force sensor attached to the slider are recorded. And the atom behaviors are quantitatively analyzed to research the adaptability of the atom behaviors.

\item Free space interaction tasks

The interaction tasks in free space are classified according to the DoF of the end effector. The same behavior patterns are used to perform the same task under different physical configurations. Several physical quantities that can represent the completeness of the tasks are chosen to be recorded during the experiment. 

\item Human-Robot interaction tasks

The tip of the manipulator is maintained at a specified point. Anthropogenic disturbance are introduced to test the compliance and kinematic redundancy of the manipulator. The position of the tip is recorded to show the basic performance of the HPN arm and the characteristics in human-robot interaction.

\item Confined space interaction tasks

The manipulator is limited. Then experiments of basic performances and interaction tasks are performed to research the ability of the manipulator to work in confined space. 
\end{enumerate}

\subsection{Hardware platform}

\begin{figure}[htbp]
    \centering
    \includegraphics[width=0.95\columnwidth]{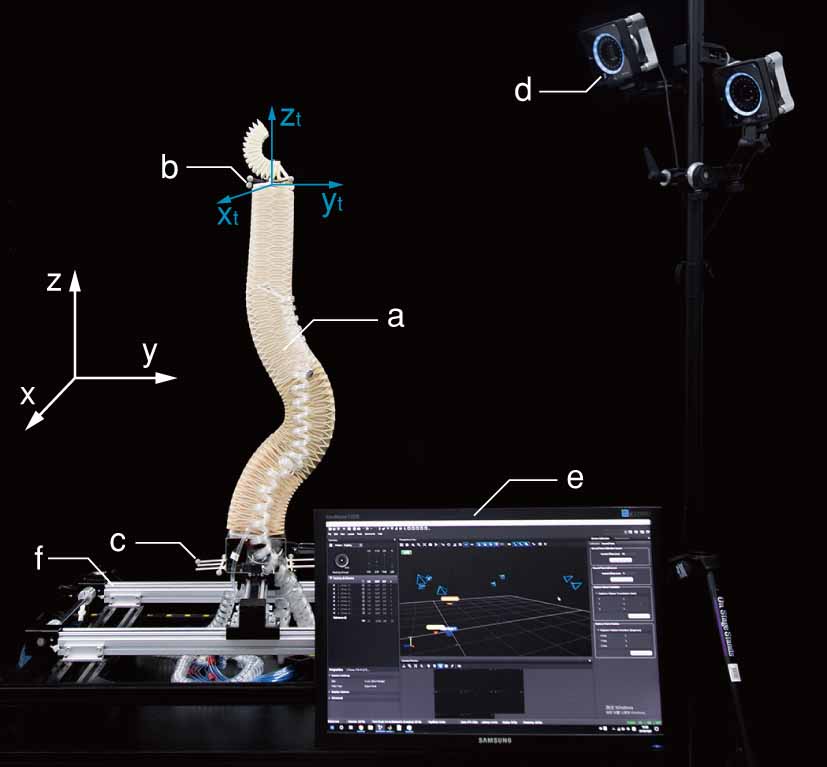}
    \caption{Hardware Platform. $a$ is HPN manipulator, $b$ and $c$ is the markers of tip and base of the HPN Arm, $d$ is the MCS camera, $e$ is PC, $f$ is two-dimensional mobile platform. Note that the coordinate system of the arm tip is the same as the coordinate system of the space in the figure.}
    \label{app_hardware}
\end{figure}

The experiment platform is shown in Figure \ref{app_hardware}. It consists of five parts: sensor system, HPN Arm2, HPN gripper, two-dimensional mobile platform, controller and computing device. The sensor system are used to obtain information. And the computing device is used to process the information, generate the control signal and control the motion of the manipulator through controller. In this experiment, motion capture system (MCS) is used as sensors. Eight MCS cameras are placed evenly across the workspace of the manipulator to obtain information of the manipulator and tasks. A two-dimensional mobile platform controls the movement of the base of the manipulator to expand its reachable space. The controller is based on proportional valves, as described above in section 2.5.

\subsection{Results and discussion}
\subsubsection{Qualitative and quantitative experiments of atom behaviors.}
\begin{figure*}[ht]
    \centering
    \includegraphics[width=0.95\textwidth]{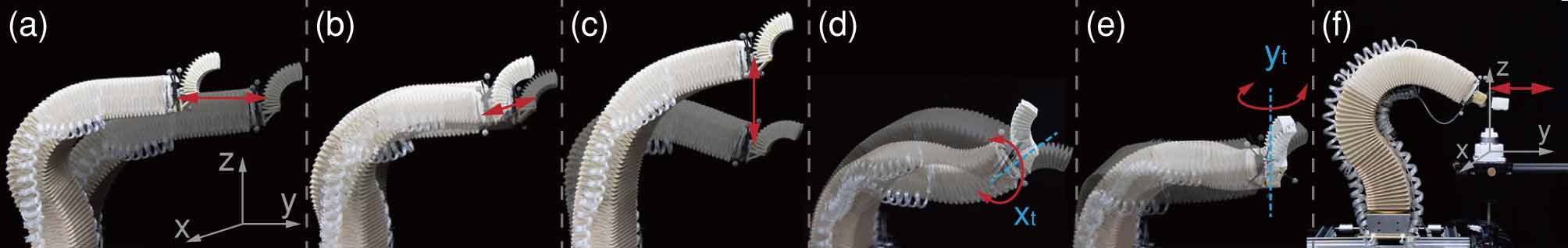}
    \caption{The atom behaviors and their performance on linear slide rail. (f) shows the forward-backward atom behavior.}
    \label{atom_behaviors}
\end{figure*}

Figure \ref{atom_behaviors}(a)-(e) demonstrate five atom behaviors, including the translational motion along three directions, and the rotational motion along x-axis and y-axis. A linear slide rail and a force sensor are used to analyze the atom behaviors qualitatively and quantitatively. Figure \ref{atom_behaviors}(f) demonstrates the manipulator performing the forward-backward motion on the slide rail, the angle and orientation of the slide rail are changed to perform different experiments. The procedure of this experiment is the arm come to grip the handle of the slider firstly, and then performs the same atom behavior with different orientations of slide rail. 

\begin{figure*}[ht]
    \centering
    \includegraphics[width=0.95\textwidth]{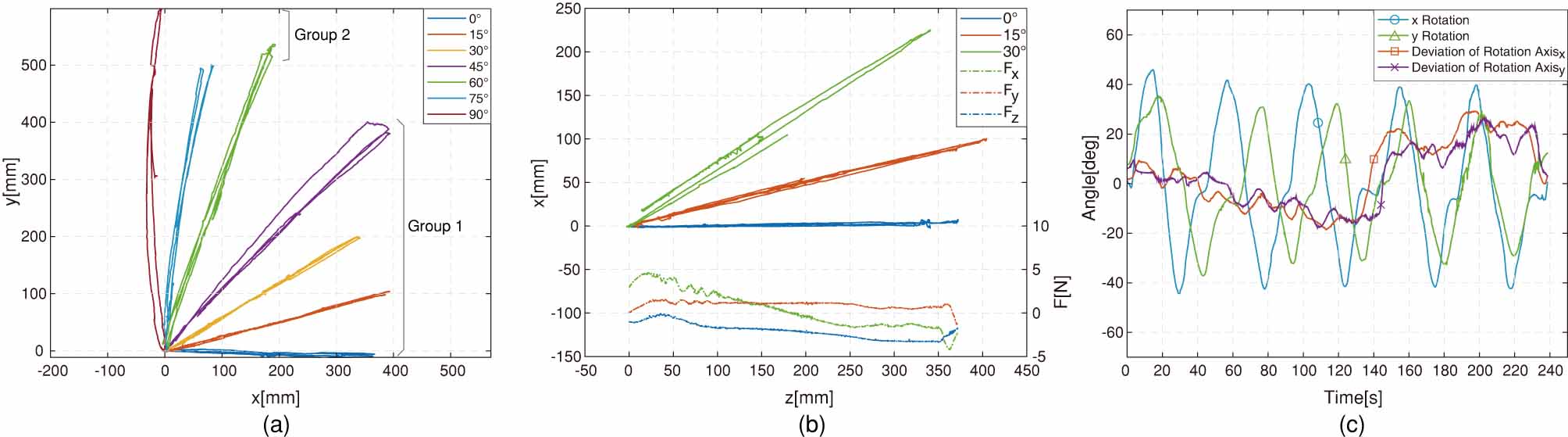}
    \caption{Experiments of basic behavior patterns. (a) demonstrates the trajectory of the slider in the x-y plane. The slide rails are placed in the first quadrant at 15 degrees to each other. And the forward-backward and left-right behaviors are performed to the slider on the slide rails with different angle respectively. The slider are made to move back and forth twice. Group 1 represents the path while performing forward-backward behavior, and Group 2 represents the path while performing left-right behavior. (b) demonstrates the path of the slider within the x-z plane, and the data of the force sensor while moving along z-axis(0 degree). (c) demonstrates the rotational angle while the rotating behavior is performed in the case where the axis of rotation of the hand is not identical with the actual axis of rotation. The curves in this figure are, in turn, the rotational angle of two rotational atom behaviors, the deviation between the rotational axis of slider's handle and the rotational axis of two atom behaviors. }
    \label{atom_behaviors_data}
\end{figure*}

This experiment shows the passive compliance of the soft manipulator, embodied in the adaptability of the atom behaviors. In Figure \ref{atom_behaviors_data}(a), the atom behavior is moving forward and back, while the angle between the direction of the slide rail and the atom behavior is 0 degree, 15 degrees, 30 degrees, and 45 degrees, the interaction between the soft manipulator and the slide rail makes the tip of the manipulator eventually move along the slide rail. Figure \ref{atom_behaviors_data}(b) shows the adaptability of atom behavior along the directions of z-axis. From the curve of force output, it can be inferred that large undesired force is generated along the directions of x-axis and z-axis. Considering Figure \ref{atom_behaviors_data}(a), this shows that the passive compliance of the HPN soft manipulator is approximately isotropic. Figure \ref{atom_behaviors_data}(c) shows the adaptability of rotating behavior. Even if the rotation axis of this atom behavior is not exactly identical with that of the task, the passive compliance of the manipulator would still make it adapt to and execute the specified task.

The experiments of basic behaviors show the manipulator’s adaptability, which is the basis of the manipulator to perform tasks without sensing or modeling the environment accurately. It is also the basis of the manipulator to be compatible with task uncertainty.

\subsubsection{Experiment of free space interaction tasks.}
\begin{figure*}[ht]
    \centering
    \includegraphics[width=0.95\textwidth]{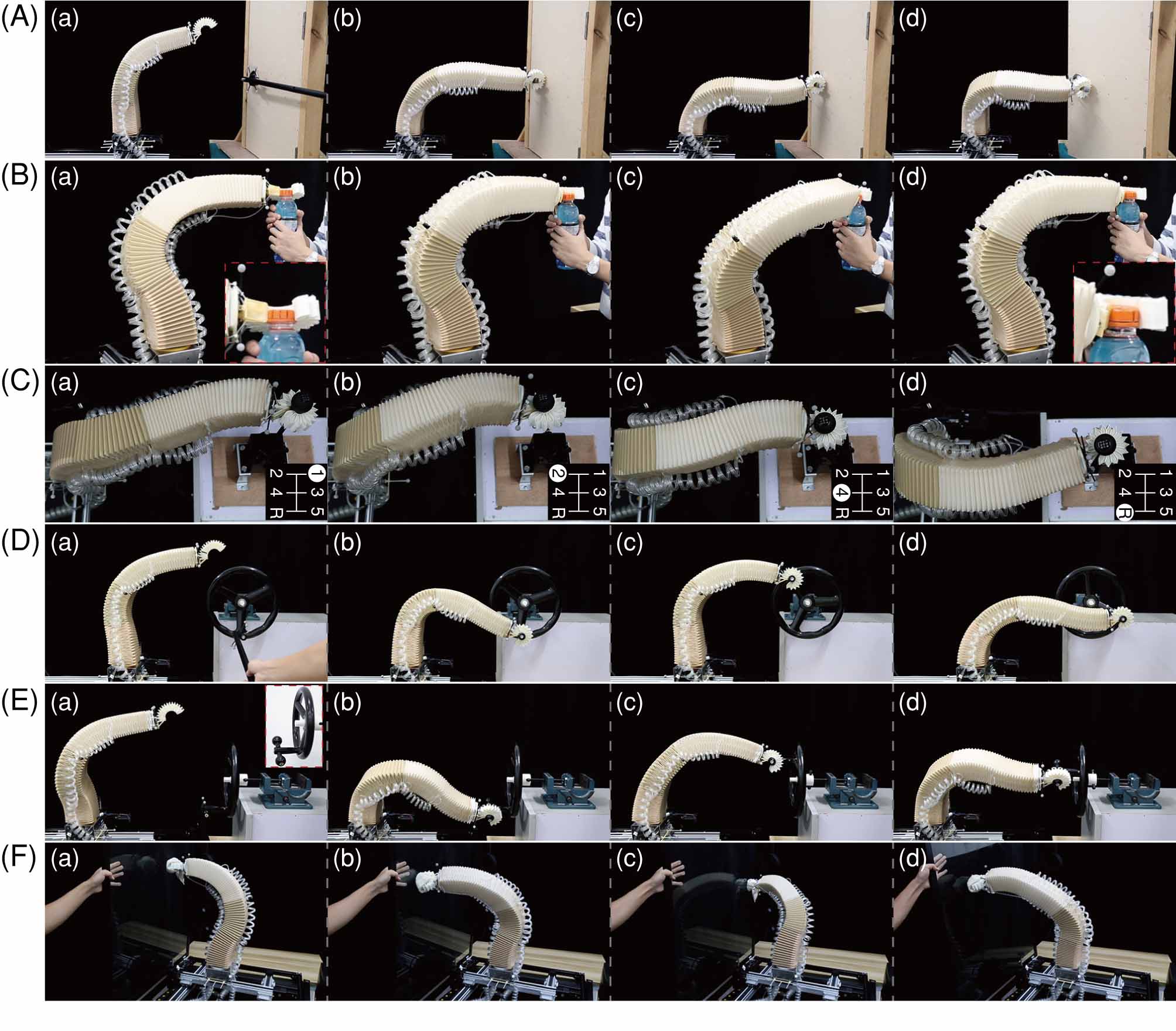}
    \caption{The HPN Arm performed a series of interaction tasks. In the tasks, the DoF of the tip of the manipulator increases step by step. And the same task with different configurations are performed (see Extension 6-10 for details). (A) demonstrates the experiment of opening and closing the door. The task has only 1 DoF. It has three procedure: rotate the door handle, open the door and close the door. (a-d) shows the procedure of opening the door. First, the position of the door handle is specified (a), then the manipulator moves to the handle and grasps it (b), then handle is pressed down (c), and finally the door is opened by being pulled back (d). (B) demonstrates the experiment of unscrewing the bottle cap. This task has only 1 rotational DoF. (a-d) shows the state of the manipulator at different time. From the close-up view at the bottom right corner of (a) and (d), it can be clearly observed that the bottle cap is unscrewed. (C) shows the experiment of shifting gear. The task has two DoFs. The number at the below right corner shows the current gear. (D) shows the experiment of rotating a handwheel parallel to the manipulator. This task has two DoFs. (a-d) shows the process of performing the task. First, the approximate position of the handle is specified (a). Then the manipulator moves to the handle and grasps it (b). Then the corresponding behavior patterns are performed (c-d) until the task is completed. (E) shows the experiment of rotating the handwheel vertical to the manipulator. This task has 3 DoFs. The handwheel is shown at the upper right corner of (a). The procedure is similar to (D), while the pre-defined behavior patterns were different. (F) demonstrates the experiment of cleaning the glass which has 5 DoFs. The tilt angle of the glass is changed during the performance of the task to introduce deviation.}
    \label{applications}
\end{figure*}

An interaction task is characterized by complex contact situations between the manipulator and the environment. The ability to perform interaction tasks is the basis of robots to be applied in unstructured environment. We perform experiments of interaction tasks based on HPN Arm2. The most important characteristic of an interaction task is the degrees of freedom of the end effector during the task. Here interaction tasks with different DoFs are chosen to perform the experiments, including opening drawer, opening door, unscrewing bottle cap, shifting gear, turning hand wheel parallel and vertical to the soft arm, cleaning glass. Prior knowledge is given, and the corresponding behaviors are used to perform the tasks respectively (see Figure \ref{applications}). The results show that the soft manipulator can achieve interaction tasks with different DoFs using simple control strategy, and can be compatible with different configurations in the tasks, for example, the position of the glass, the radius of the hand wheel, the position and the orientation of the drawer.

\begin{figure*}[ht]
    \centering
    \includegraphics[width=0.95\textwidth]{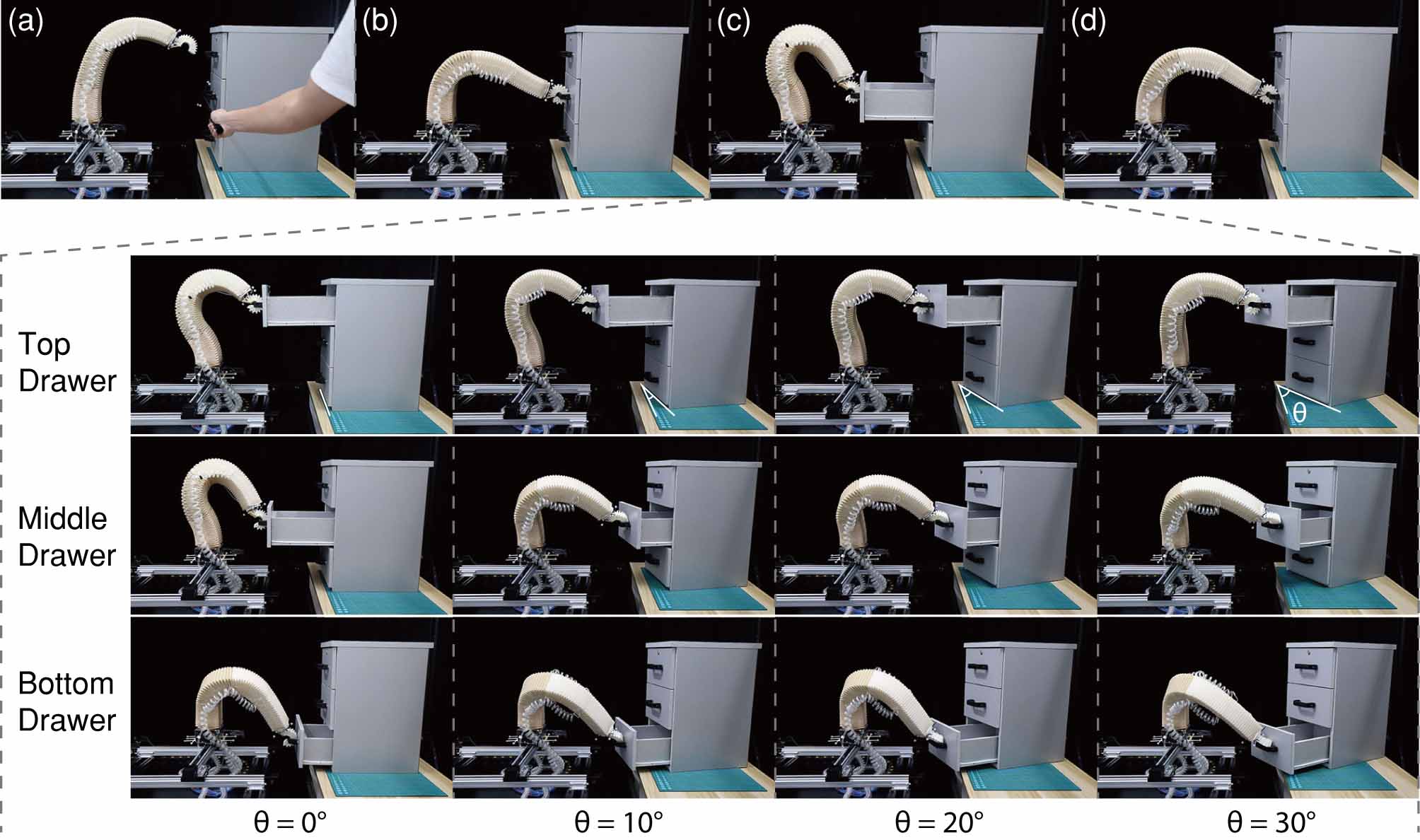}
    \caption{The procedure and results of opening the drawer. It has only 1 DoF. (a-d) shows the procedure. Firstly, the position of the handle is specified (a), then the manipulator moves to the handle and grasps it (b). Then the drawer is pulled back (c), and pushed forward (d).The posture of the soft arm when drawers with different configuration are opened to the maximum. And the nether figure is results of the soft arm opening drawers of different heights (top, medium, bottom) and different angles (0 degree, 10 degrees, 20 degrees, 30 degrees). }
    \label{open_drawer}
\end{figure*}

\begin{figure*}[ht]
    \centering
    \includegraphics[width=0.95\textwidth]{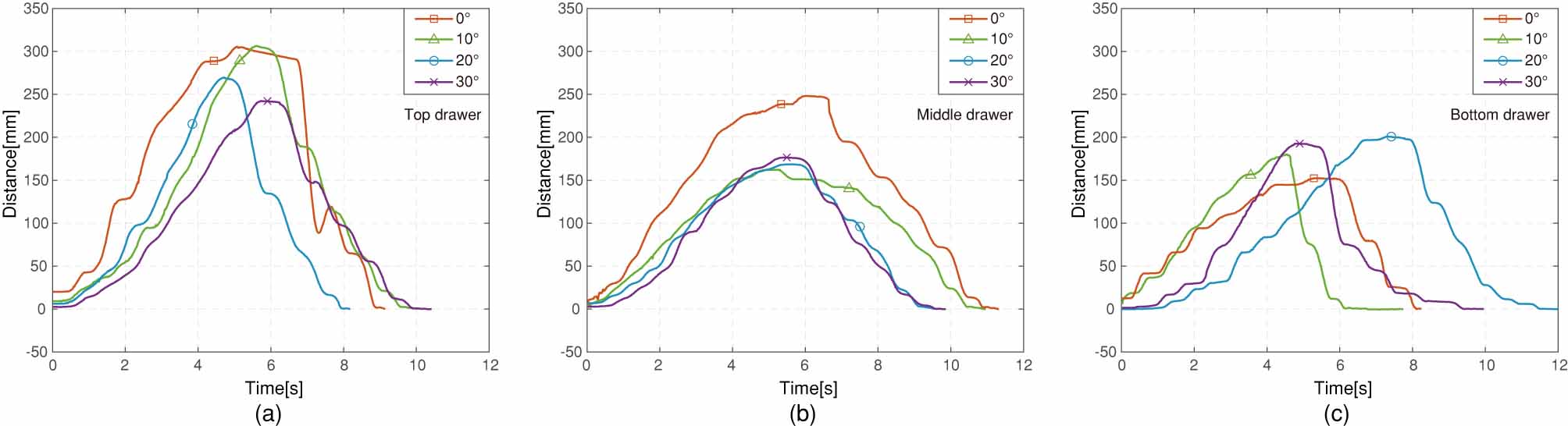}
    \caption{The result of the experiment of pulling the drawer. It shows how the position of the tip change over time when opening drawers with different orientation and different height. Due to the deformation of the arm itself, drawers of different heights or angles can be pulled out at different displacement, ranging from 15 cm to 30 cm. The cliff-like decline on the curve of opening the top drawer at zero degree is a rapid motion after storing energy, this characteristic is analyzed in task 4.}
    \label{open_drawer_data}
\end{figure*}

\paragraph{Free space interaction task 1: Open drawer.}
This task has one DoF, using the two atom behaviors: move forward and move backward. Angle (0 degree to 30 degrees) and height (top, medium, and bottom) of the drawer is changed, but same behaviors are used to perform this task, shown in Figure \ref{open_drawer} (see also Extension 11). Displacement of the drawer being pulled out is selected as the analytical indicator of task completion. Figure \ref{open_drawer} shows the arm posture when the soft arm opens the drawer with different configurations at the maximum angle. It can be seen that the same atom behaviors work differently on the same task with different configurations. This also demonstrates the adaptability of the arm as a result of passive compliance.

\begin{figure*}[ht]
    \centering
    \includegraphics[width=0.95\textwidth]{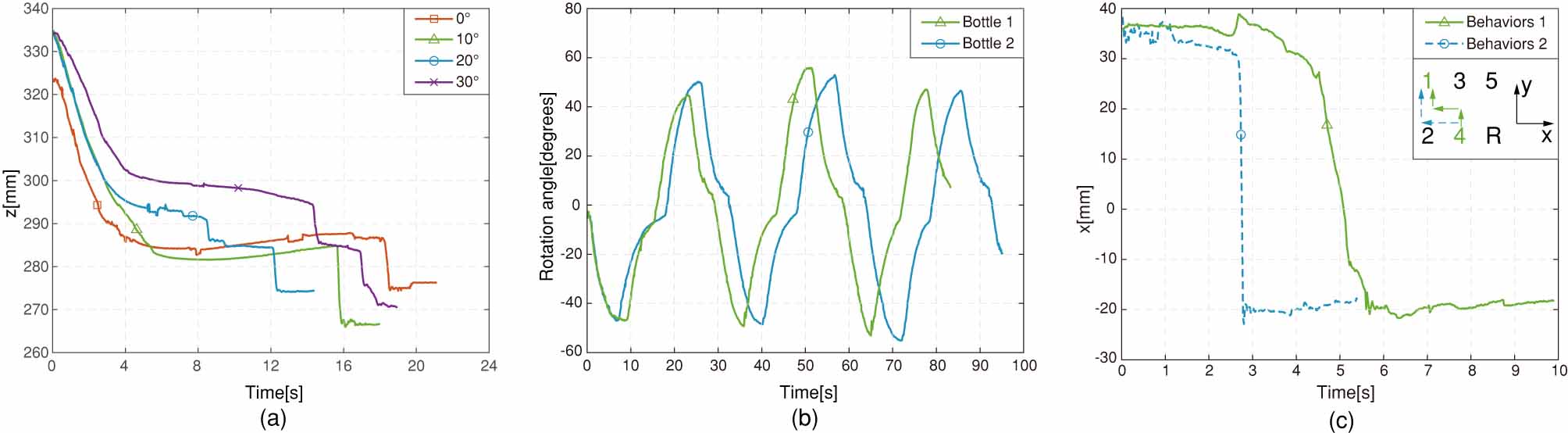}
    \caption{(a) shows how the height of the tip change over time in the experiment of opening the door. The orientation of the door is 0 degree, 10 degrees, 20 degrees, 30 degrees respectively. Approximately, it is necessary to press down 6 cm for opening the door. The cliff-like decline is the rapid motion after storing energy by antagonize static friction. (b) shows the result of unscrewing the bottle cap. The change of the angle between the orientation of the tip and its initial orientation over time is illustrated. The two curves show the results for the bottle caps with different radius. The bottle cap was screwed for three cycles. In each cycle it was screwed for ±45 degrees. (c) shows how the position along the x-axis of the manipulator change with time and two behaviors in the experiment of shifting gear. The velocity along x-axis generated by behavior 2 is much larger than that generated by behavior 1.}
    \label{applications_first_three_data}
\end{figure*}

\paragraph{Free space interaction task 2: Open door.}
As an interaction task, opening/closing the door has three processes: turning the door handle, opening, and closing. We define the atom behaviors as moving down, moving backward, and moving forward respectively. During this task, the orientation of the door is changed, but the same atom behaviors are used to perform the task. This procedure is shown in Figure \ref{applications}(A) (see also Extension 10).
During this task, there is no need to model the rotation axis of the handle and the door and perceive the contact force and torque in real time. The height of the arm tip is selected as the analytical indicator of task completion to analyze the procedure of turning the handle downward. Figure \ref{applications_first_three_data}(a) shows the change in the height of the arm tip in this procedure. After turning the handle, doors at different orientations are opened and closed by moving it backward and forward that similar to opening drawer.

\paragraph{Free space interaction task 3: unscrew a bottle cap.}
Unscrewing bottle cap is the task with one rotation DoF. The predefined atom behavior is to rotate along the plumb line. Along with the gripper, this behavior could rotate different bottle cap, see Figure \ref{applications}(B)(see also Extension 9). Choose the angle between the orientation of the tip and its initial orientation as the task indicator. At the position where we perform the task, the manipulator could output a moment within about 45 degrees.

\paragraph{Free space interaction task 4: shift gear.}
The task of shifting gear has 2 DoFs. The task is performed based on four atom behaviors of the manipulator, which are moving forward, moving backward, moving left, moving right. A certain sequence of gears is specified: Neutral→1→2→3→4→1→R, see Figure \ref{applications}(C) (see also extension 8). Then two modes are defined to perform the task, namely Behavior 1 and Behavior 2. Behavior 1: decompose the shift between gears to basic atom behaviors and then perform them step by step. For example, the corresponding behaviors of 4→1 are: moving forward, moving left and moving forward. Behavior 2: move after storing energy. The shift between gears is also decomposed to basic atom behaviors, and the motion to move left and right are replaced by energy storing motion. Again take shifting 4→1 for an example, first the motion to the left is performed, the tip won’t move due to natural constraints, but the manipulator stores potential energy due to passive deformation. Then the motion of moving forward is performed, and the stored energy is released, the tip could reach 1 immediately. We choose the procedure of shifting 4 to 1, perform the task using these two modes, and record the x coordinate of the tip, see Figure \ref{applications_first_three_data}(c). It can be observed that the manipulator would move rapidly to the left when releasing the stored energy, which makes the mission efficiency increase by 50\%. It shows the ability of the manipulator to store energy, which is significant for tasks related to dynamic interaction.

\begin{figure*}[ht]
    \centering
    \includegraphics[width=0.95\textwidth]{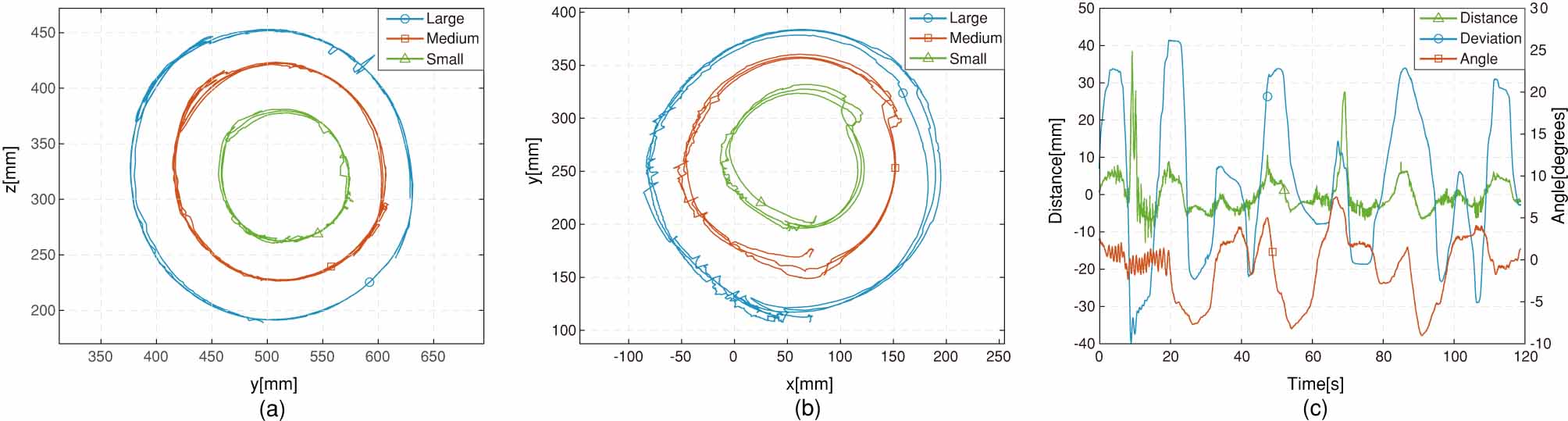}
    \caption{The result of the experiments of interaction tasks. (a) shows the result of rotating the handwheel parallel to the manipulator. The trajectory of three circles of motion is illustrated. The different colors of the path represent the handwheels of different radius. (b) shows the result of rotating the handwheel vertical to the manipulator. (c) shows how the recorded physical quantities of cleaning the glass change over time. Specifically, the data are the distance between the tip and the glass, the projection of the deviation of the tip from its initial position on the surface of the glass, and the angle between the glass and the vertical plane. The change of the configuration includes two stages. First, small changes are applied to the orientation of the glass (0s-20s), and then the glass swings with a large amplitude (20s-120s). The amplitude of the motion of the manipulator is approximately the same under different disturbances. And in either of the stage, the manipulator detached from the glass once and then resumed contact with the glass in a short time.  }
    \label{applications_last_three_data}
\end{figure*}

\paragraph{Free space interaction task 5: turn a parallel hand wheel.}
A handwheel with 2 DoFs is chosen and placed parallel to the manipulator as shown in Figure \ref{applications}(D). The task is performed using a behavior pattern including four atom behaviors: move backwards, move up, move forward, move down (see extension 7). Three kinds of handwheels with different radius are used. And the trajectory of the handle in 2D plane is recorded. It can be inferred from Figure \ref{applications_last_three_data}(a) that the same behavior patterns could be used to perform the task of turning handwheel with different radius. In the process, there is no need to sense the radius of the handwheel. 

\paragraph{Free space interaction task 6: turn a vertical hand wheel.}
In this task, a handwheel with three DoF is chosen. And it is placed vertical to the manipulator as shown in Figure \ref{applications}(E). A behavior pattern including four atom behaviors, namely left, up, right and down is used to perform the task (see extension 7). Figure \ref{applications_last_three_data}(b) illustrates the trajectory of the handle. Comparing with task 5, it could be found that the behavior patterns predefined for the two kinds of tasks are almost the same. So it can be concluded that, a single behavior pattern has the potential to perform interaction tasks with different DoFs. This shows the adaptability of behavior patterns at the interaction task level.

\paragraph{Free space interaction task 7: clean glass.}
While performing the task of cleaning glass, the tip has 5 DoFs. The atom behaviors is specified as pushing on the glass and moving left or right according to prior knowledge. This behavior could be derived combining forward and left/right. And this behavior is used to perform the task. In this process, the orientation of the glass was changed all the time, so the relative position between the glass and the manipulator is changed. While performing the task, there is no need to sense the exact change of the orientation of the glass and the contact force. It is not necessary to modify the control strategy neither (See extension 6). Figure \ref{applications_last_three_data}(c) shows the orientation of the glass and the position of the tip relative to the glass. The position and orientation of the tip (not the gripper) and a point attached to the glass are recorded. The position of the gripper is represented in the Coordinate System of the glass. Its z coordinate ( the z-axis is orthogonal to the plane of the glass) is taken as the deviation from the glass, and the distance between the current position of the tip and its initial position represents the distance. In the actual experiment, when the tilt angle of the manipulator is too large or the manipulator is at the far left and the far right of the workspace, the manipulator is observed to detach from the glass, but it could resume their contact in a short time. The adaptability of the manipulator shown in this experiment is also meaningful in many other tasks, such as moving along a curved surface.

\subsubsection{Experiment of human-robot interaction task. }
Human-robot interaction has a wide range of significance in the application of robots in unstructured environments. To demonstrate the human-robot interaction feature of the manipulator, the tip of the manipulator is held at a specified point in the workspace. We interact with the manipulator actively and test the compliance and the kinematic redundancy of the manipulator. The position of the tip is recorded to show the human-robot interaction features of the HPN arm.

\begin{figure*}[ht]
    \centering
    \includegraphics[width=0.95\textwidth]{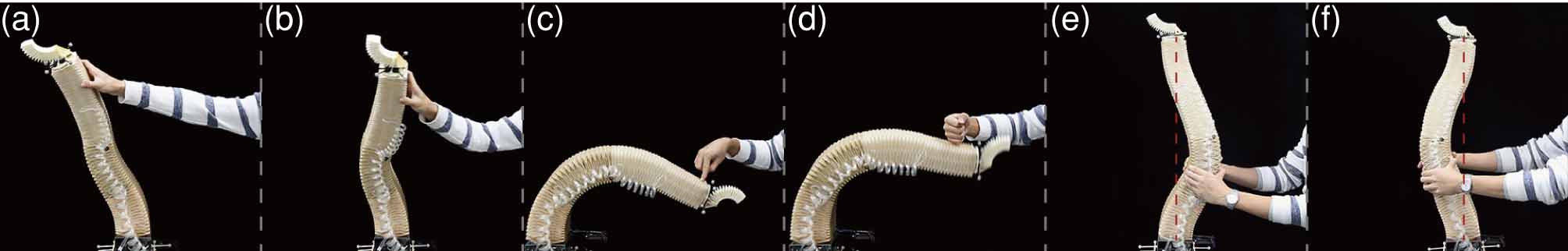}
    \caption{Experiments of human-robot interaction task based on HPN Arm. (a) and (b) demonstrate the wide range of compliance of the manipulator. (c) and (d) show the disturbance resistance of the manipulator.  (e) and (f) demonstrate the kinematic redundancy of the manipulator.}
    \label{hold}
\end{figure*}

\begin{figure*}[ht]
    \centering
    \includegraphics[width=0.67\textwidth]{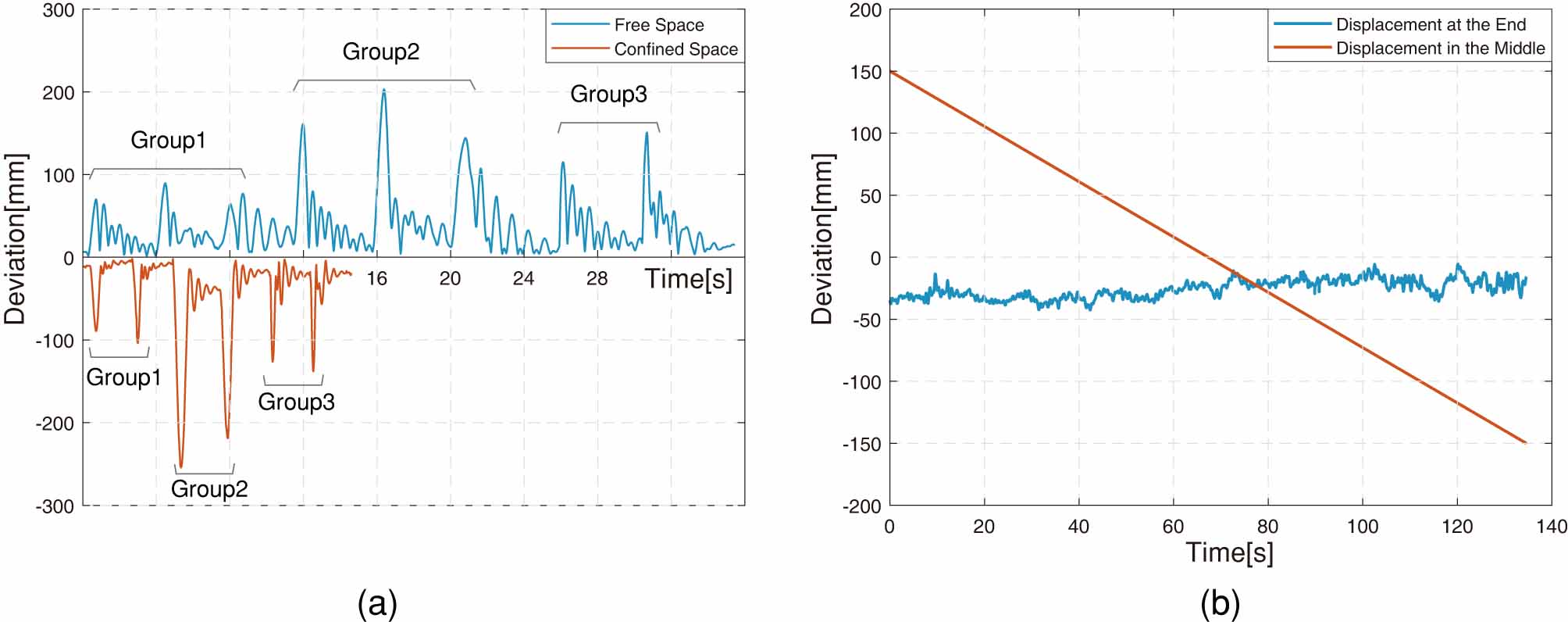}
    \caption{Results of human-robot interaction task. (a) compares the experiments of human-robot interaction tasks in free space and confined space in which the two segments nearest to the base of the manipulator is confined. The specific properties in confined space will be analyzed in experiments of comfined space interaction tasks. Group 1 is the deviation when small disturbances are introduced, group 2 is the deviation when large disturbances are introduced, and group 3 is the deviation when the instantaneous impact is introduced. The different peaks in each group are the disturbances in different directions at the same position. (b) illustrates experiments of kinematic redundancy. Disturbances of about ±15 cm were performed at the medium segment, and the deviation of the tip is within 4 cm.}
    \label{hold_data}
\end{figure*}

The tip of the manipulator maintained at a specified point in the workspace, and different kinds of interactions are introduced, see Figure \ref{hold} (see also Extension 5). The interactions are divided into three kinds according to the duration and the amplitude: small disturbances (5-7 cm), large disturbances (15-20 cm), and instantaneous impact of three directions at the tip. The highest point of the workspace and a certain trivial point were chosen as target points to perform the experiment. The deviation of the tip of the manipulator from the target point is recorded and the relation between the deviation and time is illustrated in Figure \ref{hold_data}. The configuration of the kinematic redundancy experiment is introducing anthropogenic disturbance at the medium segments, and then the deviation of the tip of the manipulator from the target is recorded. (See Figure \ref{hold}(e)(f), Figure \ref{hold_data}(b))

Figure \ref{hold_data}(a) shows the response of the manipulator to the disturbances and impact at different time. In the whole process, the position, direction, type of contact are not perceived. This experiment shows the characteristic of the HPN arm to interact with people. In order to achieve this, traditional rigid robots need to implement active compliance using force control and compliant components such as series elastic actuators. In the human-robot interaction, the soft manipulator responds rapidly and has a low computational cost. Those inherent properties are not limited to situations of human-robot interaction. They are also useful in the case where robots interact with the unstructured environment.

The soft manipulator also shows kinematic redundant properties, see Figure \ref{hold}(e)
(f)(see also Extension 5). When there are disturbances in the medium of the manipulator, the manipulator can adjust its posture within a certain range to keep the position and orientation of its tip almost unchanged. Comparing with the traditional rigid robot with kinematic redundant, the implementation of the kinematic redundant of the soft manipulator does not need complex control method, nor does it need to sense the external force. 

\subsubsection{Experiment of confined space interaction tasks. }
In unstructured environments, it’s common for the arm to be confined. Next, we will demonstrate the basic performance of the soft arm in a confined space, its ability to interact with people and perform interaction tasks. In the experiment of confined space interaction tasks, point tracking and interaction tasks were performed with the two segments near the base of the arm are limited.

\begin{figure*}[ht]
    \centering
    \includegraphics[width=0.95\textwidth]{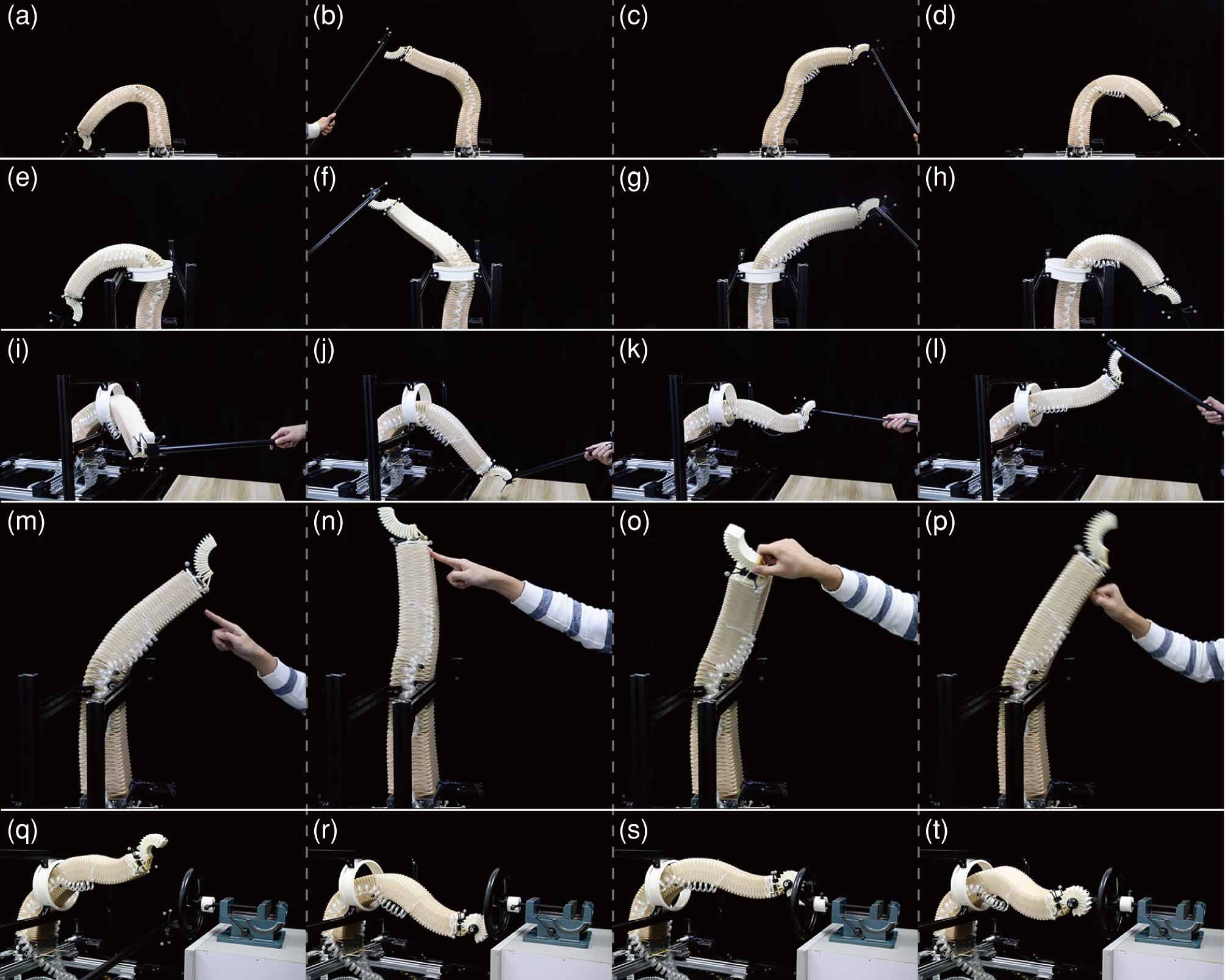}
    \caption{Experiment of confined space interaction tasks. (a - d) are the point tracking in free space. (e - h) are the point tracking with constraints in the vertical plane. Four points in the same position as that in free space were selected for comparison. (i - l) are the point tracking with constraints in horizontal plane; (m - p) is the compliance demonstration in confined space, including disturbance (m) (n) and impact (p); (q - t) are the procedures by which the arm performs the interaction task of rotating a hand wheel vertical to the arm in the confined space: first the position of the handle is specified (q), then the manipulator moves to the handle and grasps it (r), perform the behaviors to complete the task (s) (t). }
    \label{confined_space}
\end{figure*}

In the confined space, since the arm will interact with the environment and there will be constraints, the HPN arm will produce movement that complies with the constraints. 
 
Figure \ref{confined_space} shows properties of the soft arm in confined space (See extension 12). Comparing with what happens in free space (a-d), the posture of the arm to reach the same position may be different (e-h) and the workspace of the soft arm is limited (i-l) in confined space. (m-p) are the demonstration of human-robot interaction of the soft arm in confined space, which show that the end of the soft arm still has complete passive compliance in confined space, and the deviation of the tip of the manipulator from the target point is recorded that shown in Figure \ref{hold_data}. The HPN arm can recover faster from disturbances in confined space because effective length of the arm is shorter. Rotating a hand wheel vertical to the arm, which requires relatively complex motions and three degrees of freedom, is chosen in the experiment of confined space interaction tasks. During this task, the arm interacts with the environment and is forced to change its posture. Thus soft arm can still have a certain output space in confined space. For a hand wheel (medium size) larger than this tube, the arm is still able to rotate it using limited atom behaviors (See extension 12).
 
In summary, the human-robot interaction tasks, free space interaction tasks and confined space interaction tasks of the soft arm are demonstrated in this section. In these tasks, traditional robots might need to know the position, task configuration, constraints, etc., and plan its motion carefully. We demonstrates the ability of the soft arm to perform tasks involving interaction without relying on accurate perception, planning, and control. The control of all tasks is based on simple behaviors defined in orthogonal reference frames, demonstrating the simplification and advantages of soft arms in interaction tasks.

\subsection{Application Summary}

Continuous passive compliance is the main characteristics that distinguish soft manipulators from rigid manipulators. Passive compliance not only brings flexibility and safety to soft manipulators but also defines a new way for manipulators to interact with the environment, which is the real advantage and potential of soft manipulators. Understanding this interaction advantage is conducive to promoting the application of soft robots and the progress in robotics. 

In this paper, we propose that the advantages of the soft arm should be reflected in the interaction, explains why the soft arm can simplify the interactive tasks, and gives methods based on atom behaviors to take advantage of it. A large number of experiments have been done to prove these methods and explanation.

Our work only shows the advantages of soft robots in interaction tasks, but the potential goes beyond that. For example, because of tasks are simplified, it's easier to create higher-level planners to accomplish complex tasks. As a start, this paper hopes to trigger more applications of soft robots, which in turn will attract more researchers and resources, and ultimately promote the rapid development of this field.

\subsubsection{Limitation}

\begin{enumerate}[(a)]
\item Since our soft arm does not have DoF of torsion, if there is torsion during the task, the fingers will twist and generate relatively large torsion on the manipulated object. This problem can be solved to some extent by adding DoF of torsion to the gripper. There will be some problems if torsion is required in the task because our arm doesn't have the ability to twist. But we can still output two other torsions if the arm can move or change its direction. In addition, we are going to design a soft arm with torsion DoF like an elephant trunk.

\item This application section is a preliminary exploration. For example, the location of the handle is given through the MCS marker and baton. But it can be easily improved by our visual algorithms.

\item In this section, atom behaviors are given artificially as prior knowledge. In future work, high-level planning will be used to automatically generate behaviors based on different tasks.

\item In all of our experiments, the base of the arm is fixed. If the base is movable, the arm will be more flexible and able to perform more complicated tasks. For example, the door can be opened wider if the base is movable. When designing the hardware, we designed a step motor platform. In the following work, we will explore the performance of the arm combined with this platform. In the future, we will try to combine the arm with the mobile robot to further explore the possible application such as human-robot interaction tasks.

\item Now we use motion control to perform tasks, and in some cases where fixed point force output is indispensable, there will be a problem: undesired force might be output on undesired directions because estimated model control is not accurate and feedback is needed to eliminate the deviation. However, since the tip of the arm is fixed and the marker does not move, the deviation caused by the  estimated model control will always exist, and there is a certain probability for this deviation to increase. Adding markers to the middle of the arm, rather than relying on estimation method, can alleviate this problem to some extent. But only by accomplishing force control with force sensors can this problem be solved.

\end{enumerate}

\section{Conclusion}
% 本文系统性的探究了软体手臂，从设计、制备到控制再到应用的各个方面的问题。以做出类似象鼻一样灵活、强大的软体手臂为目标，提出三个关键问题：如何设计大负载的软体手臂？软体手臂是否可以建模？软体手臂有什么优势？文章围绕三个问题展开，每一部分都给出了最终有可能形成理论的猜想与判断，并在其指导下给出 promising 的方法。大量的实验都一定程度上证实了提出的猜想与方法的合理性。

% HPN设计 architecture 及相应的整套系统（包括优化方法、制备方法、动力系统），在控制和应用的探究中，被证明是稳定、有效且易用的：我们在做控制、应用的时候基本不需要再关注设计相关的问题，具体来说， HPN arm3整套系统在高强度使用半年以上的时间（每天平均5小时）中，并没有出现手臂结构疲劳、老化、破损，气动动力系统失效等任何问题，偶尔（总共不到3次）出现的气囊漏气问题，很容易（几分钟之内）就可以通过更换气囊解决。这意味着软体机器人平台有望在科研领域商业化、标准化。这将有助于改变目前搞软机器人需要自动从头制备的局面，使得在其他方面擅长（如视觉、人工智能、人机协作等）而并不擅长设计制备的团队能够直接投入力量到软体机器人领域的研究中。这样软体机器人这个非常需要多学科共同努力的领域才能真正高速发展。

% Application部分的有所任务都是在model-less控制方法的基础上完成的，整个过程中，除了上述涉及较大输出力的时刻，都高效又稳定（可观看视频）。证明了这一简单的控制方法在解决软体手臂的控制问题上的有效性。至此，整个领域不应该再说软体机器人手臂难控制，只能说软体机器人手臂相对刚性手臂，需要任务空间的反馈才能实现高效控制。

% 目前机器人手臂具有成功应用的场景就是工业自动化，很多人会去想软体手臂如何应用到工业应用中。鉴于软体机器人的模型很难建得像刚性机器人那样精准，其控制的精度与效率跟刚性机械臂没有可比性，所以我们判断软体手臂的应用肯定不包括工业自动化。更精准、更高效，一直是推动孕育于自动化需求下刚性机器人及相应技术体系发展进步的内在动力。而软体机器人的应用并不在工业自动化领域，因此——特别是在领域的起步探究阶段——没有必要过分追求精准、高效。机器人学是以应用为目的的学科，所有理论、技术都应围绕应用展开（脱离应用容易走偏）。具体说来，像控制方面，根据需求去搞控制，当搞出满足当下需求的 controller 就可以去探究应用了，根据探究中的具体问题和新需求再去提升 controller。

% 为什么说软体机器人的优势在于交互？反过来思考，软体机器人与柔性关节的连杆机器人和做到主动柔顺的刚性机器人的区别在于无限自由度。而由于软体机器人的驱动数量有限，无限自由度的特性只有在交互中才能发挥体现。

% 软体机器人交互的未来，不只只是本文展示出来的 interaction tasks。强人工智能很可能会因为软机器人的发展而产生。

% 一场技术变革正将从软体机器人领域引发，会产生全新的技术体系。本文，历时数年，展示的这一整套系统的工作，作为一个简单的框架，有望促进技术变革的出现与新技术体系的发展。

This paper systematically discusses the design, fabrication, control, and application of the soft arm. With the goal of creating a flexible and powerful soft arm like an elephant trunk, we ask three key questions: how to design a soft arm with large load capacity? Can the soft arm be modeled accurately? What are the advantages of the soft arm? This paper focuses on these three questions, each part of which gives hypotheses and judgment that may eventually form a theory and gives promising methods under their guidance.

In our work of control and application, the HPN system (including the hardware, fabrication methods, etc.) was proved to be very stable, durable and maintainable. We have been testing this system for over 1000 hours, but there have never been any problems such as structure fatigue, aging, damage, failure of the aerodynamic power system, etc. The occasional airbag leakage can be easily fixed by replacing the airbag within minutes. This means that soft robot has the potential for standardization and commercialization. This will help to change the current situation that soft robotics researcher needs to begin with the materials and fabrication. Researchers who are not professional of them can focus on their expertise such as computer vision, artificial intelligence, etc. In this way, soft robotics, a field that requires a lot of interdisciplinary efforts, can really develop at a high speed.

In addition to a reliable platform, an effective control algorithm is also critical. All tasks in the application section were completed efficiently and stable with our estimated model control (see video), which proved the effectiveness of this control strategy in the control of the soft arm.

Precision and efficiency have always been the inner drives for the development of rigid robots and corresponding technologies, which have been widely applied to industrial automation. Since it is difficult to model a soft robot as accurately as a rigid robot, its control accuracy and efficiency are not comparable with that of a rigid robot. The applications of soft robots may not be industrial automation, so there is no need to be overly precise and efficient, especially in the early stage of soft robotics. As an applied science, all theories and technologies in robotics should focus on application. So in the control aspect, controllers that meet the current needs can be applied to explore new applications and can be improved according to subsequent specific problems.

The advantage of soft robots lies in active interaction. Their main advantage over multiple soft link robots and rigid robots with active compliance are infinite DoFs. Because of the limited number of actuators in a soft robot, the characteristics of infinite DoFs can only be distinguished in interaction.

The future of soft robot interaction is not only what we presented in this paper. A technological revolution is taking place in the field of soft robotics, with a set of new technologies. Strong artificial intelligence is likely to result from the development of soft robots. As a simple framework, this systematic work is expected to promote the emergence of the technological revolution and the development of a new technological system.

\begin{funding}
This research is supported by the National Natural Science Foundation of China under grant 61573333.
\end{funding}

\bibliographystyle{SageH}
\bibliography{ref}

\begin{appendices}

\begin{table*}[htbp]
\centering 

\section{Appendix: Index to Multimedia Extensions}
\caption{Table of Multimedia Extensions}
\label{Table of Multimedia Extensions}
\begin{tabular}{p{50pt}p{40pt}p{368pt}}
\toprule

Extension& Media type& Description\\
\hline
1& Video& This is a video to demonstrate the basic structure of Honeycomb PneuNet and HPN Arm 1. Based on HPN Arm 1,  Low-level control is implemented to perform several tasks.\\

2& Video& This is a video to demonstrate 2D Two-level approach of HPN Arm control. Positioning and tracking experiments are carried out to evaluate the effectiveness of this control method. \\

3& Video& This is a video to demonstrate 3D Estimated-model based control of HPN Arm. Tracking experiments are carried out to evaluate the effectiveness of this control method. Pick and Place are demonstrated as well.\\

4& Video& This is a video to demonstrate 2D Model-free control scheme based on HPN Arm. The training process of Q-Learning is demonstrated. Positioning task is chosen to evaluate the robustness of this control method.\\

5& Video& This is a video to demonstrate the human-robot interaction characteristic of HPN Arm 2. Passive compliance and kinematic redundancy of soft manipulator are demonstrated. \\

6& Video& This is a video to demonstrate HPN Arm 2 cleaning glass.  The relative position of end-effector and glass are changing manually.\\

7& Video& This is a video to demonstrate HPN Arm 2 turning hand-wheel. Different size hand wheels are put parallel and vertical to the Arm.\\

8& Video& This is a video to demonstrate HPN Arm 2 shift gear. Energy storage effect is observed in this process. \\

9& Video& This is a video to demonstrate HPN Arm 2 open bottle. Two different bottles are opened with the same behavior.\\

10& Video& This is a video to demonstrate HPN Arm 2 open doors with different configuration by the same behavior. \\

11& Video& This is a video to demonstrate HPN Arm 2 open drawers with different configuration by the same behavior.\\

12& Video& This is a video to demonstrate HPN Arm 2 perform confined space interaction tasks.\\

\toprule
\end{tabular}

\end{table*}

\end{appendices}

\end{CJK*}
%中文

\end{document}